\documentclass[pdflatex,iicol,sn-vancouver-ay]{sn-jnl}

\usepackage{todonotes}
\usepackage{xcolor}
\usepackage{colortbl}
\usepackage{tabularx}
\usepackage{paralist}
\usepackage{multirow}
\usepackage{threeparttable}
\usepackage{subfig}
\usepackage{adjustbox}

\usepackage{float}

\usepackage{graphicx}%
\usepackage{multirow}%
\usepackage{amsmath,amssymb,amsfonts}%
\usepackage{amsthm}%
\usepackage{mathrsfs}%
\usepackage[title]{appendix}%
\usepackage{xcolor}%
\usepackage{textcomp}%
\usepackage{manyfoot}%
\usepackage{booktabs}%
\usepackage{algorithm}%
\usepackage{algorithmicx}%
\usepackage{algpseudocode}%
\usepackage{listings}%

\usepackage{eso-pic}
\usepackage{xcolor}

\theoremstyle{thmstyleone}%
\theoremstyle{thmstyletwo}%

\theoremstyle{thmstylethree}%

\newcommand{\bms}[1]{{\small\textbf{#1}}}

\begin{document}
	
\AddToShipoutPictureFG*{%
	\AtPageUpperLeft{%
		\raisebox{-15mm}[0pt][0pt]{%
			\makebox[\paperwidth][c]{%
				\parbox{0.92\paperwidth}{%
					\centering
					\color{gray}
					\fontsize{8.5}{9.5}\selectfont
					\textbf{Note:}
					This preprint has not undergone peer review or any post-submission
					improvements or corrections. The Version of Record of this article is
					published in the \textit{International Journal of Computer Vision}, and is
					available online at
					\href{https://doi.org/10.1007/s11263-026-03001-z}
					{https://doi.org/10.1007/s11263-026-03001-z}.
					
					\textbf{For citation, please refer to the peer-reviewed Version of Record.}
					
					View-only full text:
					\href{https://rdcu.be/fCvDT}{https://rdcu.be/fCvDT}.
				}%
			}%
		}%
	}%
}

\title[Article Title]{E-RGB-D: Real-Time Event-Based Perception with~Structured~Light}


\author*[]{\fnm{Seyed Ehsan} \sur{Marjani Bajestani}}\email{ehsan.marjani@polymtl.ca}

\author[]{\fnm{Giovanni} \sur{Beltrame}}

\affil[]{\orgdiv{Department of Computer Engineering and Software Engineering}, \orgname{Polytechnique Montreal}, \orgaddress{\street{2500 Chem. de Polytechnique}, \city{Montreal}, \postcode{H3T 0A3}, \state{Quebec}, \country{Canada}}}


\abstract{Event-based cameras (ECs) have emerged as bio-inspired sensors that report
	pixel brightness changes asynchronously, offering unmatched speed and
	efficiency in vision sensing. Despite their high dynamic range, temporal
	resolution, low power consumption, and computational simplicity, traditional
	monochrome ECs face limitations in detecting static or slowly moving objects
	and lack color information essential for certain applications. To address these
	challenges, we
	present a novel approach that integrates a Digital Light Processing (DLP)
	projector, forming Active Structured Light (ASL) for RGB-D sensing. By
	combining the benefits of ECs and projection-based techniques, our method
	enables the detection of color and the depth of each pixel separately. Dynamic
	projection adjustments optimize bandwidth, ensuring selective color data
	acquisition and yielding colorful point clouds without sacrificing spatial
	resolution. This integration, facilitated by a commercial TI LightCrafter 4500
	projector and a monocular monochrome EC, not only enables frameless RGB-D
	sensing applications but also achieves remarkable performance milestones. With
	our approach, we achieved a color detection speed equivalent to 1400 fps and
	4~kHz of pixel depth detection, significantly advancing the realm of computer
	vision across diverse fields from robotics to 3D reconstruction methods. Our
	code is publicly available: \href{https://github.com/MISTLab/event_based_rgbd_ros}{github.com/MISTLab/event\_based\_rgbd\_ros}}




\maketitle

\section{Introduction}\label{sec1}

Event-based cameras (ECs) are innovative sensors that detect changes in pixel
brightness asynchronously. Unlike traditional frame-based cameras, ECs do not
capture full images, they generate events containing pixel coordinates, timestamps,
and the polarity of brightness changes whenever a change exceeds a certain
threshold. These sensors enable researchers to detect movement and changes at
extremely high speeds with very low latency, minimal power consumption, and low
bandwidth requirements, as shown by~\cite{Bajestani_2025_RoboCup}. It makes ECs
highly suitable for high-speed vision-based applications such as depth estimation.

\begin{figure}[htbp]
	\centerline{{\includegraphics[width=0.5\linewidth]{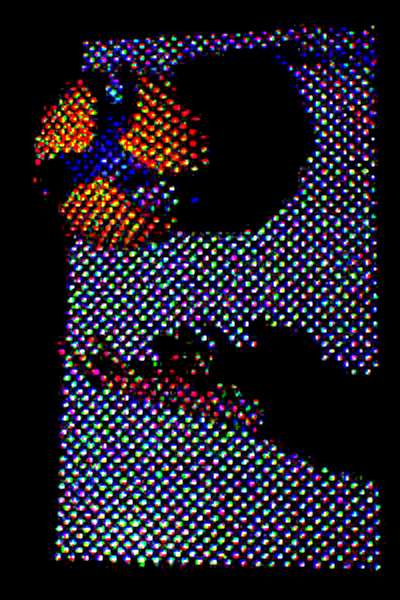}}
		{\includegraphics[width=0.5\linewidth]{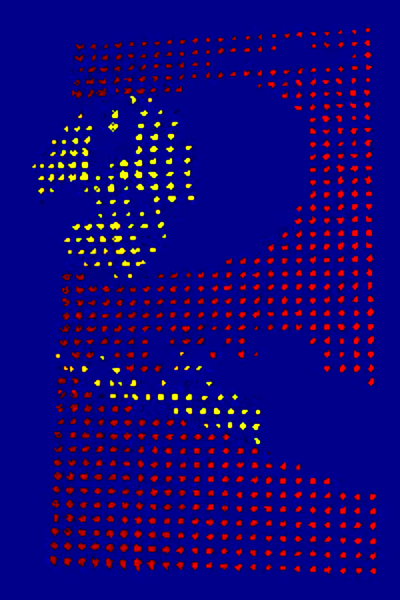}}}
	\caption{Color (left) and depth (right) detection of a volleyball thrown
		in front of a VGA monocular event-based camera, reconstructed at 120 fps
		from \(\sim 1.5\)~m using the proposed method.}
	\label{fig:ball_1}
\end{figure}

Depth estimation is a vital element in both computer vision and robotics, used in
applications like 3D modeling, augmented reality, and navigation. Structured Light
(SL) systems, which project known patterns onto a scene and observe the
deformations with a camera, have traditionally been used for this purpose as described by ~\cite{jason2011dlp}.
While accurate, these systems are limited by factors such as device bandwidth and
projector light power, impacting acquisition speed and performance under various
lighting conditions. The high temporal resolution and high dynamic range (HDR) of
ECs can address these limitations, as their asynchronous nature allows for fast,
efficient data capture without the redundancy seen in frame-based systems. However,
despite their advantages, traditional monochrome ECs have limitations in capturing
color information.

In our previous work by~\cite{Bajestani_2023_WACV}, we proposed combining an EC with a
Digital Light Processing (DLP) projector to form an Active Structured Light (ASL)
system for color sensing with a monochrome camera.
Unlike passive color event cameras, which are currently limited in resolution and availability, our method actively generates color signals using structured projection, enabling robust reconstruction even under low-light or high-speed conditions.
We extended our previous work and
explained how we could achieve event-based color and depth measurements for each pixel separately.
Integrating DLP projectors with ECs overcomes the constraints of traditional SL systems. The ECs'
ability to suppress temporal redundancy and their high dynamic range enables
effective operation in diverse lighting conditions, enhancing the depth estimation
process. Additionally, the dynamic projection adjustments allow for selective color
data acquisition, ensuring efficient use of bandwidth while maintaining spatial
resolution. Our method enhances depth and color detection, providing robust E-RGB-D
sensing at 1.4 to 4~kHz per pixel. The main contribution is achieving ultra-fast and
real-time, event-based color and depth measurements per pixel that can work in different
situations within the scene.

Unlike frame-based methods such as X-maps, as proposed by~\cite{morgenstern2023x},
which require time-discretized event accumulation and rely on raster-scanned MEMS projectors,
as described by~\cite{brandli2014adaptive},~\cite{matsuda2015mc3d}, and~\cite{muglikar2021esl},
our system operates fully asynchronously. Each event is independently tagged with depth and color
without needing to reconstruct an intermediate 2D map or wait for a complete pattern sequence.
This frameless design enables ultra-fast, low-latency sensing even in dynamic environments,
and represents a fundamental architectural departure from prior structured light approaches.
The method supports flexible structured light pattern design, enabling adaptation to various
sensing conditions without modifying the core processing architecture.
We validate our approach through a series of experiments that compare depth and color reconstruction
performance against both event-based and traditional methods (see Section~\ref{experiments}).

Various aspects of the main contribution are explained as follows:

\paragraph{High-Resolution and High-Speed 3D Reconstruction} 
There is always a trade-off between point cloud resolution and scanning process speed in visual-based 3D scanning systems, as discussed by~\cite{matsuda2015mc3d}. Time-of-Flight (ToF) cameras based 3D scanners can quickly report depth information but have limitations in fast RGB detection, according to~\cite{helios}. We obtained 3D reconstructions with variable resolution, allowing us to acquire a high-resolution colored 3D point cloud of an environment (in the order of millimeters, similar to 3D laser/ToF scanners but color data included), as well as high-speed 3D scanning (in the order of milliseconds with more sparse patterns).

\paragraph{Motion Adaptivity Across Static and Dynamic Scenes}
Unlike frame-based SL systems which struggle with motion blur or fail in static scenes, our method adapts to both fast and slow motions by tuning the SL pattern and the event cameras bias settings. This makes it suitable for mobile platforms without requiring a fixed scanning speed, as shown by~\cite{matsuda2015mc3d}. While we have not demonstrated the system on a mobile robot, this flexibility supports potential robotic integration.

\paragraph{Texture and Color Independence}
Stereo systems are sensitive to surface texture, as noted by~\cite{morar2020comprehensive}, but our method uses active structured light and performs depth estimation independently of object texture or surface color. Additionally, by projecting light in different wavelengths, our approach improves depth reconstruction even in challenging conditions such as when the surface color closely matches the illumination wavelength or reflects little light. While color detection may degrade in such cases, depth estimation remains robust compared to traditional passive methods.

\paragraph{Light and Power Efficiency}
Structured light approaches like Gray coding or phase-shifting, as discussed by~\cite{gupta2012micro}, are limited by projector brightness and camera dynamic range. Our system uses a high dynamic range sensor, as proposed by~\cite{brandli2014adaptive}, operates robustly in low-light or high-dynamic scenes, as noted by~\cite{muglikar2021esl} and ~\cite{morgenstern2023x}, and minimizes power consumption by controlling the light source in response to detected events. This enables efficient 3D sensing for mobile or embedded applications, as demonstrated by~\cite{wu2020exposure},~\cite{tin20163d}, and ~\cite{li2024learning}.

\paragraph{Fast, Frame-Free Depth Computation} 
Our method avoids accumulating events into full frames or spatio-temporal maps, instead performing per-event depth estimation via direct stereo disparity in the projector-camera space. This design supports real-time 3D point cloud reconstruction and avoids delay-inducing computations typical in high-speed scanning systems, as shown by~\cite{muglikar2021esl}.

\paragraph{Scalable and Flexible Pattern Design}
Our system can generate a wide range of projection patterns (e.g., coded dots, color sequences) to support specific sensing tasks. This flexibility enables adaptation to different resolution, speed, or semantic needs, and does not depend on specialized hardware. While high-speed coded patterns may require custom projection hardware, our framework is designed to accommodate these as optional extensions.

The rest of the paper is structured as follows: Section~\ref{mono2rgb} discusses color detection methods; Section~\ref{stereodepth} covers event-based triangulation-based depth measurement; Section~\ref{acls} describes our color and depth detection method; Section~\ref{experiments} details our results under various conditions; and Section~\ref{conclusion} outlines concluding remarks and future work.

\section{Monochrome to Color}\label{mono2rgb}
Color information is essential for tasks like segmentation and recognition, as noted by~\cite{tremeau2008color}.
Colorization involves creating a color image from a monochrome sensor or grayscale image without sacrificing resolution.
This process relies on external data about the image's colors obtained from external device, as described by~\cite{jang2020deep}, user input, as shown by~\cite{levin2004colorization}, and a trained neural network that incorporates the scene's color information, as demonstrated by~\cite{zhang2016colorful} and~\cite{cohen2024colorful}. It can be both time-consuming and costly.

The first generation of event cameras (ECs) were monochrome, with color ECs only recently becoming available, as introduced by~\cite{li2015design},~\cite{moeys2017sensitive}, and~\cite{moeys2017color}. However, color ECs have a lower resolution than monochrome ECs due to sensor size limitations and the need for color filters, as explained by~\cite{bayer1976color},~\cite{khashabi2014joint}, and~\cite{ramanath2005color}.

To maintain resolution while tripling the bandwidth requirements,
~\cite{marcireau2018event}, utilized dichroic filters on three ECs. They combined the output of three ECs, enabling the capture of color information in three distinct event streams.

\begin{figure}[htbp]
	\centering
	{\includegraphics[width=\linewidth]{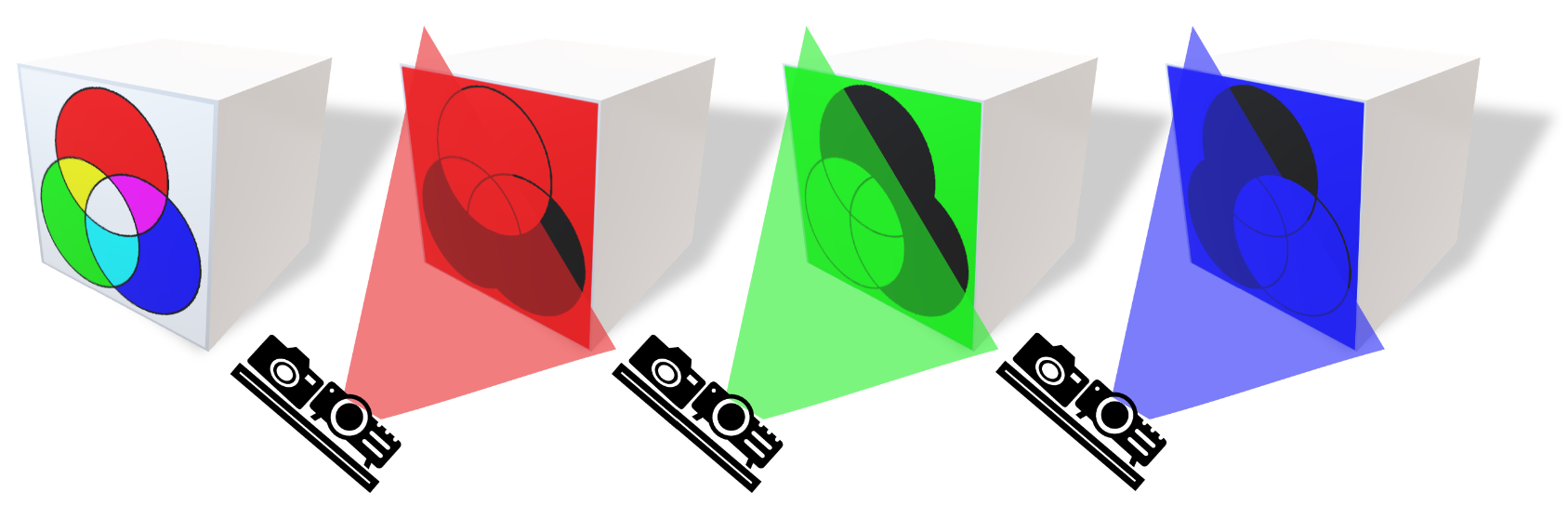}}
	\\
	{\includegraphics[width=0.19\linewidth]{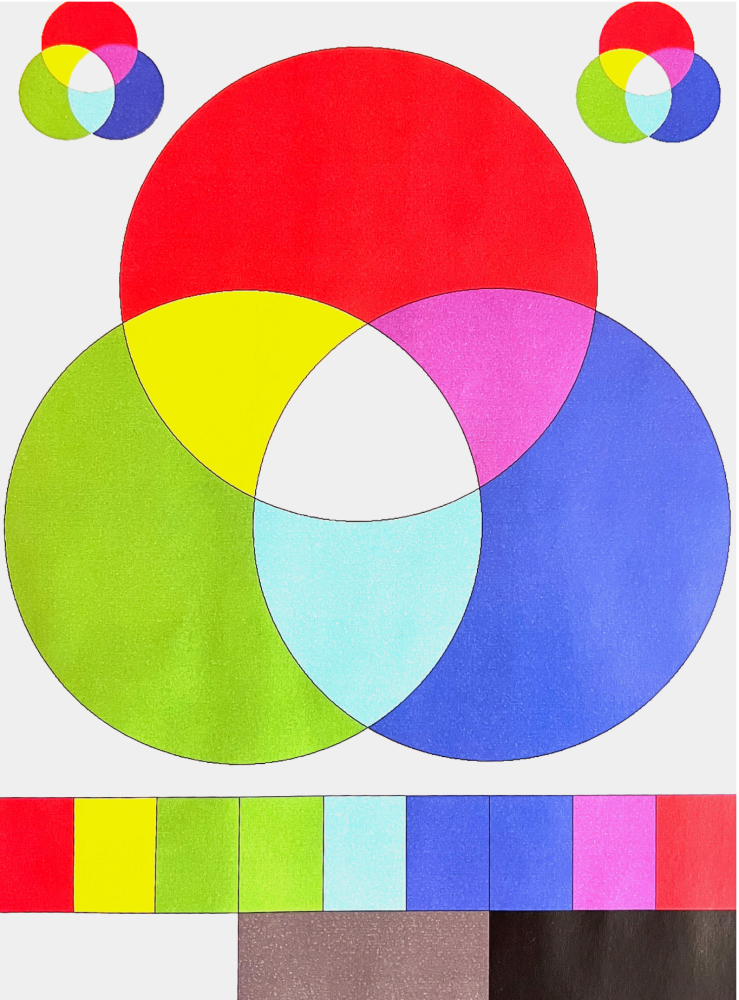}}
	{\includegraphics[width=0.19\linewidth]{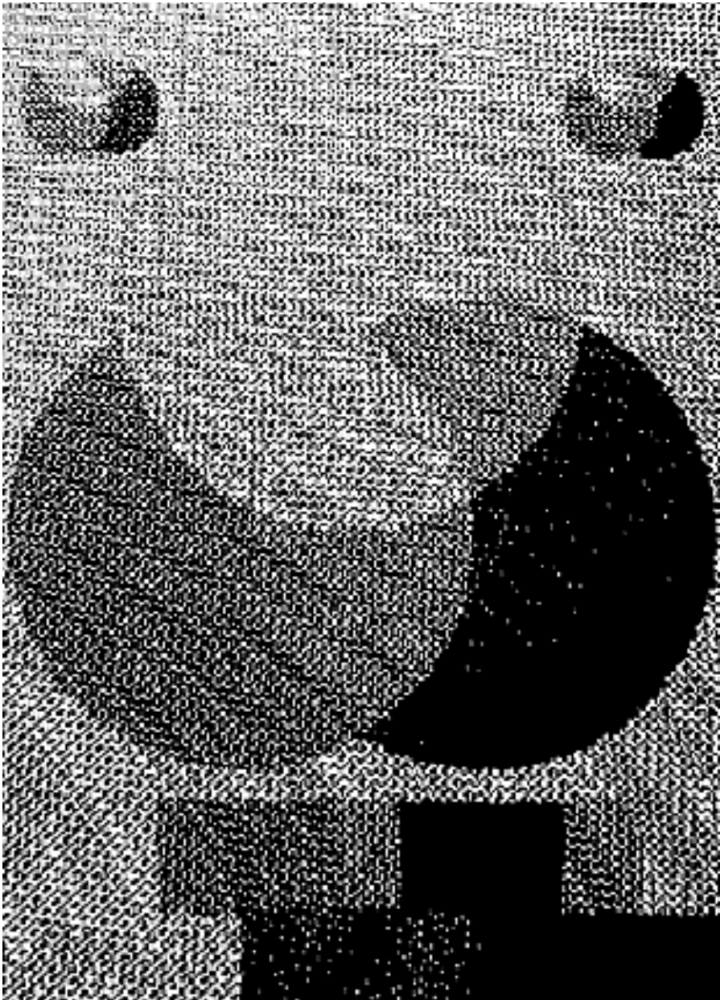}}
	{\includegraphics[width=0.19\linewidth]{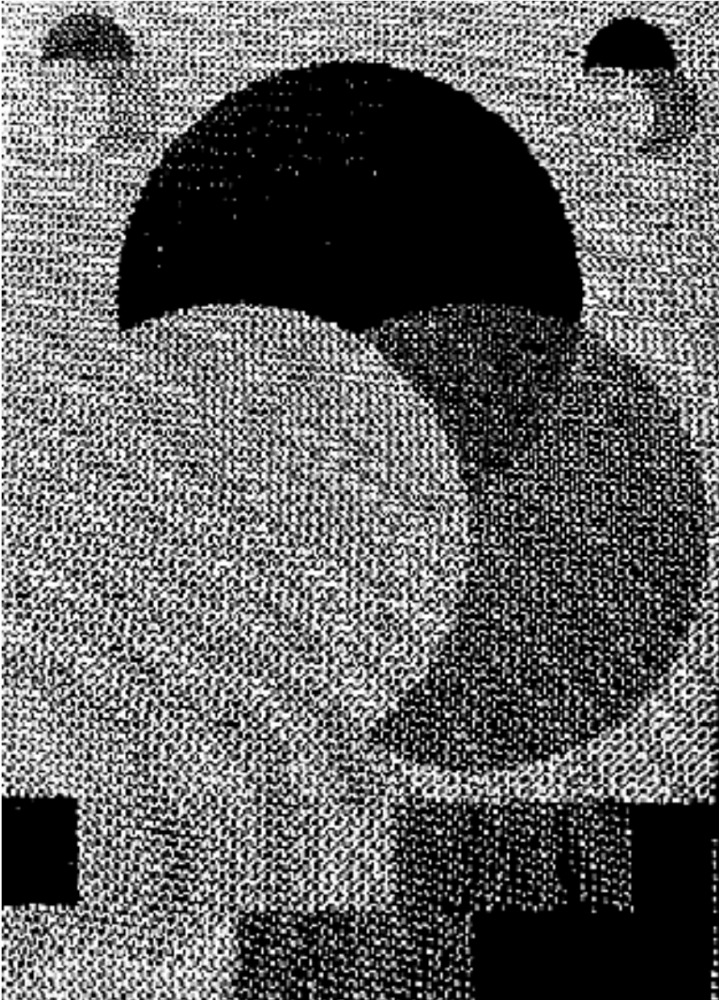}}
	{\includegraphics[width=0.19\linewidth]{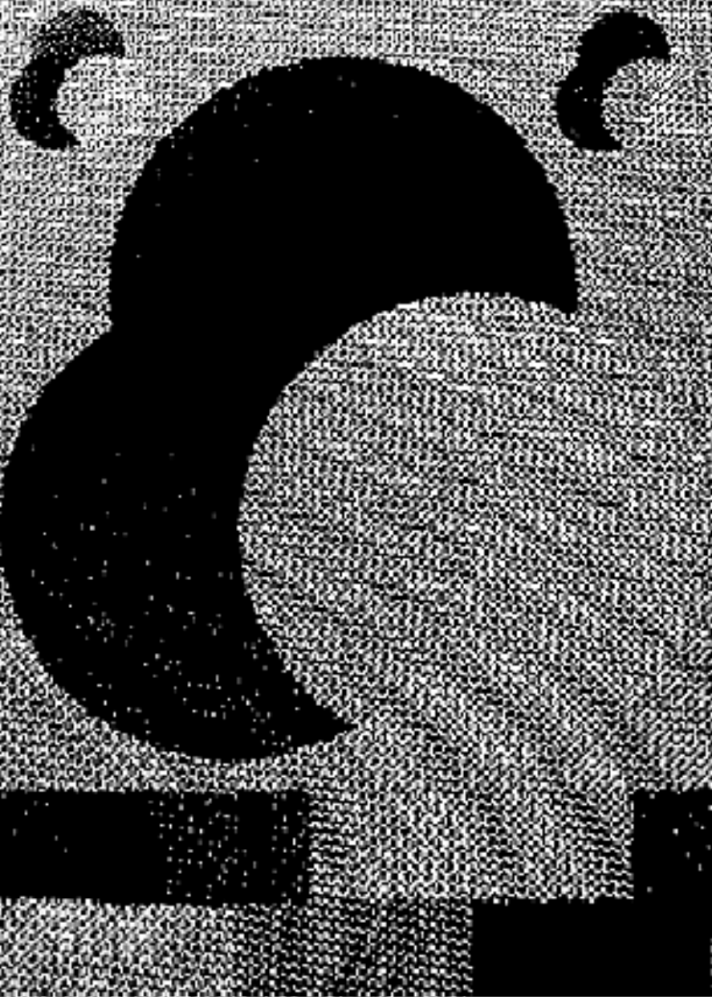}}
	{\includegraphics[width=0.19\linewidth]{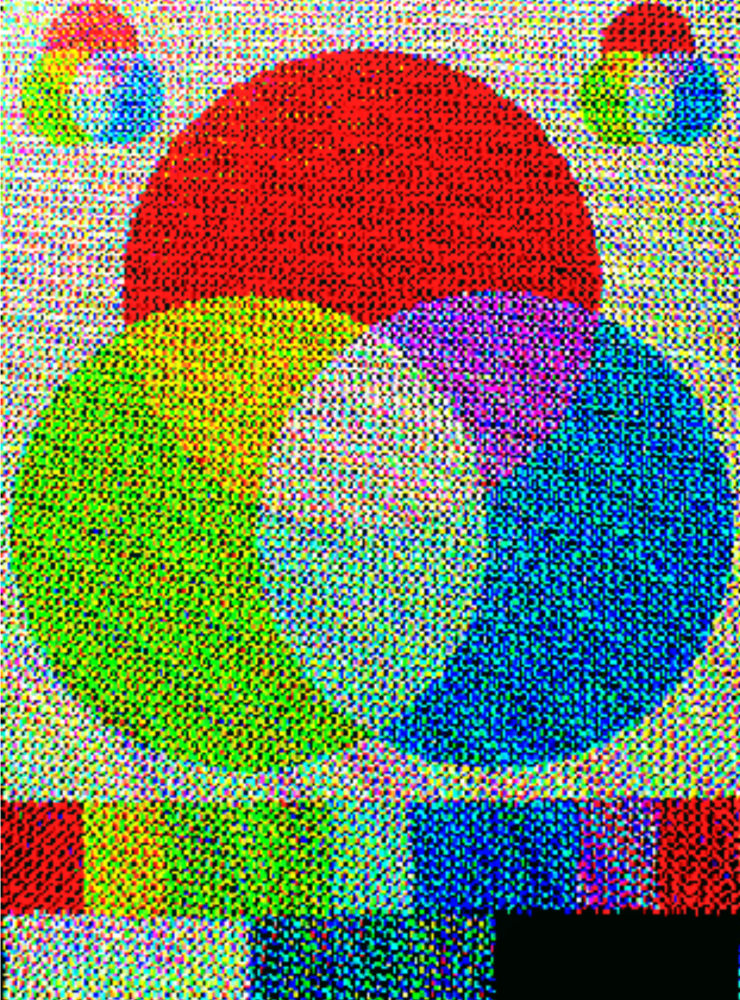}}
	\caption{Color detection of a printed color wheel, as introduced by~\cite{Bajestani_2023_WACV}. Top: The proposed procedure involves projecting various light patterns in different wavelengths/colors. Bottom, from left to right: high-resolution Ground Truth (GT), reconstructed images captured by a VGA monochrome EC in the channels of Red, Green, Blue, and the fully reconstructed image.}
	\label{fig:mono2rgb}
\end{figure}

In our initial work by~\cite{Bajestani_2023_WACV}, we utilized a DLP projector to emit light patterns, referred to as ASL, onto a scene. The EC then captured the reflections of these patterns, generating events that were tagged with the scene's color information. As Fig.~\ref{fig:mono2rgb} shows the proposed procedure, we are able to reconstruct full color image from a monochrome EC. Section~\ref{acls} describes our method for color and depth detection in details in terms of pattern frequency and coverage. Some visible noise patterns in the reconstructed color images (e.g., Fig.~\ref{fig:mono2rgb}) are due to optical interactions between the camera and projector lenses. These patterns can appear due to lens distortions or interference, particularly when using high-intensity structured projection. This factor will be addressed in future hardware refinements.

\section{Depth Sensing via Triangulation}\label{stereodepth}
In order to gather more information about the scene, fusion methods with an additional sensor are considered. While the interference-based methods are known as extremely accurate for micro-scale measuring, Time-of-flight methods are well known as low-accuracy measuring methods for a large-scale scene. Triangulation-based methods would be in the middle of them, as described by~\cite{marrugo2020state}. Triangulation-based techniques, including stereo vision and SL, have been demonstrated to provide precise depth information at short distances. In this paper, we focused on event-based depth measuring via SL.

\subsection{Event-based depth sensing with SL}
ECs can identify SL related events due to their frequency or contrast changes. By changing the frequency of SL, the camera can detect points individually. It required to be expressed that there are two different fusions of SL and EC. The first type is when a SL device is used to \emph{create} events and capturing those events by the camera (same as the proposed method by~\cite{martel2018active}). The second method is to use a SL device (or any other 3D depth sensors) to obtain a depth pre-reconstruction of the scene and subsequently adding the depth information to the captured events. For instance,~\cite{weikersdorfer2014event} used an external SL projector (Kinect) to receive the depth information of the scene. In parallel, an EC is used to obtain events (the brightness changes of each pixel due to movements), and the two outputs are merged to construct the 3D events. Therefore, despite having an active external lighting source, movement is required to generate the necessary data. Using the Kinect separately in parallel is power-consuming, and the process time would not be shortened.

In general, SL can be considered as an active stereo-vision method which uses the same concept (triangulation) for depth measurement. However, in this case, a pattern is emitted on the surface and the camera will capture the pattern deformation. Consequently, the structured pattern replaces one of the cameras in stereo-camera system. A typical SL system uses one camera and one projector; however, to reach a higher resolution or a faster full measurement, more cameras can be added as shown by~\cite{willomitzer2017single} and ~\cite{zhao2019miniature}.

\begin{table*}[!t]
	\begin{center}
		\caption{Summary of Previous SL-based Systems Addressing Depth Estimation with monocular EC}
		\label{tab:related}
		\resizebox{\linewidth}{!}{%
			\begin{threeparttable}
				\begin{tabular}{llll}
					\toprule
					\textbf{Method}	& \textbf{EC}	& \textbf{Sensor} & \textbf{Projector}\\
					\midrule
					\cite{brandli2014adaptive} & DVS128 & \mbox{$128\times128$} & Laser line 500 Hz \\
					MC3D ~\cite{matsuda2015mc3d} & DVS128 & \mbox{$128\times128$} & Laser point 60 fps\\
					FTD ~\cite{leroux2018event} & ATIS0 & \mbox{$304\times240$} & DLP TI LightCrafter 3000\\
					FPP ~\cite{mangalore2020neuromorphic,li2024event} & DAVIS346 & \mbox{$346\times260$} & DLP TI LightCrafter 4500\\
					ESL ~\cite{muglikar2021esl} & Prophesee Gen3 & \mbox{$640\times480$} & Laser point 60 fps\\
					X-maps ~\cite{morgenstern2023x} & Prophesee Gen3 & \mbox{$640\times480$} &  Laser point 60 fps\\
					\multirow{2}{*}{SEG ~\cite{lu2024sge}} & \multirow{2}{*}{Prophesee Gen4} & \multirow{2}{*}{\mbox{$1280\times720$}} & Laser point 60 fps (for static scene) \\
					&  &  & DLP projector OPR305185 (for dynamic scene)\\
					\textbf{Ours} E-RGB-D & Prophesee Gen3 & \mbox{$640\times480$} & DLP TI LightCrafter 4500\\
					\bottomrule
				\end{tabular}
			\end{threeparttable}
		}
	\end{center}
\end{table*}

The concept of using an external projector to employ SL and prevent stereo challenges with a monocular EC is foundational to modern systems employing event cameras, as shown by~\cite{brandli2014adaptive},~\cite{matsuda2015mc3d},~\cite{muglikar2021esl}, and~\cite{morgenstern2023x}. Table~\ref{tab:related} summarizes previous structured light (SL) systems that have tackled the issue of depth estimation using event cameras. The SL methods are different in terms of patterns types.

\textbf{Coded patterns SL:}\label{sec:rel_EBSL}
Strips SL are introduced to perform the area scanning. If they are unique, the system can identify points and calculate the depth. If these strips are in black and white (\emph{binary coding}), series of patterns are needed to identify the corresponding points. Thus compared to the other patterns, binary patterns are sensitive to object movement. Consequently, a high frequency of switching the binary patterns is assisting. DLP projectors have the ability to switch patterns in the order of kilohertz, as demonstrated by~\cite{wang2022enhancing} and \cite{lu2024sge}. To have a higher resolution, 2D and hybrid patterns are introduced by~\cite{yu2020high}.

\cite{leroux2018event} present a 3D reconstruction method using an Asynchronous Time-based Image Sensor (ATIS) with \mbox{$304\times240$} pixels and a DLP projector. Instead of line structured light (SL), Frequency-Tagged Dots (FTD) are projected onto the scene. The known pattern and distances allow calculation of pattern deformation and point depth. This method, however, is highly sensitive to movement and ambient light changes, which affect the number of captured events.

\cite{mangalore2020neuromorphic} and 
\cite{li2024event} used a Dynamic Active-pixel Vision Sensor (DAVIS346, \mbox{$346\times260$} pixels) with a DLP LightCrafter 4500 for 3D reconstruction. They developed a Fringe Projection Profilometry (FPP) system using a moving fringe pattern, allowing the EC to scan multiple lines simultaneously, which is faster than line-scanning methods. The EC's advantage is detecting shadowed areas, unlike frame-based cameras where shadows and dark regions appear the same. However, a pre-recording of the scene without the object is needed to ``inpaint'' shadowed areas, and the camera's limited event reporting capacity can lead to some events being eliminated.

\textbf{Simple patterns SL:}\label{sec:rel_Laser}
The simplest method, known as the statistical pattern, involves a random distribution of dots. This method is used in various commercial devices such as
\cite{kinectv1},
\cite{intel3d}, and
\cite{orbbec3d} due to its simplicity and small footprints. 
\cite{huang2021high}, also frequently projected a single pseudo-random pattern using a DLP6500 projector. They generated event frames from the event stream and utilized a digital image correlation method to calculate displacements and derive 3D surfaces of target objects. While using a single dot-based pattern increased scanning speed, it sacrificed detailed information compared to dense information achievable with line-based patterns. Moreover, implementing a discrete pattern led to inaccurate dot location and sensitivity to ambient light, resulting in low-resolution 3D depth measurements, as reported by~\cite{marrugo2020state}.

Another way to acquire higher resolution is to use line instead of discrete dots. In that way, high accuracy measurement in one direction will be attained. This method is also combined with a line laser for short range scanning. However, to measure the depth in all directions the line direction needs to vary.

\cite{brandli2014adaptive} used a laser line and an EC to scan the surface of an object. Using a concentrated light line (laser) helped achieve more contrast and detect the line more easily with the EC. Additionally, if the environment is static, emitting the structured light (SL) helps detect only the relevant pixels. However, to obtain a 3D reconstruction of the scene using this method, it is necessary to change the line direction or move the object in front of the laser line. 

By using the line laser and the EC, 
\cite{matsuda2015mc3d}, who introduced ``Motion Contrast 3D Scanning'' (MC3D), resolved the speed-resolution trade-off issue present in traditional SL scanners. Traditional SL scanners become inoperative when illumination changes occur in the environment; however, using a high dynamic range EC yields an improved final result. The laser scanner had an exposure time of 28.5 seconds, but their proposed device had an exposure time of one second.

Expanding on the Matsuda et al's work, 
\cite{muglikar2021esl}, who introduced ``Event-based Structured Light'' (ESL), employed time maps to establish a temporal link between the projector and camera. Initially, they produced depth maps by conducting an epipolar disparity search within rectified projector time maps. Following this, an additional processing stage was implemented to enhance pixel-level coherence and reduce event fluctuations. However, this stage demands significant computational resources, preventing their method from achieving real-time performance.

\cite{morgenstern2023x}, introduced a method that converts the projector time map into a rectified ``X-map'', capturing X-axis correspondences for incoming events and enabling direct disparity lookup without additional search. This method supported real-time interactivity, making it suitable for Spatial Augmented Reality (SAR) experiences requiring low latency and high frame rates. They claimed that their method is 7 to 100 times faster than considering the entire frame, as in the ESL method, because there is no need to do a row-by-row disparity search and calculate the depth for the whole frame. We used a partially similar X-mapping method to calculate the depth for each pixel. A detailed description is provided in the next section.

\begin{figure}[htbp]
	\centering
	\includegraphics[width=0.19\linewidth]{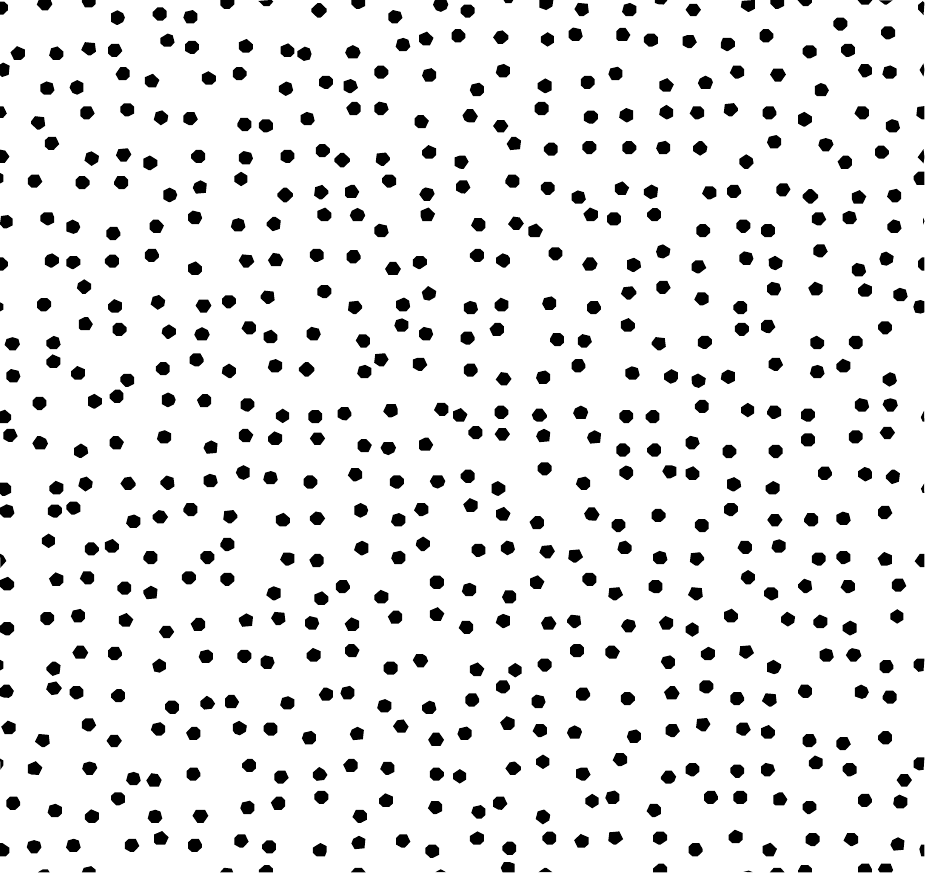}
	\includegraphics[width=0.19\linewidth]{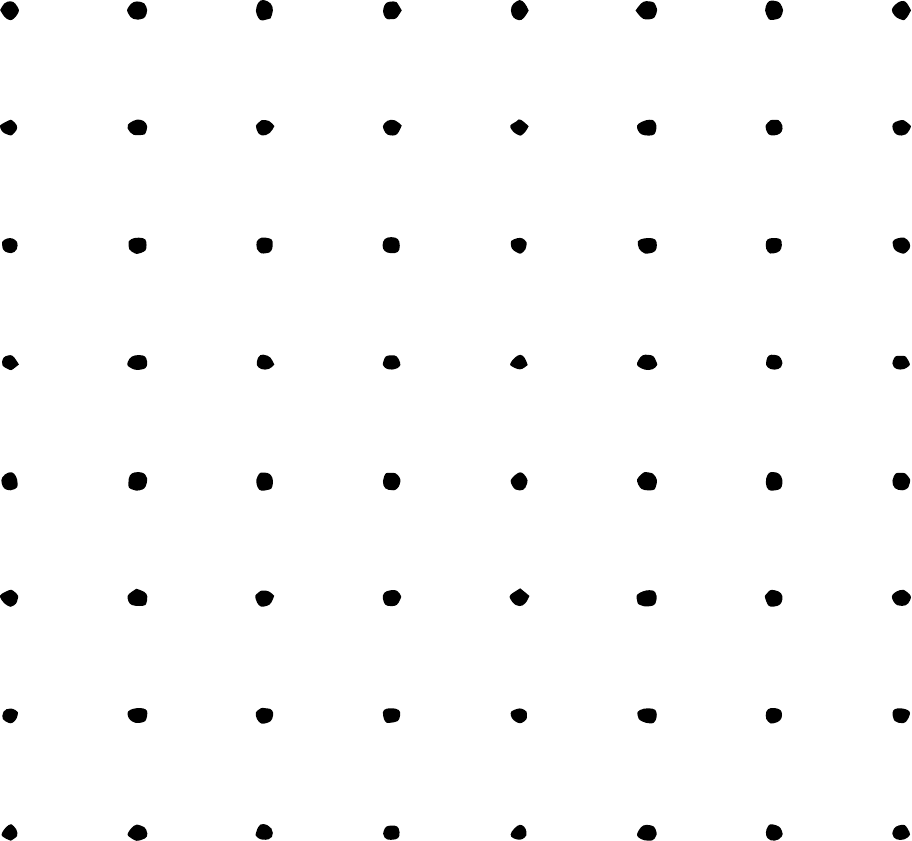}
	\includegraphics[width=0.19\linewidth]{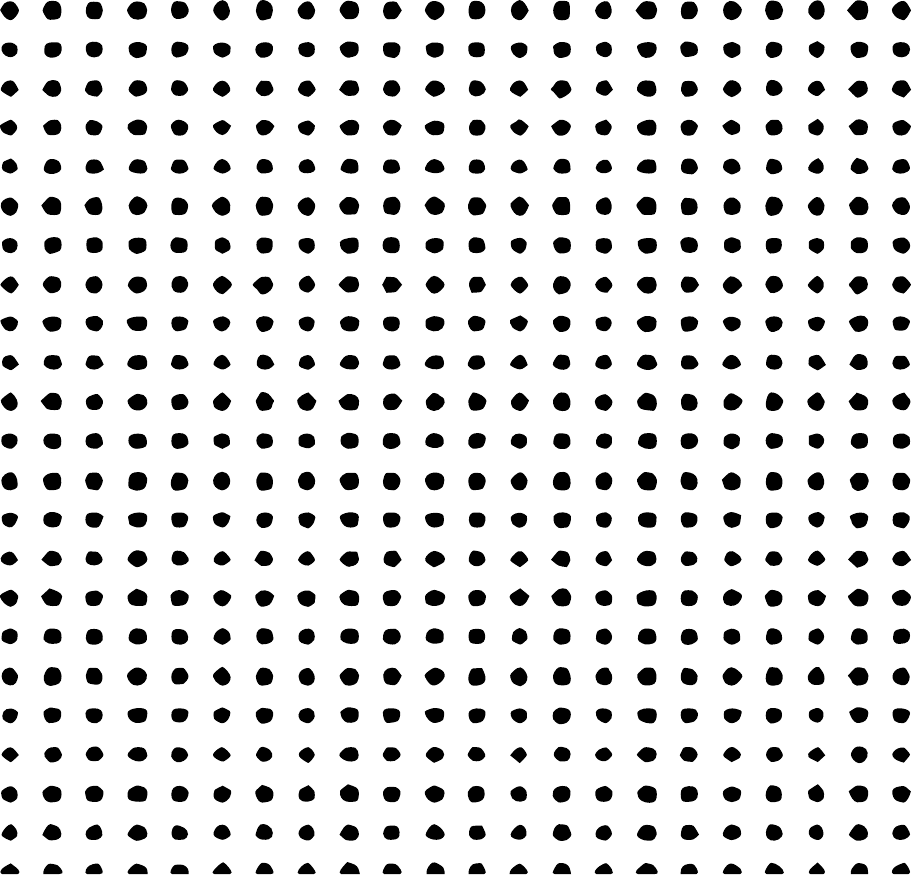}
	\includegraphics[width=0.19\linewidth]{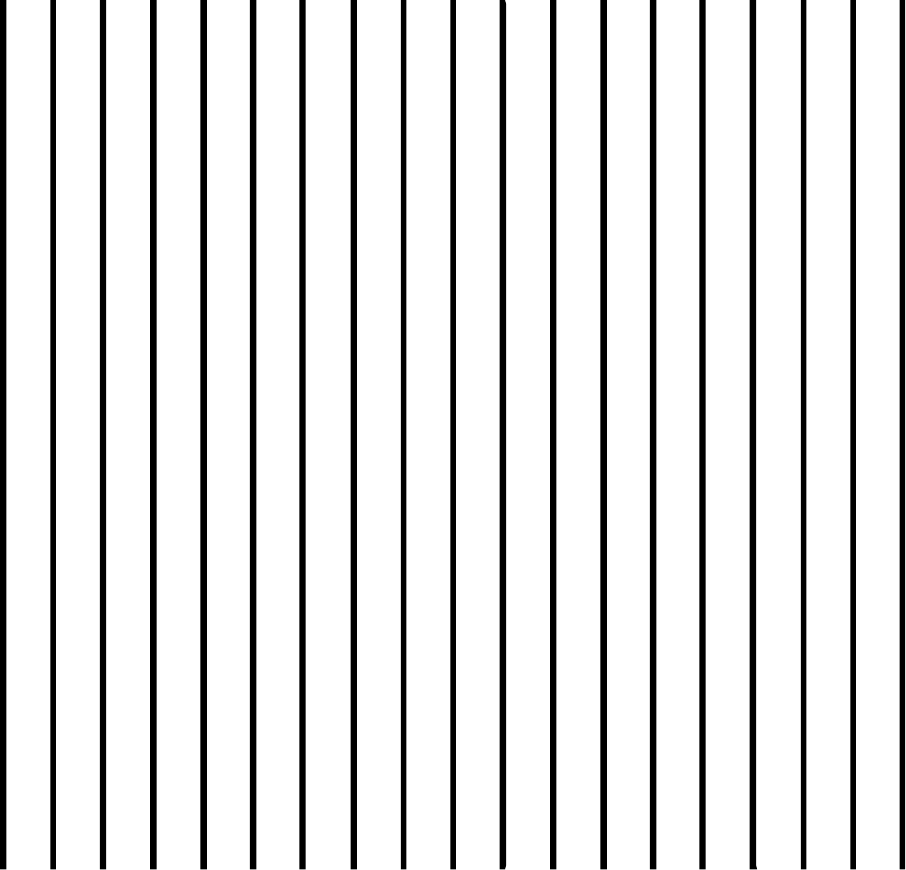}
	\includegraphics[width=0.19\linewidth]{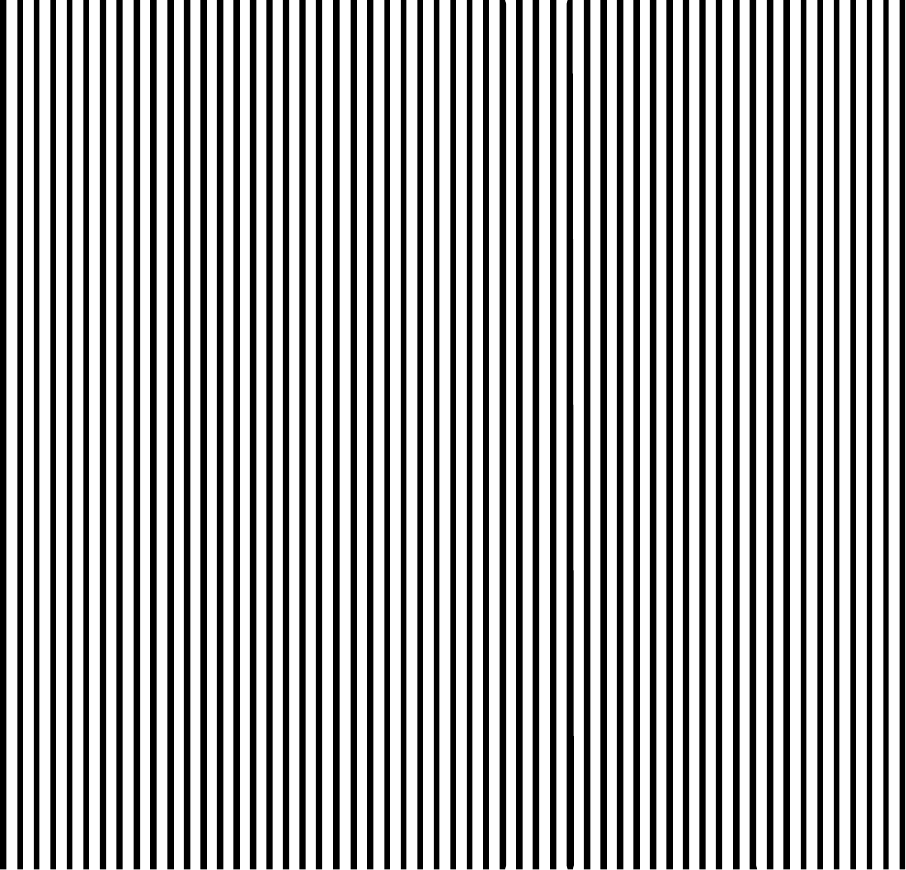}
	\vspace{1pt}
	\begin{tabular}{p{0.17\linewidth} p{0.14\linewidth} p{0.14\linewidth} p{0.14\linewidth} p{0.15\linewidth}}
		Random  & N dots & M dots & N lines & M lines\\
	\end{tabular}
	\caption{Proposed ASL pattern types for balancing speed and detail in E-RGB-D scanning. M is greater than N, meaning more points are reconstructed and have higher Coverage Percentage (CP). We did not use a pseudo-random dot pattern to detect the depth, but it could be used to add color to methods that detect only depth, such as those proposed by~\cite{huang2021high}.}
	\label{fig:patterns}
\end{figure}

\section{ASL: Adaptive Structured Light}\label{acls}
This section introduces our adaptive structured light framework. The proposed method is to adjust the SL pattern to balance the trade-off between
scanning speed and detail. Achieving a denser output requires more pixel data,
but there is a limit to the number of events an EC can process simultaneously,
as it may become bus-saturated. The percentage of Ground Truth (GT) points
estimated by the proposed method, relative to the total number of pixels in the
GT that contain data, is referred to as Fill Rate (FR), completeness, or depth
map completion, as defined by~\cite{muglikar2021esl}. However, to avoid reaching the camera's
bus-saturation limits, we control the number of projected points by changing
pattern. We refer to the ratio of ON to OFF pixels in a pattern as the Coverage
Percentage (CP), as defined by~\cite{Bajestani_2023_WACV}. Each pattern type has a different
CP, and each experiment is evaluated based on its FR.

\begin{figure}[htbp]
	\centering
	\renewcommand{\arraystretch}{1.2}
	\textbf{Mode One:} Color only\\
	{
		\begin{tabularx}{0.4\linewidth}{|*{4}{>{\centering\arraybackslash}X|}}
			\hline
			\cellcolor{gray!30}ID & \cellcolor{red!30}\( R \) & \cellcolor{green!30}\( G \) & \cellcolor{blue!30}\( B \) \\
			\hline
			\cellcolor{gray!30}250 &\cellcolor{red!30} 235 &\cellcolor{green!30} 235 &\cellcolor{blue!30} 235 \\
			\hline
		\end{tabularx}
	}
	\vspace{10pt}\\
	\textbf{Mode Two:} Depth only
	\\
	{
		\begin{tabularx}{0.4\linewidth}{|*{4}{>{\centering\arraybackslash}X|}}
			\hline
			\cellcolor{gray!30}ID & \( D_1 \) & ... & \( D_n \) \\
			\hline
			\cellcolor{gray!30}260 & 235 & ... & 235 \\
			\hline
		\end{tabularx}
	}
	\vspace{10pt}\\
	\textbf{Mode Three:} Depth then Color
	\\
	{
		\begin{tabularx}{0.7\linewidth}{|*{7}{>{\centering\arraybackslash}X|}}
			\hline
			\cellcolor{gray!30}ID & \( D_1 \) & ... & \( D_n \) & \cellcolor{red!30}\( R \) & \cellcolor{green!30}\( G \) & \cellcolor{blue!30}\( B \) \\
			\hline
			\cellcolor{gray!30}270 &235 & ... & 235 &\cellcolor{red!30} 235 &\cellcolor{green!30} 235 &\cellcolor{blue!30} 235 \\
			\hline
		\end{tabularx}
	}
	\vspace{10pt}\\
	\textbf{Mode Four:} Depth and Color
	{
		\begin{tabularx}{\linewidth}{|*{10}{>{\centering\arraybackslash}X|}}
			\hline
			\cellcolor{gray!30}ID & \cellcolor{red!30}\( D_1 \) & \cellcolor{red!30}... & \cellcolor{red!30}\( D_n \) & \cellcolor{green!30}\( D_1 \) & \cellcolor{green!30}... & \cellcolor{green!30}\( D_n \) & \cellcolor{blue!30}\( D_1 \) & \cellcolor{blue!30}... & \cellcolor{blue!30}\( D_n \) \\
			\hline
			\cellcolor{gray!30}280 &\cellcolor{red!30}235 &\cellcolor{red!30} ... &\cellcolor{red!30} 235 &\cellcolor{green!30} 235 &\cellcolor{green!30} ... &\cellcolor{green!30} 235 &\cellcolor{blue!30} 235 &\cellcolor{blue!30} ... &\cellcolor{blue!30} 235 \\
			\hline
		\end{tabularx}
		\vspace{10pt}\\
	}
	\caption{Pattern sequence and their exposure times in microseconds. In our experiments, we used mode 3 with two different values for $n$ (23 and 45), and mode 4 with $n$=23. One ID for each pattern type would be enough, but we could have different IDs for each color or depth pattern mode. While this increases the total scanning time, it makes the system more robust and trackable.}
	\label{fig:sequence}
\end{figure}

The high-resolution method, which uses line-based scanning and is less sensitive
to ambient light, as described by~\cite{matsuda2015mc3d} and~\cite{gupta2013structured}, is given the
highest priority among the proposed patterns, while the dot-based pattern is
assigned the lowest priority. Figure~\ref{fig:patterns} represents the various
patterns that E-RGB-D can work with, including line-based scanning, dot-based
scanning, and pseudo-random dot patterns. For instance, the SL pattern could be
changed according to either the remaining battery charge or the robot's motion
speed. The robot can switch to the line method and decrease the LED current
(decrease the power of the projector) when the energy level reaches its critical
state. This flexibility is a key advantage of our approach. The system is not tied to any fixed pattern encoding scheme; rather, it can dynamically switch between different structured light patterns depending on environmental conditions (e.g., lighting, motion), application needs (e.g., speed vs. resolution), or system constraints (e.g., power consumption). These patterns are supported by the high-speed binary projection capabilities of the DLP LightCrafter 4500, and are fully compatible with our per-event processing pipeline. This modular design makes our framework broadly applicable and adaptable to various scanning contexts without requiring any structural changes to the method.

To reconstruct the color and depth, we introduced different pattern sequences, we will describe them separately in detail in the following subsections.

\subsection{Color detection}\label{color}
As mentioned in Section~\ref{mono2rgb} and shown in Figure~\ref{fig:mono2rgb}, we project each pattern three times onto the scene with different wavelengths/colors, followed by capturing the reflection with the EC. These patterns could be part of a depth measuring procedure (moving line or dots) or a single pattern just to detect the color. Figure~\ref{fig:sequence} shows the pattern sequence and their exposure times in microseconds. There are 4 different types of sequences to obtain color and/or depth. For instance, when measuring depth first and then color, any pattern from Figure~\ref{fig:patterns} could be used to capture the color, even if the depth pattern is different. So, based on the needs, one pattern could be denser and have a higher CP than the other. One could use a solid pattern and decrease the Region of Interest (ROI) of the camera to prevent bus saturation while still obtaining a fully dense, colorful image for a specific area.

\begin{figure}[htbp]
	\centering
	{\includegraphics[width=0.48\linewidth]{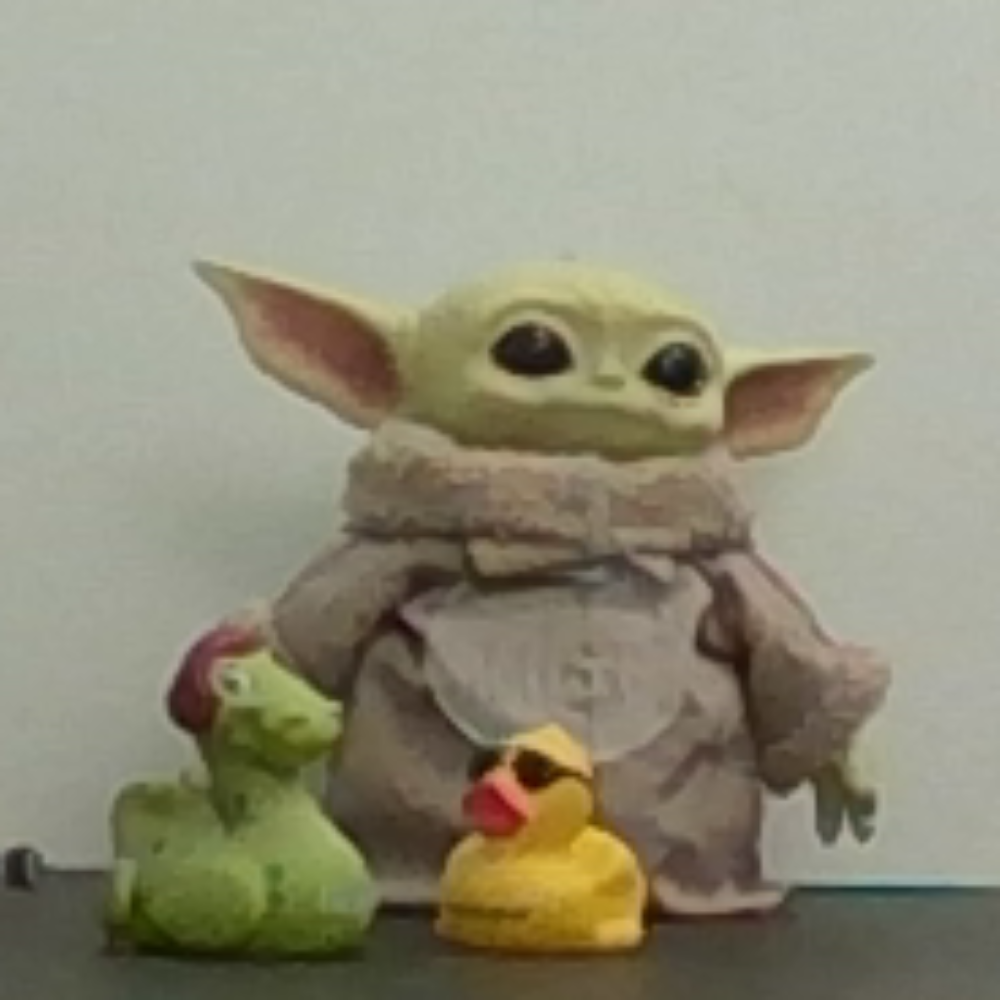}}
	{\includegraphics[width=0.48\linewidth]{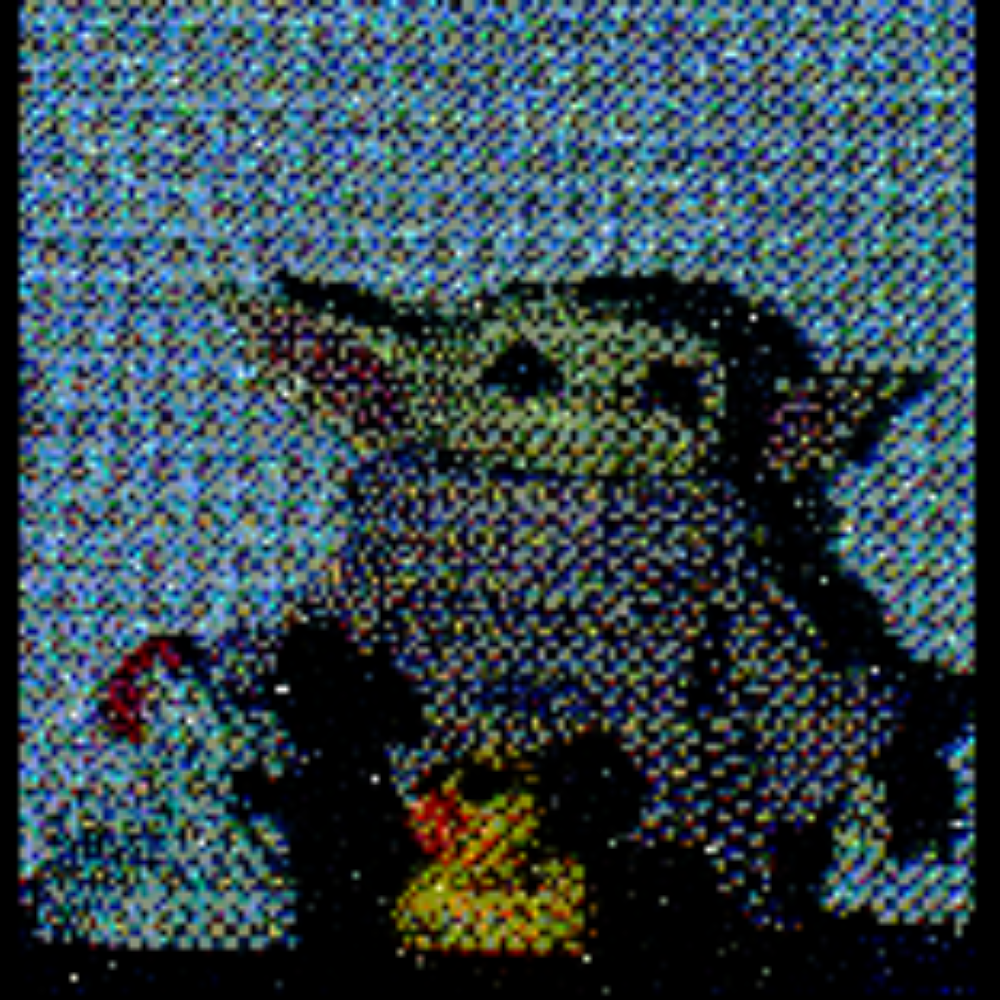}}
	\\
	{\includegraphics[width=0.32\linewidth]{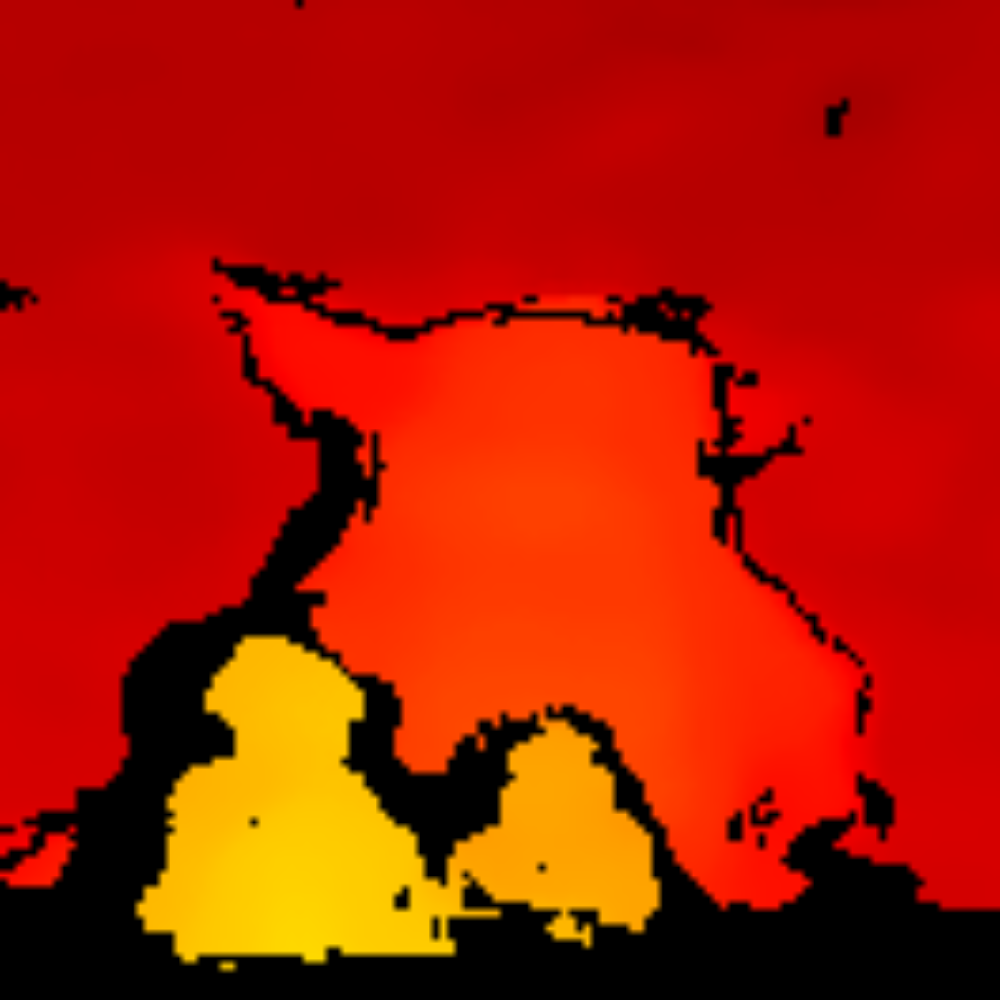}}
	{\includegraphics[width=0.32\linewidth]{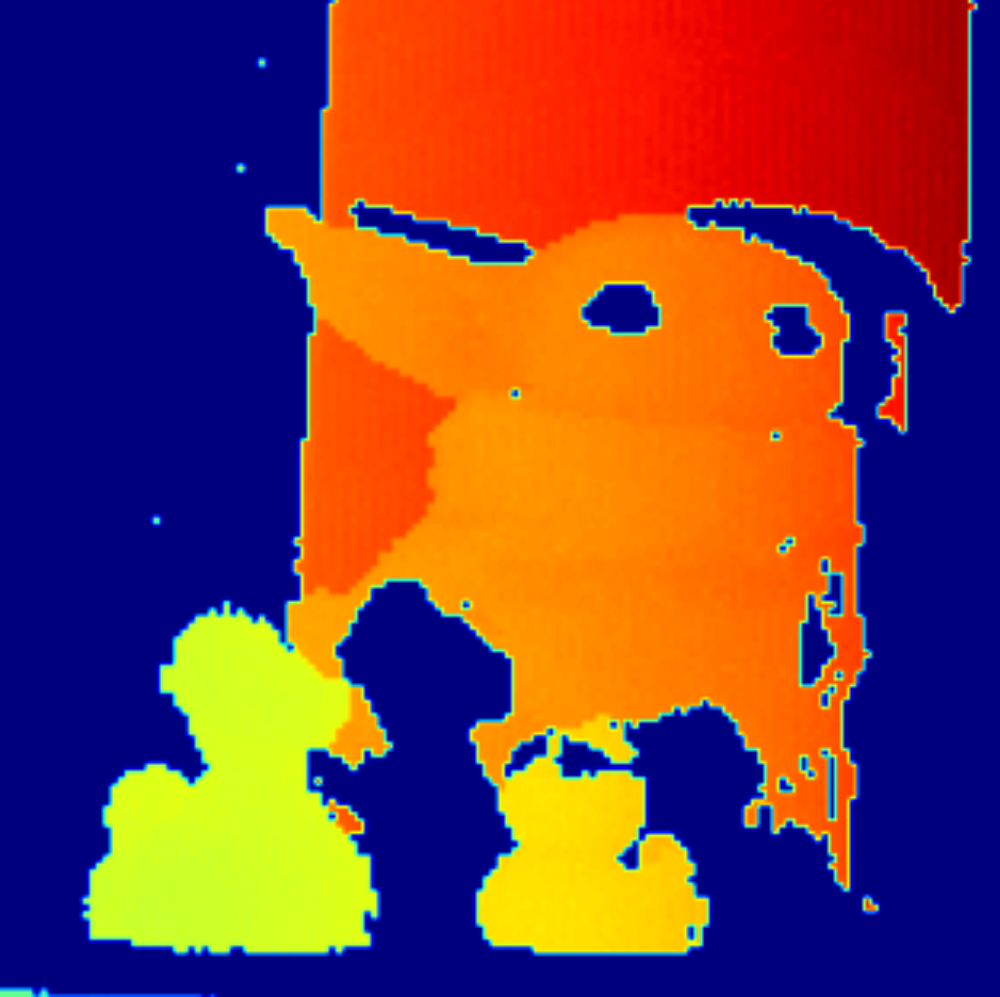}}
	{\includegraphics[width=0.32\linewidth]{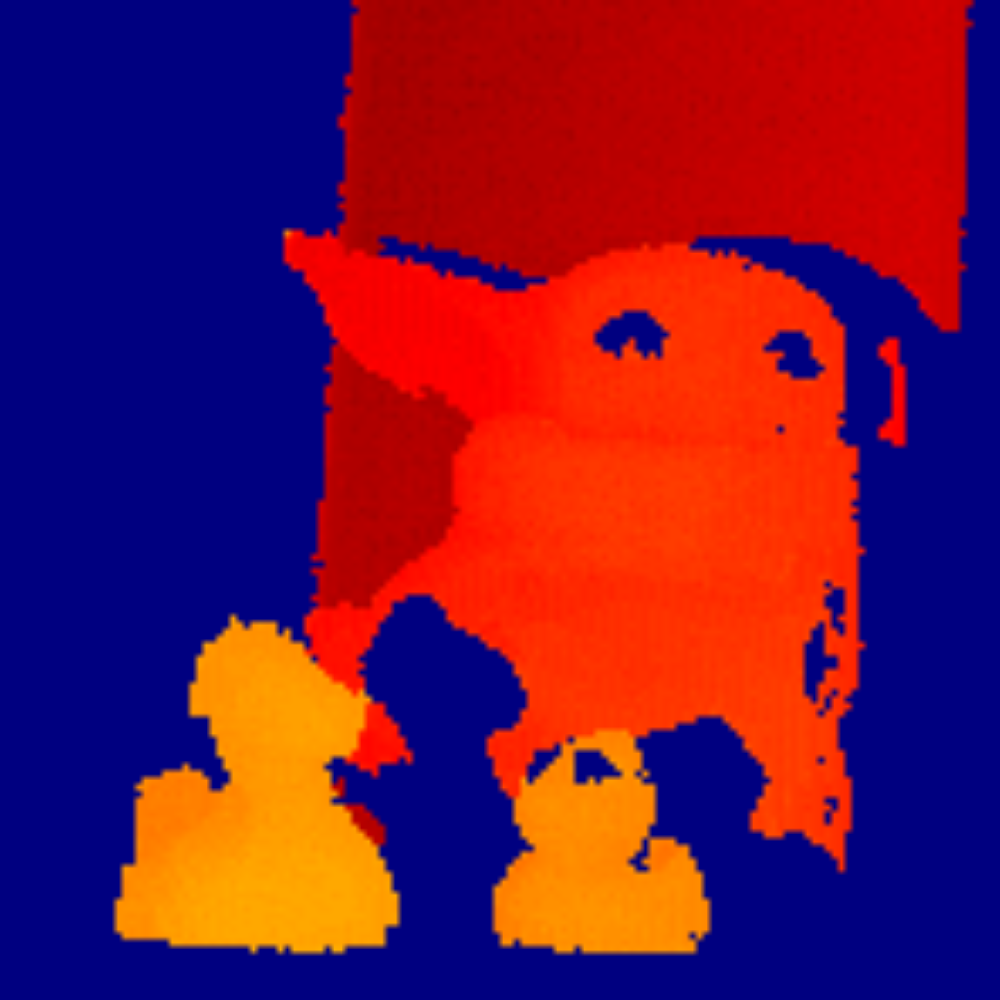}}
	\\
	{\includegraphics[width=0.32\linewidth]{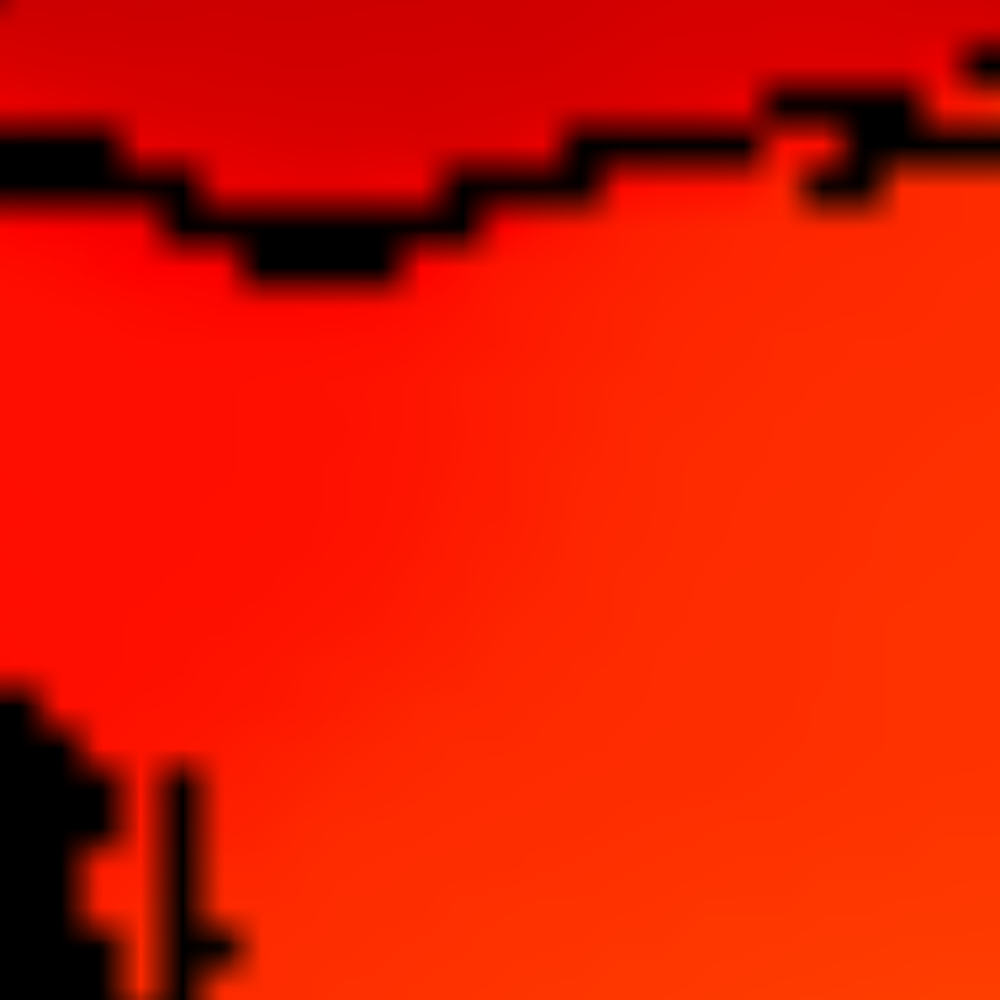}}
	{\includegraphics[width=0.32\linewidth]{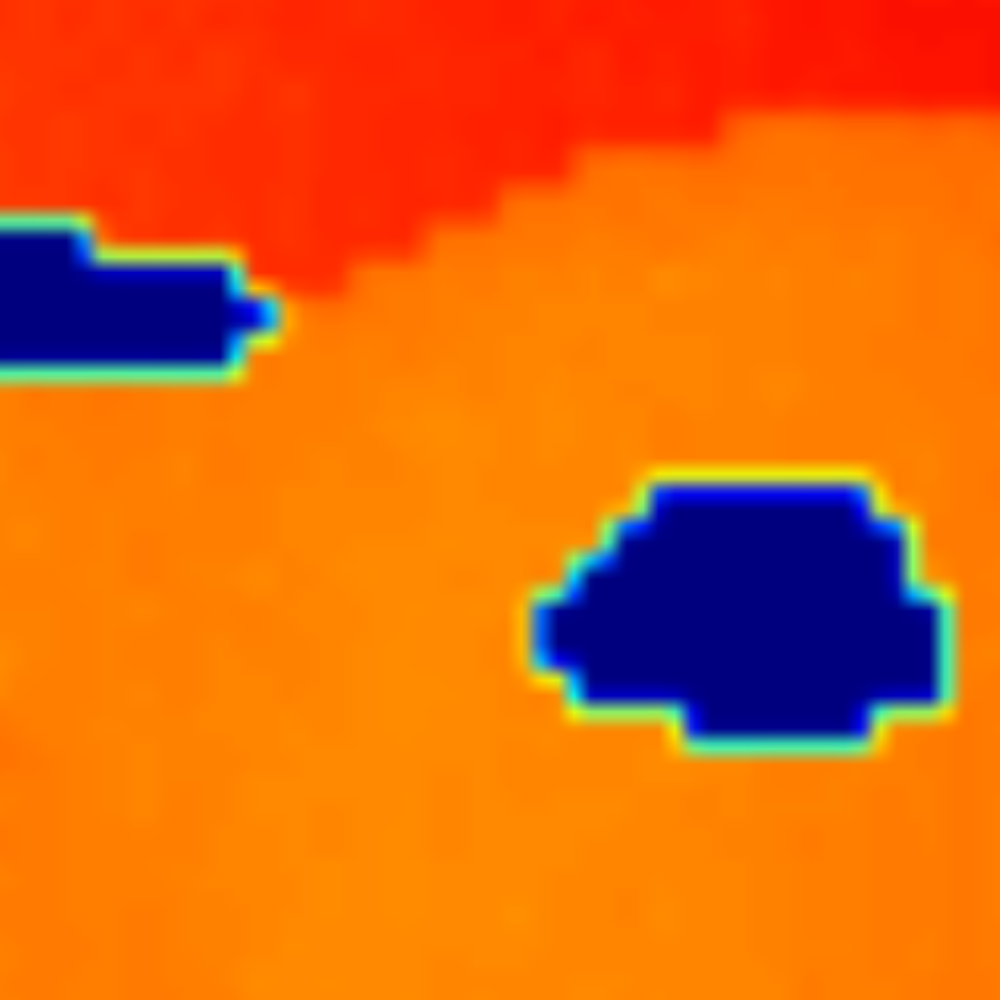}}
	{\includegraphics[width=0.32\linewidth]{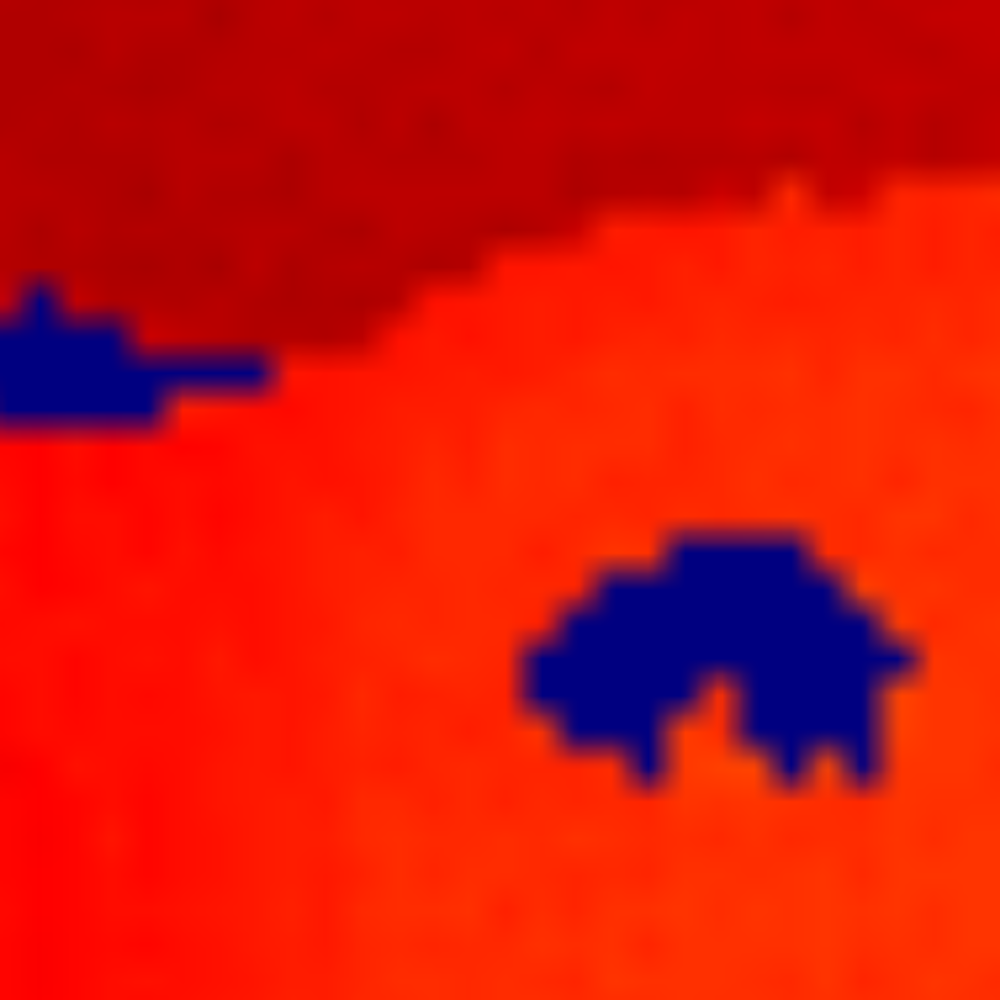}}
	\begin{tabular}{p{0.3\linewidth} p{0.25\linewidth} p{0.35\linewidth}}
		D455:~33~ms & ESL:~2.36~s & \bms{ours:~12.57~ms}\\
	\end{tabular}
	\caption{\textbf{Top Left:} Output of the D455 RGB camera.
		\textbf{Top Right:} The RGB image reconstructed by the proposed method with a Monochrome EC.
		\textbf{Middle Row:} Depth detection comparison between D455 (Left), \cite{muglikar2021esl}'s work named ESL (Middle), and our method, E-RGB-D (Right).
		\textbf{Bottom Row:} Zoomed-in view of the middle row. Note that the color differences are due to defining different minimum and maximum values for the jet-coded colorization; the actual difference in mm is detailed in Section~\ref{results}.}
	\label{fig:depth_doll}
\end{figure}

Since the projector and camera are connected through the external trigger pins, we projected a blank pattern with a specific exposure time as the ID, allowing the software to detect which pattern sequence mode is being projected by the projector. Both the ESL and X-maps methods utilize a Micro Electro-Mechanical System (MEMS) laser projector limited to 60~Hz, making them suitable for raster scanning patterns where rows are projected sequentially. In contrast, our work employs a Digital Micromirror Device (DMD projector) capable of projecting all pixels simultaneously, with a capacity exceeding 4~kHz. The top row of Figure~\ref{fig:depth_doll} shows the RGB reconstructed by our method compared to the output color image of the RealSense D455 camera.

We evaluate the quality of our color reconstruction using a Macbeth ColorChecker chart, following standard color accuracy protocols, as established by~\cite{bayer1976color}, \cite{khashabi2014joint}, and \cite{ramanath2005color}. As shown in our previous work (\cite{Bajestani_2023_WACV}), the method consistently reproduces color accurately, even though it uses only a monochrome event camera, supporting the effectiveness of the proposed active color projection approach.

\subsection{Depth detection}\label{depth}
Like X-maps, as proposed by~\cite{morgenstern2023x}, we provided a lookup table for disparity values to eliminate the need for disparity search. Because computing scene disparity by aligning time entries of the map along epipolar lines with an idealized projector time map is computationally intensive, as noted by~\cite{muglikar2021esl}. However, our DMD projector does not exhibit raster printing behavior, so we did not store the projector's $x$ coordinates in relation to $y$ and time $t$, and we do not use a temporal map either. Instead, we determine the disparity by knowing the column of the projected line. We will describe it in detail in this section.

\textbf{Direct disparity lookup table:} We formulate the problem of depth estimation using epipolar lines. After calibrating the system and setting up the stereo configuration, we create a lookup table $\text{LUT}_{c}$ that assigns each pixel on the camera plane $P_{c}(x_{c},y_{c})$ to its corresponding pixel on the rectified camera's image $P_{c_{r}}(x_{c_{r}},x_{c_{r}})$. Additionally, a lookup table $\text{LUT}_{p}$ assigns each pixel $P_{p}(x_{p},y_{p})$ on the projector plane to $P_{p_{r}}(x_{p_{r}},y_{p_{r}})$ on the rectified projector image.

\begin{equation}
	\text{LUT}_{c}(P_{c}(x_{c},y_{c})) = P_{c_{r}}(x_{c_{r}},y_{c_{r}})
\end{equation}

\begin{equation}
	\text{LUT}_{p}(P_{p}(x_{p},y_{p})) = P_{p_{r}}(x_{p_{r}},y_{p_{r}})
\end{equation}

Considering that we are projecting different numbers of lines on the object, we could have an array $Columns$ that carries the column number $x_{p}$ of each line on the projector plane. As shown in Figure~\ref{fig:sequence}, the size of this lookup array could vary from $1$ to $n$ based on the pattern mode.

\begin{equation}
	Columns = \begin{bmatrix} x_{p_{1}} & x_{p_{2}} & \hdots & x_{p_{n}} \end{bmatrix}
\end{equation}

By identifying which pattern is being projected and knowing the coordinates of the captured event, we can determine the disparity of the incoming event directly. Imagine at the time of reciving the $\text{Event}_{c}(x_{c},y_{c})$, we are projecting line number $m$; then, we can determine two points of this line on the projector rectified map as:

\begin{equation}
	\text{Top}_{p_{r}} = \text{LUT}_{p}(Columns[m], H) = (x_{T},y_{T})
\end{equation}
\begin{equation}
	\text{Bottom}_{p_{r}} = \text{LUT}_{p}(Columns[m], 0) = (x_{B},y_{B})
\end{equation}

where $H$ represents the height of the projector resolution.

Because our setup is a horizontal stereo, the epipolar lines in the rectified
images are horizontal and have the same y-coordinate. The corresponding
point is determined by calculating the intersection of the line that
passes through these two points and the horizontal line that passes through the
event's pixel on the rectified camera plane.

\begin{equation}
	\text{Event}_{c_{r}} = \text{LUT}_{c}(x_{c},y_{c}) = (x_{E},y_{E})
\end{equation}

\begin{equation}
	x_{p_{r}} = x_{T} + (y_{E} - y_{T}) \cdot slope
\end{equation}
where $\text{slope} = \frac{x_{B} - x_{T}}{y_{B} - y_{T}}$.
And the disparity whould be:
\begin{equation}
	\text{disparity} = x_{p_{r}} - x_{E}
\end{equation}

The depth \( Z \) of a point in a scene can be calculated using the formula:

\begin{equation}
	Z = \frac{f \cdot B}{\text{disparity}}
\end{equation}

where \( f \) is the focal length of the camera, \( B \) is the baseline.
Algorithm~\ref{alg:event_algorithm} provides a concise overview of the entire
process.

\begin{figure}[htbp]
	\centering
	{\includegraphics[width=0.45\linewidth]{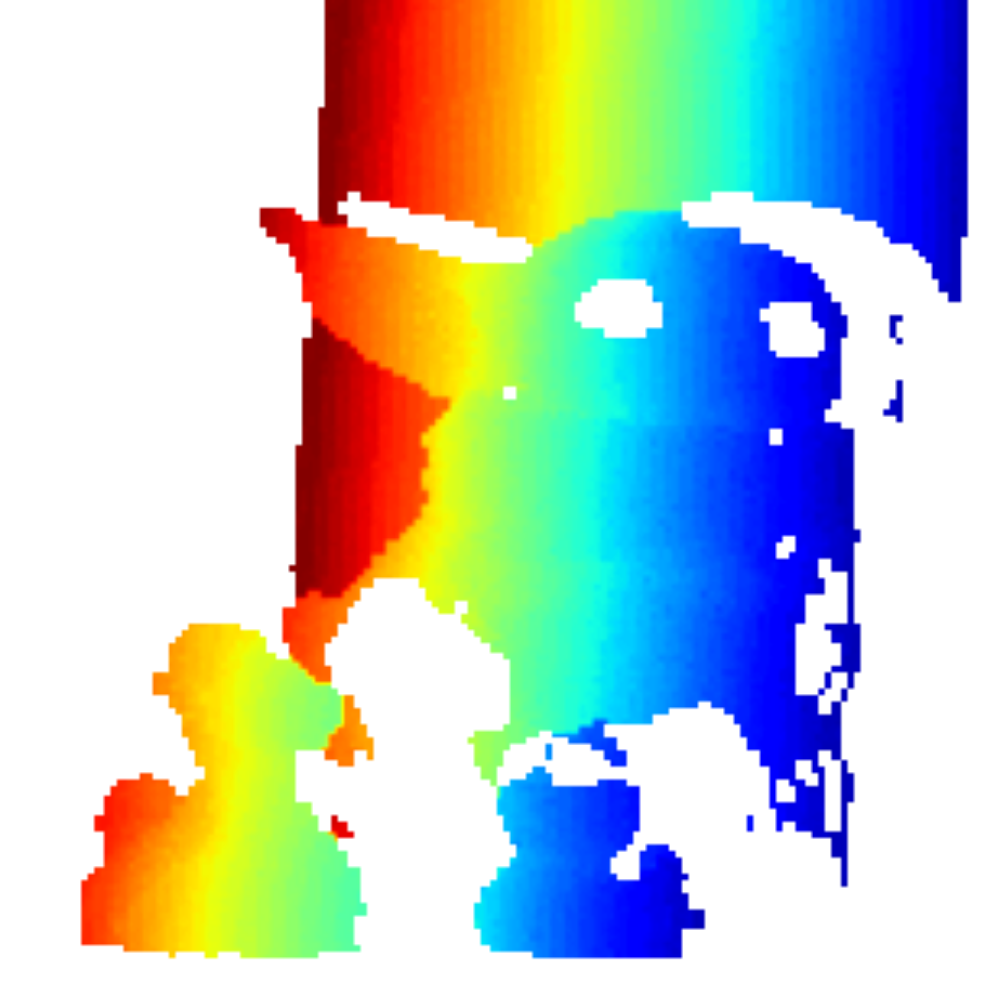}}
	{\includegraphics[width=0.45\linewidth]{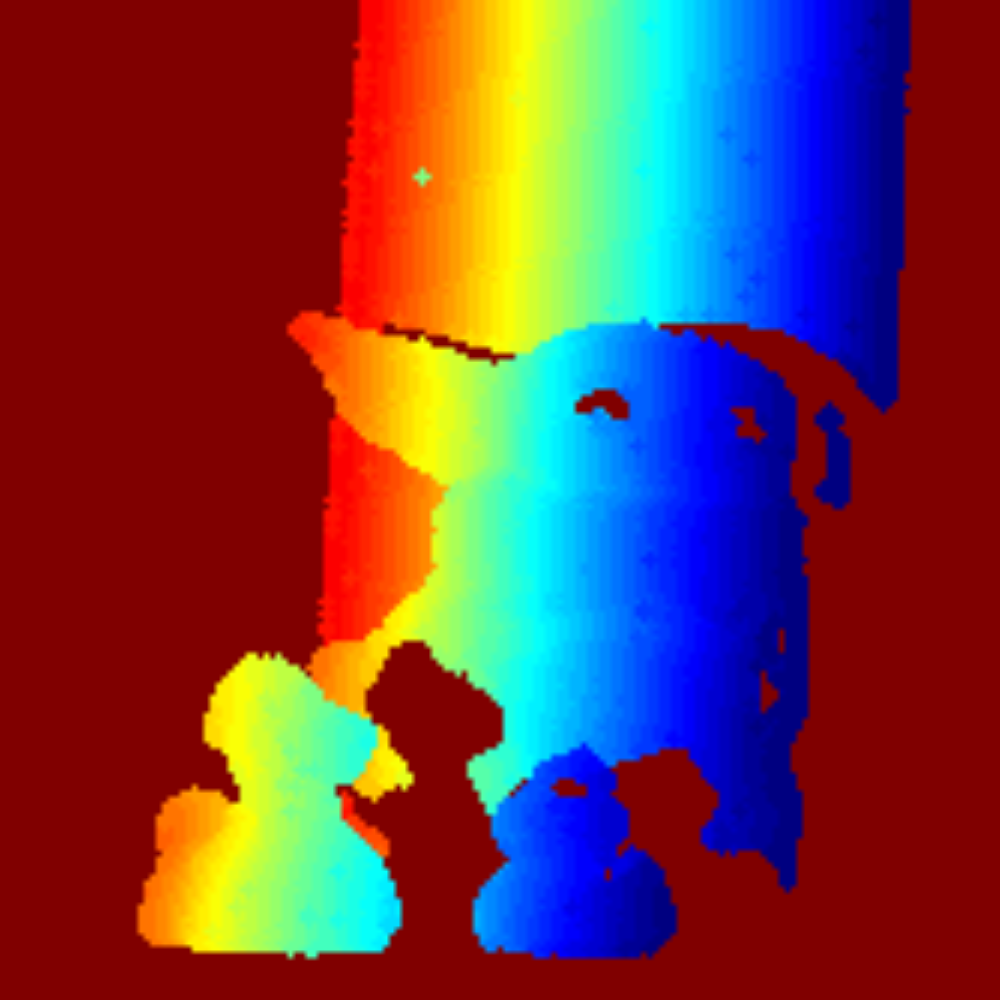}}
	\caption{Temporal map recunstructed by ESL (Left), ours E-RGB-D (Right). }
	\label{fig:temporal}
\end{figure}

We were able to generate a temporal map similar to the ones produced by using
raster printing projectors. Figure~\ref{fig:temporal} displays the temporal map
of Figure~\ref{fig:depth_doll} setup, created by ESL and E-RGB-D (ours). Although
we do not use the temporal map, this figure shows that we could generate one
simply by knowing which column is being projected by the DMD projector and
assigning a temporal index/color to that specific receiving event. However, in
the other methods, they need to receive all events and normalize the temporal
map based on the time of receiving the first and the last event.

Moreover, in X-maps, to obtain values for all possible measurements from the
camera and create the lookup reference, they needed to record events at least
for one time. However, we do not need to record anything and we can directly
find out the depth by knowing the column of the projected line on the rectified
projector image.

In practice, the projector's resolution is usually higher than that of the camera sensor, which means the EC may not capture individual projector columns or may see overlapping columns. To address this, we introduced a gap between each line in our fully dense patterns, ensuring a solid, dense temporal map with the camera.

Our method publishes events, and a separate ROS node aggregates these events and
publishes frames at various speeds. Since our system operates on an event-based
model rather than a frame-based one, it does not require complete pattern
captures. However, as a consequence of this approach, some patterns may not be
fully captured at higher frequencies. This contrasts with traditional methods
that rely on complete pattern captures to generate data. One metric for
assessing output quality is the Fill Rate (FR), which compares the number of
pixels containing data in the current frame to those in the ground truth frame.
For instance, patterns with 23 lines complete faster compared to those with 45
lines, allowing quicker coverage of the field of view and achieving a higher FR.
However, reducing the number of lines sacrifices detail. This trade-off between
speed and detail is a deliberate aspect of our approach, which is not achievable
with other methods. Raster-based projectors frequently project a solid pattern,
while methods using coded patterns like grayscale or binary codes require
capturing all patterns to report depth. They cannot increase speed by reducing
detail, nor can they enhance detail by sacrificing speed.

\begin{algorithm}
	\caption{Stamping events with depth and color}
	\label{alg:event_algorithm}
	
	\begin{algorithmic}[1]
		\State Initialize $\text{LUT}_{p}$, $\text{LUT}_{c}$, and $Columns$ based on calibration and pattern mode
		
		\While{camera triggers events}
		\If{external trigger}
		\State Increment $m$ and/or indicate the new color
		\Else
		\State Capture $\text{Event}_{c}(x_{c},y_{c})$
		\State Retrieve $\text{Event}_{c_{r}}(x_{E},y_{E})$ from $\text{LUT}_{c}(x_{c},y_{c})$
		\State Retrieve $x_{p}$ from $Columns[m]$
		\State Retrieve $\text{Top}_{p_{r}}(x_{T},y_{T})$ and $\text{Bottom}_{p_{r}}(x_{B},y_{B})$
		from $\text{LUT}_{p}(x_{p}, H)$ and $\text{LUT}_{p}(x_{p}, 0)$
		\State Calculate $x_{p_{r}} = x_{T} + (y_{E} - y_{T}) \cdot \frac{x_{B} - x_{T}}{y_{B} - y_{T}}$
		\State Compute $\text{depth} = \frac{f \cdot B}{x_{p_{r}} - x_{E}}$
		\State Publish depth and color-stamped events alongside the colored point cloud on ROS topics.
		\EndIf
		\EndWhile
		
	\end{algorithmic}
\end{algorithm}

\section{Experiments}\label{experiments}

This section assesses the performance of our event-based SL system for depth and
color estimation. We begin by introducing the hardware setup
(Section~\ref{setup}), along with the baseline methods and ground truth used for
comparison (Section~\ref{baseline}). Subsequently, we conduct experiments on
static scenes to quantify the accuracy of the proposed method and on dynamic
scenes to demonstrate its high-speed acquisition capabilities
(Section~\ref{results}).

\subsection{Setup}\label{setup}
\begin{figure}[htbp]
	\begin{center}
		\setlength{\tabcolsep}{1pt}
		\begin{tabular}{cc}
			\includegraphics[width=0.4\linewidth]{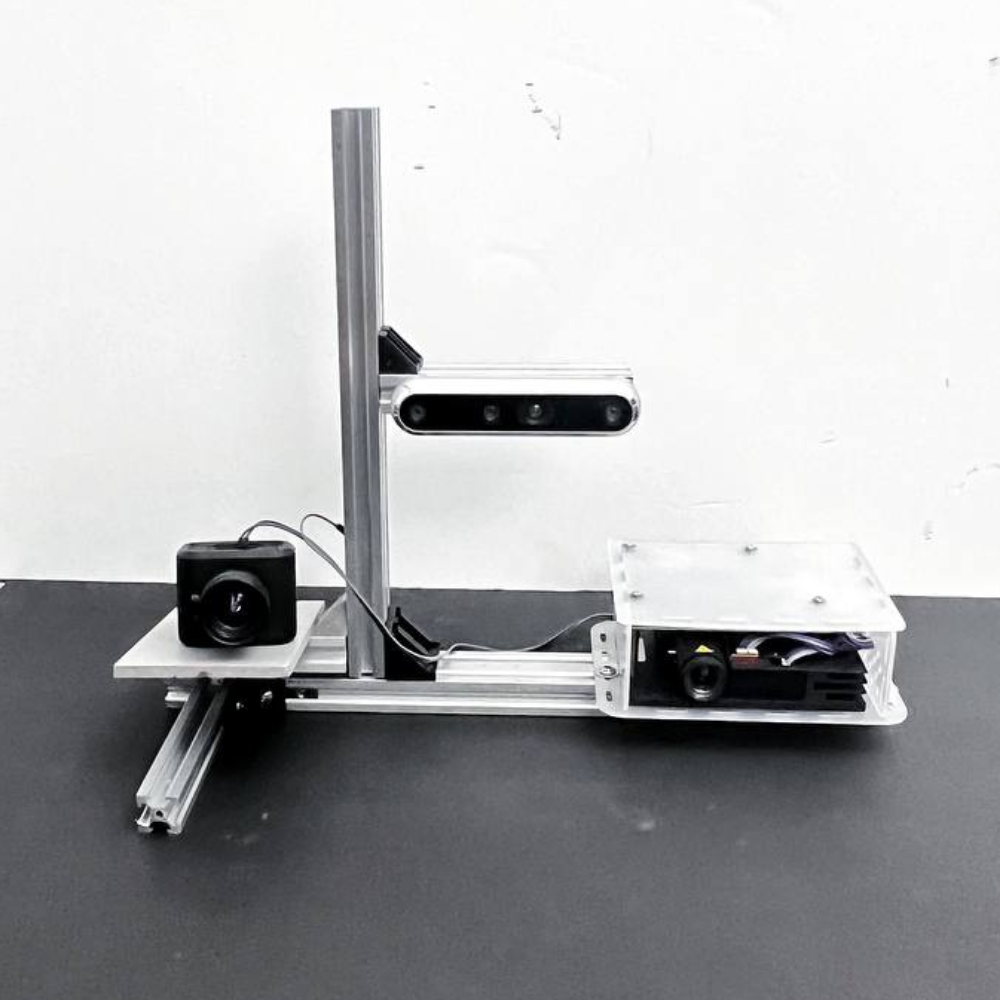}&
			\includegraphics[width=0.4\linewidth]{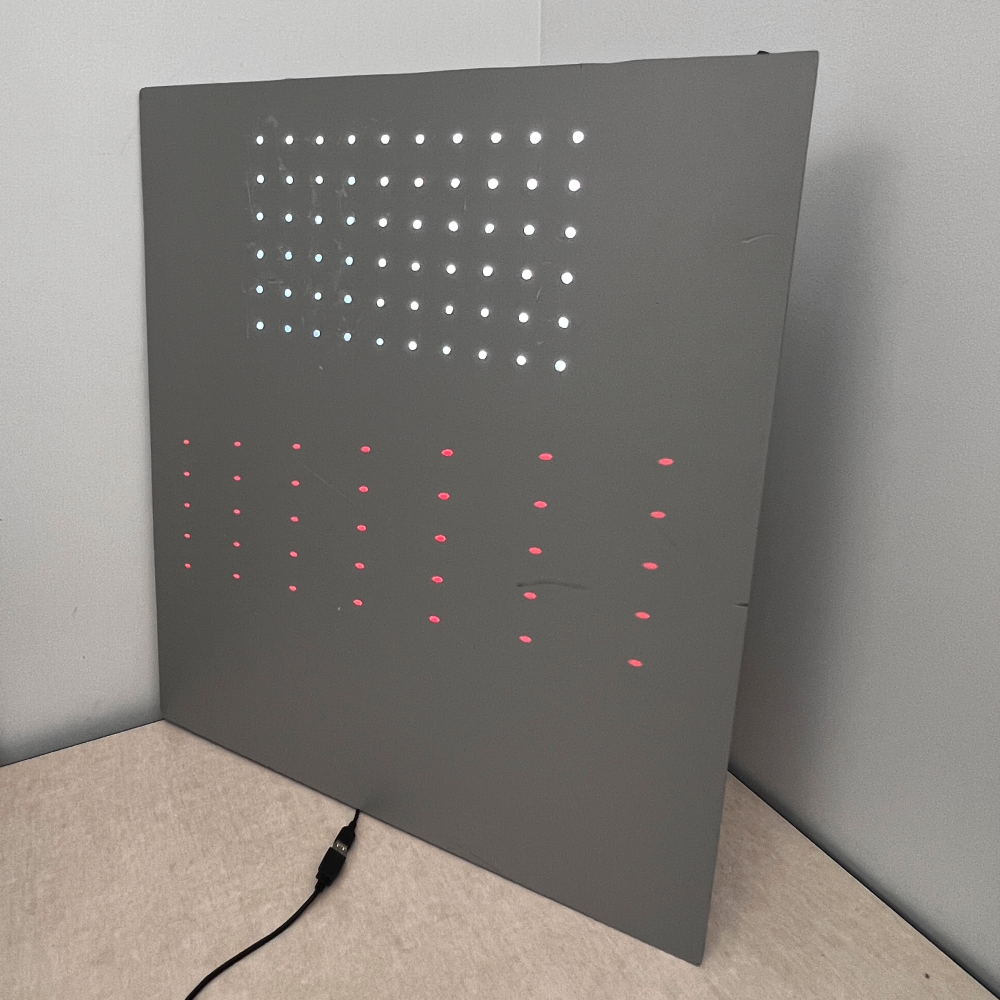}\\
		\end{tabular}
		\caption{\textbf{Left:} The experimental setup includes a DLP LightCrafter 4500 Evaluation Module, a Prophesee evaluation kit (Gen3-VGA), and an Intel RealSense D455 (only used for comparison).
			\textbf{Right:} The calibration circle dot-board used for calibrating the EC and camera-projector setup, with white dots from LEDs and red dots projected by the projector.}
		
		\label{fig:setup}
	\end{center}
\end{figure}

\textbf{Camera:} The event camera used in this study is the Prophesee Evaluation Kit 1 (EVK1) equipped with a Gen~3.0 sensor (\cite{prophesee}).
This dynamic vision sensor offers a
resolution of \mbox{$640\times480$} pixels with a 15~\textmu m pixel pitch and only detects
contrast changes. It features a dynamic range greater than 120~dB, an average
latency of 200~$\mu$s, and timestamps events with microsecond precision
(Figure~\ref{fig:setup}).

\begin{figure}[htbp]
	\centering
	{\includegraphics[width=0.22\linewidth]{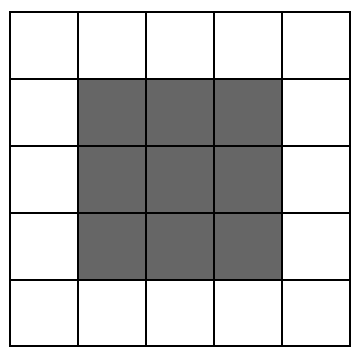}}
	\rotatebox{45}{\includegraphics[width=0.2\linewidth]{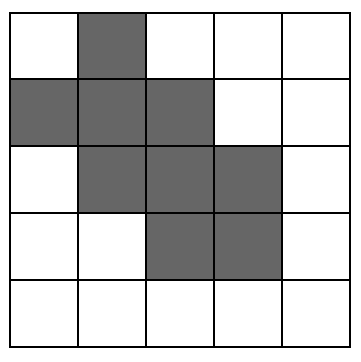}}
	{\includegraphics[width=0.22\linewidth]{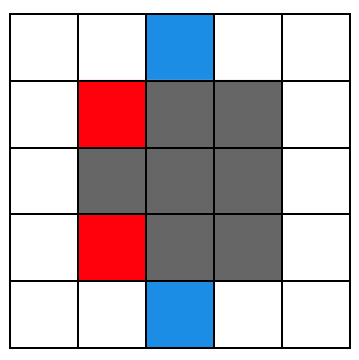}}
	{\includegraphics[width=0.22\linewidth]{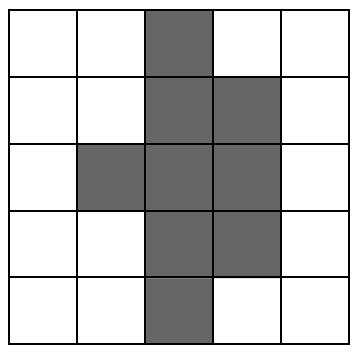}}
	\vspace{1pt}
	\begin{tabularx}{\linewidth}{*{4}{>{\centering\arraybackslash}X}}
		A & B & C & D \\
	\end{tabularx}
	\caption{The pattern of one dot with the size of \mbox{$3\times3$} pixels in our dot-based patterns. If we project (A) due to the diamond pixel configuration of the DMD, we will see (B) on the object surface. Which could lead to mislocating the center of the projected dot. Therefore, we should move red-colored pixels to blue ones in (C) and project (D) to achieve the pattern (A) on the object's surface with a 45-degree rotation.}
	\label{fig:diamond}
\end{figure}

\textbf{Projector:} To project binary patterns at high speed (over 4~kHz), we used the~\cite{dlpLC4500} DLP LightCrafter 4500 projector, which supports a resolution of \mbox{$912\times1140$} in a diamond pixel configuration and operates with a 235~$\mu$s exposure period, allowing a 4.225~kHz pattern-switching rate that is easily captured
by the EC (Figure~\ref{fig:setup}). Because of the diamond pixel array of the
DMD, the pixel data does not appear on the DMD exactly as it would in an
orthogonal pixel arrangement. Figure~\ref{fig:diamond} shows the pattern that we
used to project dots in our dot-based patterns.

\textbf{Calibration:}
To calibrate the system, we introduced a camera-projector calibration procedure specifically designed for event-based SL. Our method involves calibrating the intrinsic parameters of the event camera by utilizing a standard calibration tool (OpenCV;~\cite{opencv_library}) applied to images obtained by converting events into intensity frames. This conversion is done while observing a flickering circle-grid pattern from various angles. We chose circle patterns over checkerboards due to their superior performance in terms of both the quality and stability of the final calibration results across multiple iterations, as shown by~\cite{learning_opencv3}. Unlike checkerboards, which may lead to uncertainties in corner detection, circle-grid patterns allow for more precise extraction of circle centers, for instance, through the calculation of the center of gravity of all circle pixels.

Our approach for this calibration is straightforward and adaptable. We begin by setting up a circle grid of flickering LEDs on a flat surface and projecting another circle grid pattern using the projector beside it on the board. After calibrating the intrinsic parameters of the event camera using the LED circle-grid pattern, we proceed to calibrate the extrinsic parameters of the camera-projector setup and the intrinsic parameters of the projector.

Once the camera calibration is complete, we determine the board's relative position, which in turn allows us to calibrate the projector. This streamlined method enables the simultaneous calibration of all parameters through simple operations, significantly reducing the complexity typically associated with calibration processes. Furthermore, it is universally applicable to all event-based SL systems. Figure~\ref{fig:setup} shows the calibration circle dot-board used for calibrating the system.

\cite{luo2014simple} proposed an alternative approach for calibrating the SL camera-projector system. Their method involves introducing four reference planes and generating lookup tables for pixel correspondences. While this method can enhance calibration quality, it adds complexity to the procedure.

Similarly, 
\cite{wang2020temporal} presented a calibration technique based on Temporal Matrices Mapping (TMM). They utilized two temporal matrices to establish pixel correspondences between the SL projector plane and the event camera (EC) plane. Although this approach can improve calibration accuracy, it also increases the overall complexity of the calibration process.

\textbf{Software:}
We have implemented our method on the Robot Operating System (ROS). The designed Qt-based Graphical User Interface (GUI) enables us to control the camera settings online and allows the ROS nodes to publish color- and depth-stamped events alongside RGB frames and colorful point clouds on ROS topics, as shown by~\cite{Bajestani_2025_RoboCup}. More details can be found on our Event-based RGBD ROS Wrapper (see~\cite{Bajestani_2023_WACV}) Git repository at \href{https://github.com/MISTLab/event_based_rgbd_ros}{github.com/MISTLab/event\_based\_rgbd\_ros}.

\subsection{Baseline and Ground Truth}\label{baseline}

As shown in Table~\ref{tab:related}, the most recent related works are the
ESL, as proposed by~\cite{muglikar2021esl}, the X-maps, as introduced by~\cite{morgenstern2023x}, and the
SEG, as described by~\cite{lu2024sge}. Since all of them used a laser point projector with a
scanning speed of 60 Hz, they were automatically excluded from comparison in
high-speed real-time event-based scanning.

Although X-maps presents a high-speed structured light method using event cameras, it relies on a raster-scanning laser projector and constructs a rectified X-map from a time-discretized spatio-temporal event cuboid. Our system, in contrast, does not accumulate or discretize events into frames; it instead performs disparity estimation and 3D reconstruction per event in real-time. Due to these fundamental differences in both hardware and methodology, direct benchmarking with X-maps is not feasible. Therefore, we focused our comparisons on ESL, as proposed by~\cite{muglikar2021esl}, which is compatible with recorded data and serves as a fair, reproducible baseline for evaluating our method in a similar structured light context.

Since, the ESL used a raw recorded event
file and pre-data-processing to synchronize and convert the raw event data file
in the absence of triggers. This makes it more accurate (but slower) than the
X-maps method because it considers all events together rather than processing
them in real-time. More importantly, we could still generate the time-maps even
by using a DLP. We did not use the SEG method in dynamic conditions because
their code is not yet publicly available.

In this task, obtaining the GT is challenging due to the lack of methods capable
of producing dense depth with accuracy above the millimeter level for natural
scenes. Therefore, ESL, as proposed by~\cite{muglikar2021esl}, adopts averaged
MC3D, as introduced by~\cite{matsuda2015mc3d}, as their ground truth, whereas
X-maps, as described by~\cite{morgenstern2023x}, employ optimized ESL. While we acknowledge the
potential for accuracy enhancement through averaging operations in depth
estimation, our approach differs due to reconstructing color alongside depth.
Given the utilization of a distinct point scanning method and projector type, we
developed our method to generate the ground truth by accumulating more events
over an extended time window (one second) and performing averaging across 10
scans.

\subsection{Results}\label{results}

\begin{figure}[h]
	\centering
	\begin{tabularx}{\linewidth}{p{0.28\linewidth}p{0.28\linewidth}p{0.28\linewidth}}
		\includegraphics[width=\linewidth]{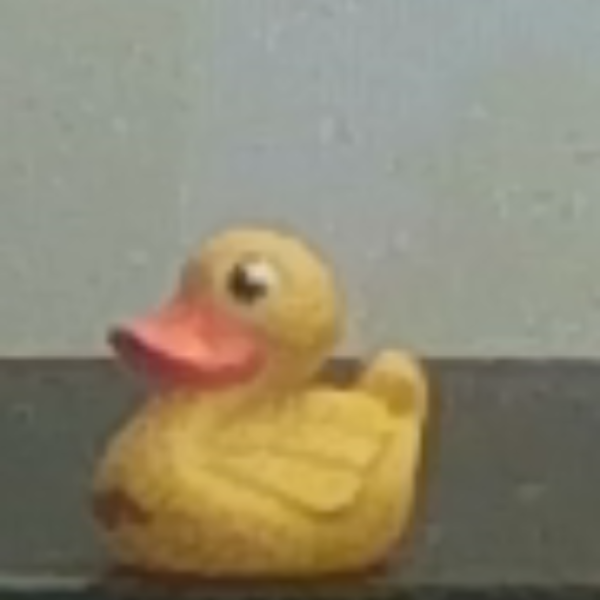}&
		\includegraphics[width=\linewidth]{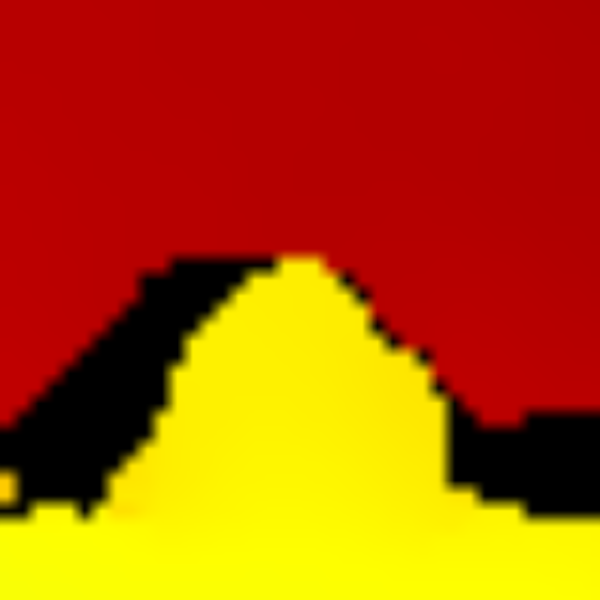}&
		\includegraphics[width=\linewidth]{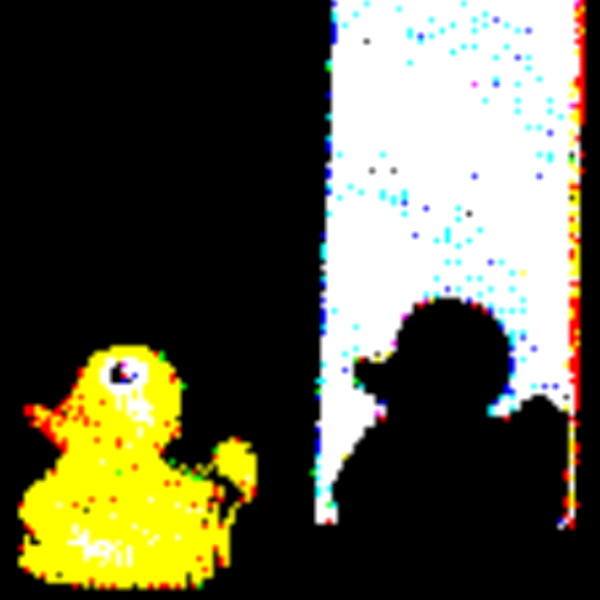}\\
		\small D455~33~ms & \small D455~33~ms & \small \textbf{Ours} 17.23~ms\\
		\includegraphics[width=\linewidth]{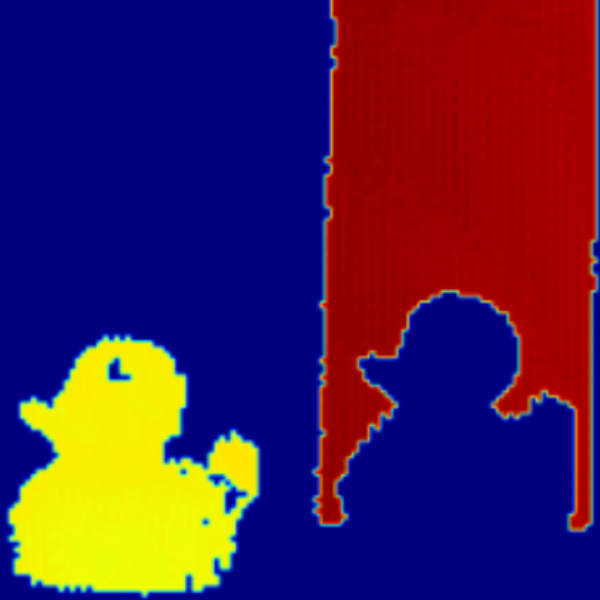}&
		\includegraphics[width=\linewidth]{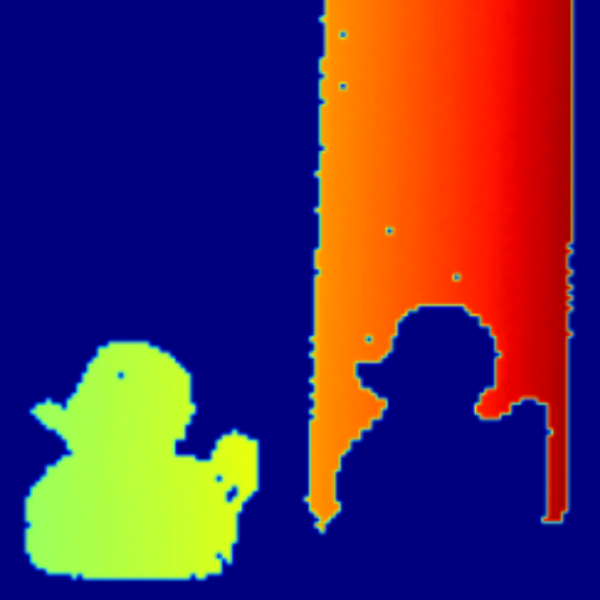}&
		\includegraphics[width=\linewidth]{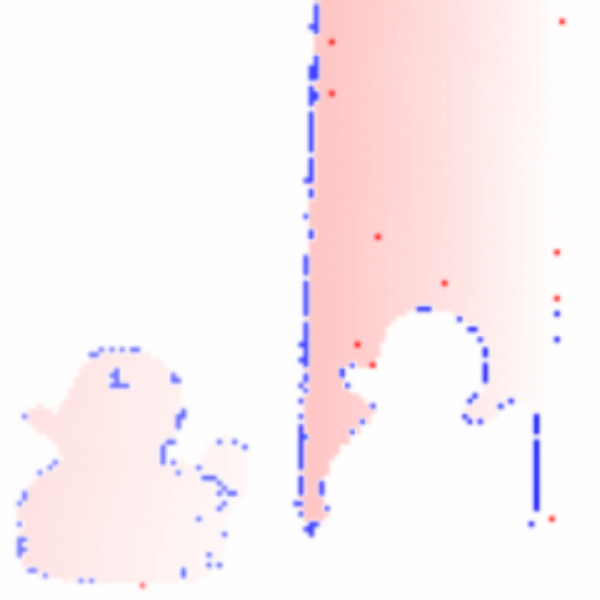}\\
		\small GT 10~s & \small ESL 1.55~s & \small GT Difference\\
		\includegraphics[width=\linewidth]{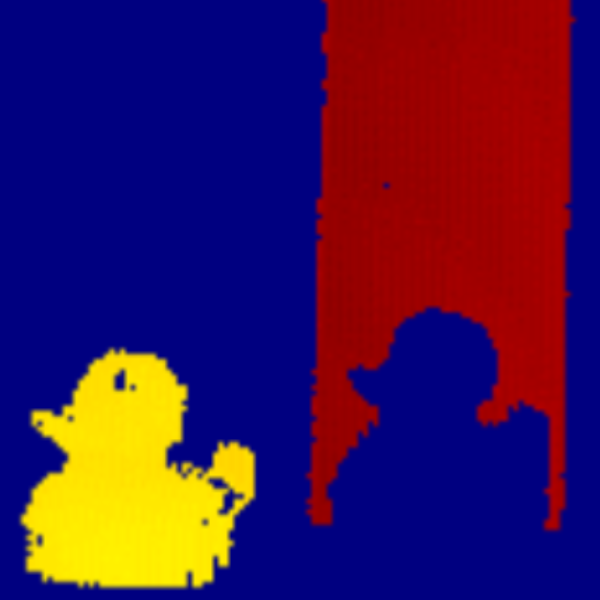}&
		\includegraphics[width=\linewidth]{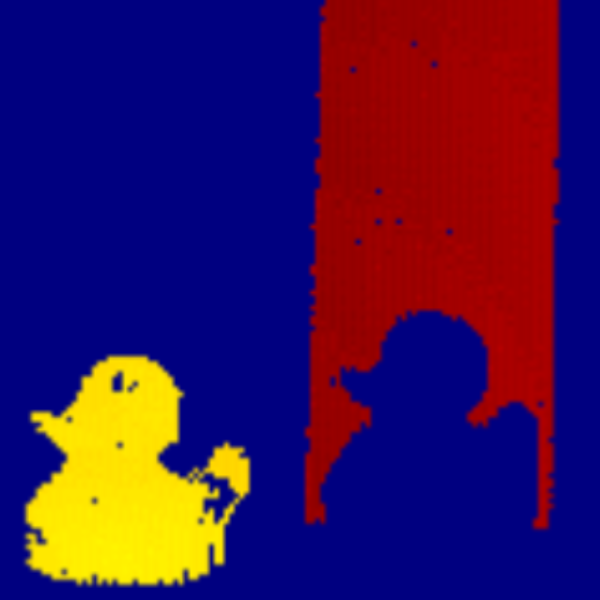}&
		\includegraphics[width=\linewidth]{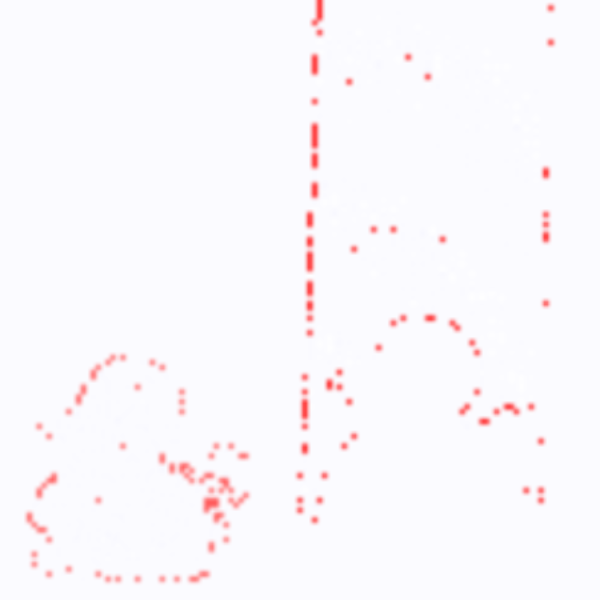}\\
		\small GT 10s~& \small \textbf{Ours} 17.23~ms &\small GT Difference \\
	\end{tabularx}
	\caption{Comparison of color and depth detection for Duck setup (M4L23 pattern).}
	\label{fig:static_duck}
\end{figure}

\begin{figure*}[htbp]
	\centering
	\begin{tabularx}{\linewidth}{p{0.075\linewidth}p{0.075\linewidth}p{0.075\linewidth}p{0.075\linewidth}p{0.075\linewidth}p{0.075\linewidth}p{0.075\linewidth}p{0.075\linewidth}p{0.075\linewidth}p{0.075\linewidth}}
		D455 33.33~ms RGB &
		\textbf{Ours} M3D45 RGB & \textbf{Ours} M4L23 RGB &
		D455 33.33~ms Depth &
		\textbf{Ours} M3D45 Depth & ESL M3D45 Depth &
		\textbf{Ours} M4L23 Depth & ESL M4L23 Depth &
		\textbf{Ours} M3L45 Depth & ESL M3L45 Depth \\
		\hline
		\addlinespace
		\end{tabularx}
		\begin{tabular}{ccccccccccc}
		\rotatebox{90}{\small{Yoda}}
		{\includegraphics[width=0.090\linewidth]{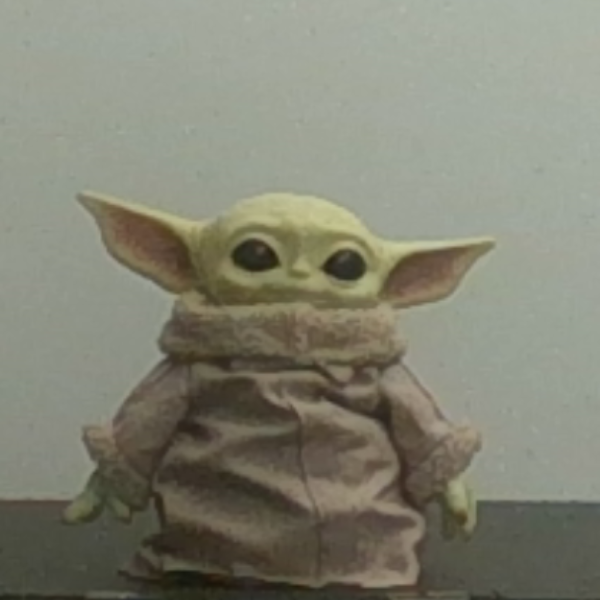}}
		{\includegraphics[width=0.090\linewidth]{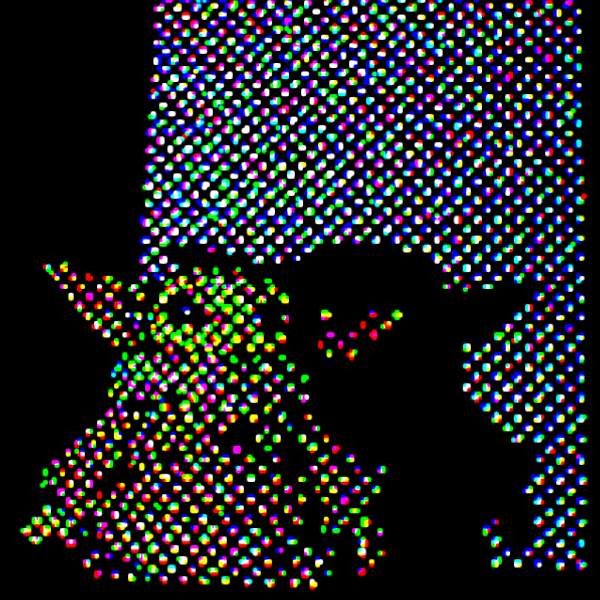}}
		{\includegraphics[width=0.090\linewidth]{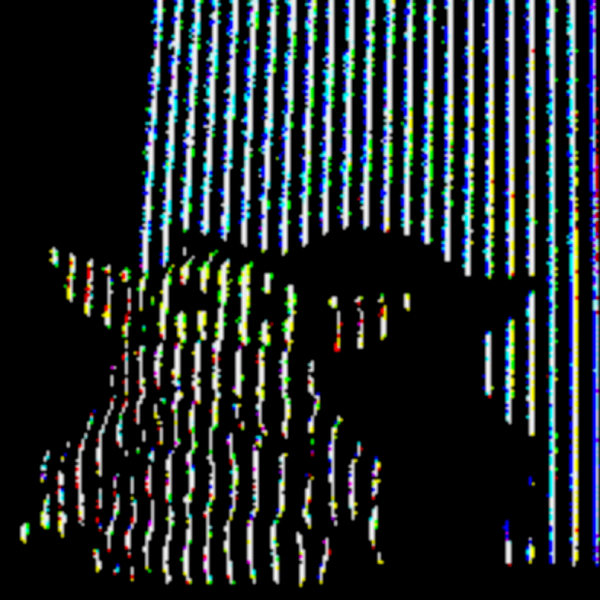}}
		{\includegraphics[width=0.090\linewidth]{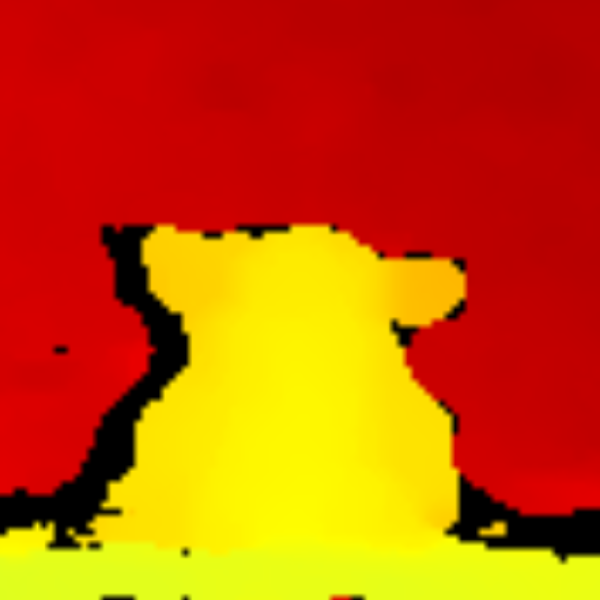}}
		{\includegraphics[width=0.090\linewidth]{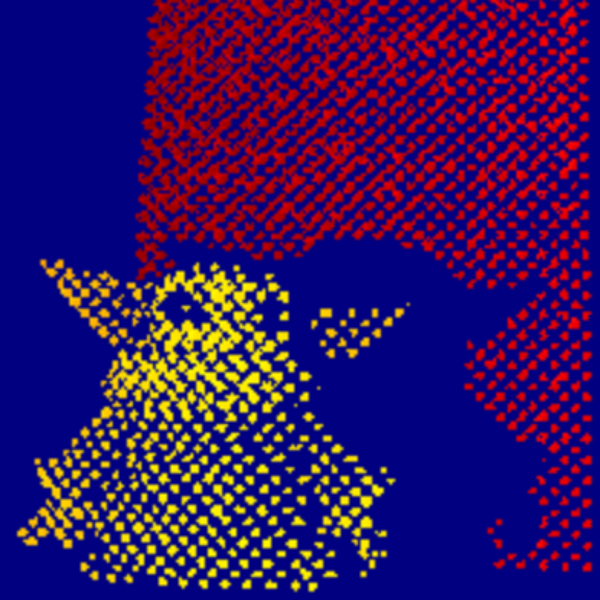}}
		{\includegraphics[width=0.090\linewidth]{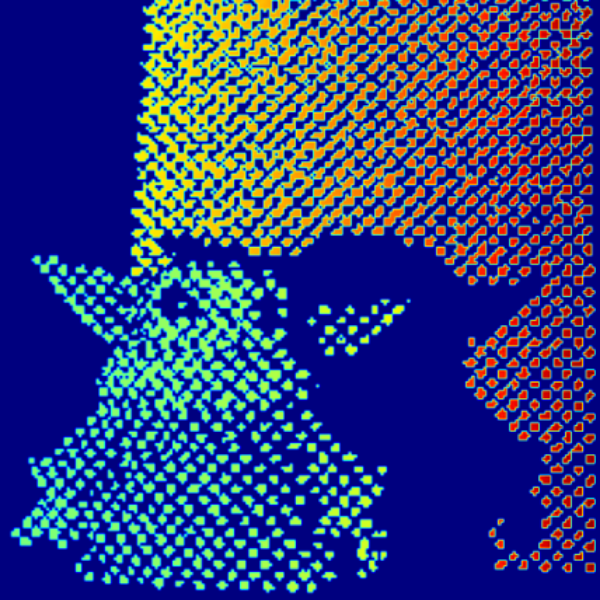}}
		{\includegraphics[width=0.090\linewidth]{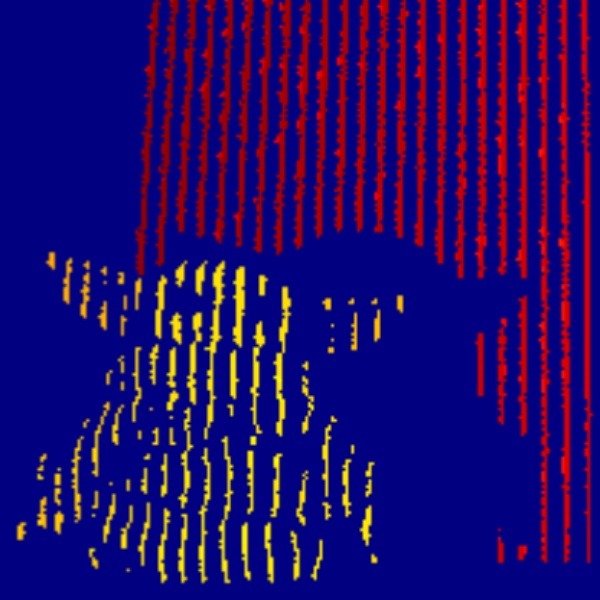}}
		{\includegraphics[width=0.090\linewidth]{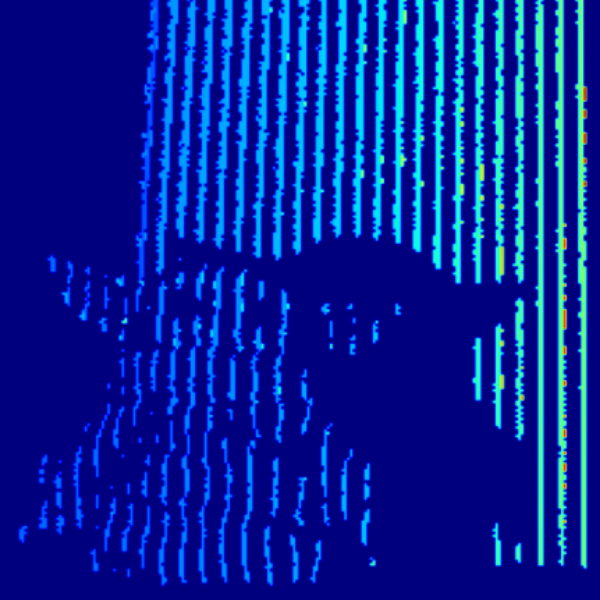}}
		{\includegraphics[width=0.090\linewidth]{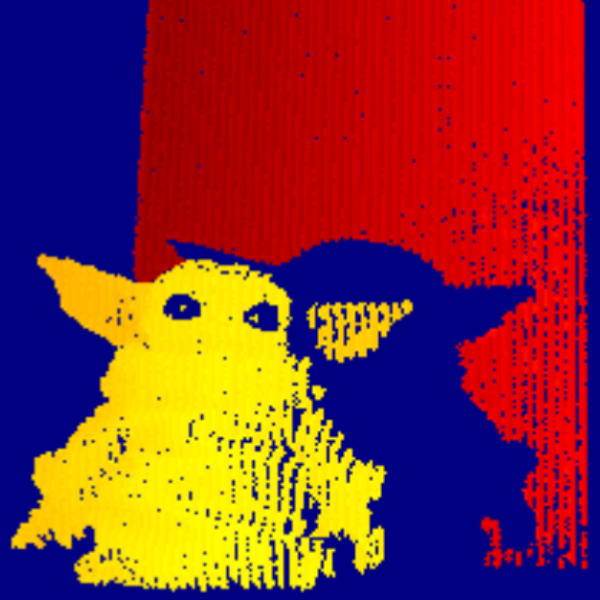}}
		{\includegraphics[width=0.090\linewidth]{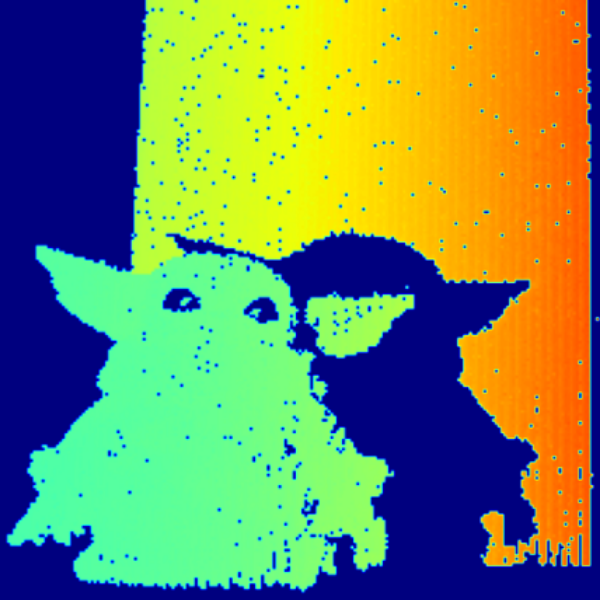}}
		\end{tabular}
		
		\begin{tabularx}{\linewidth}{p{0.001\linewidth}p{0.07\linewidth}p{0.075\linewidth}p{0.07\linewidth}p{0.07\linewidth}p{0.076\linewidth}p{0.066\linewidth}p{0.076\linewidth}p{0.066\linewidth}p{0.076\linewidth}p{0.066\linewidth}}
		& RMSE: Time: &
		20.04 \bms{12.57~ms} & 22.42 \bms{17.23~ms} &
		&
		330.23 \bms{12.57~ms} & 300.84 2.35 s &
		228.03 \bms{17.23~ms} & 241.12 1.93 s &
		335.03 \bms{12.57~ms} & 199.83 5.37 s \\
		\hline
		\addlinespace
		\end{tabularx}
		
		\begin{tabular}{ccccccccccc}
		\rotatebox{90}{\small{Foam}}
		{\includegraphics[width=0.090\linewidth]{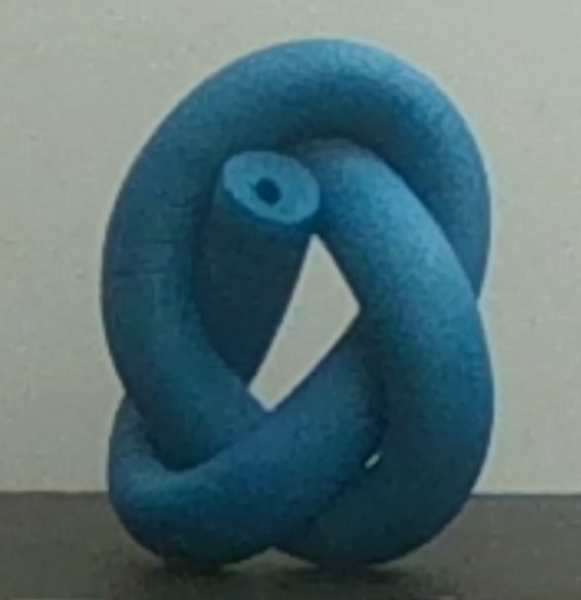}}
		{\includegraphics[width=0.090\linewidth]{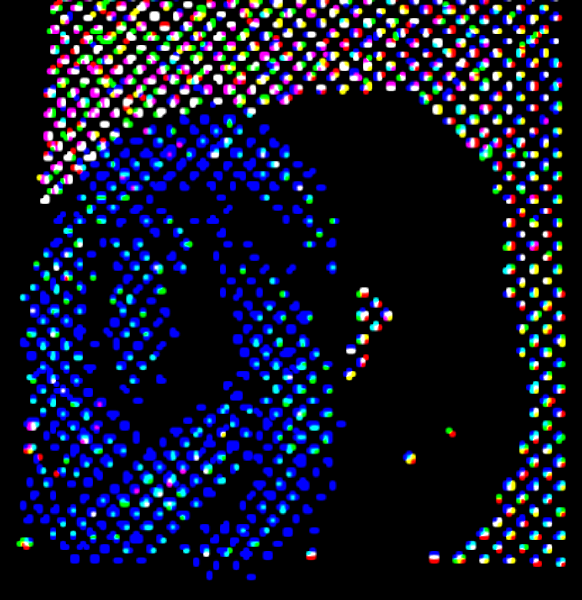}}
		{\includegraphics[width=0.090\linewidth]{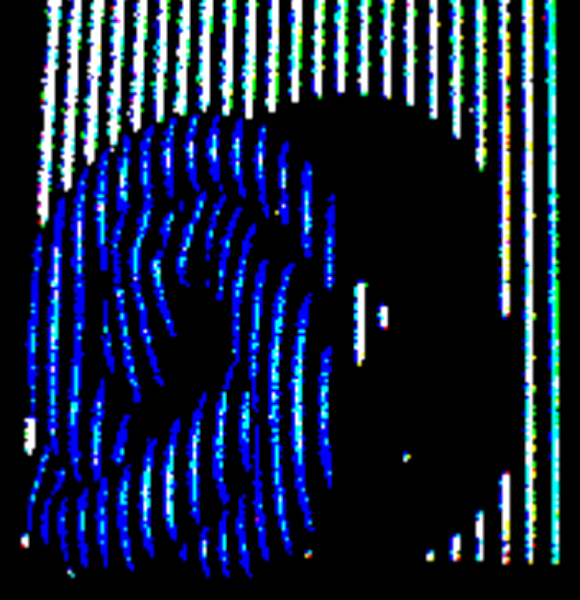}}
		{\includegraphics[width=0.090\linewidth]{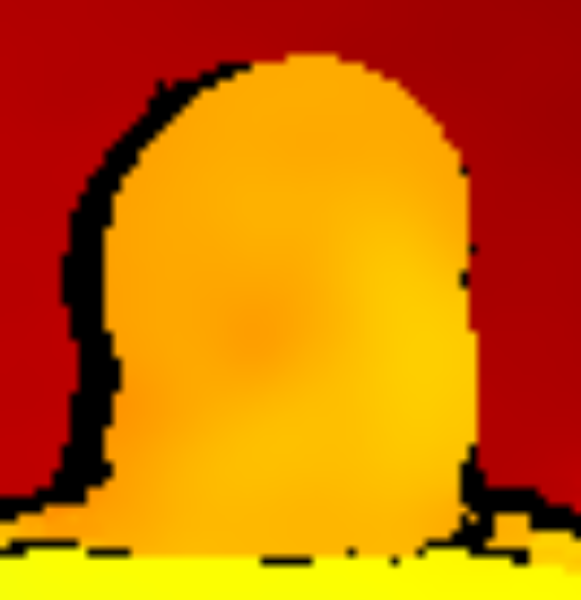}}
		{\includegraphics[width=0.090\linewidth]{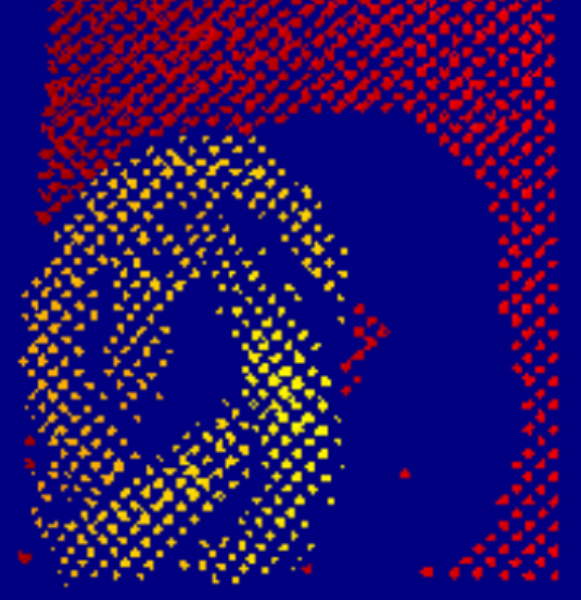}}
		{\includegraphics[width=0.090\linewidth]{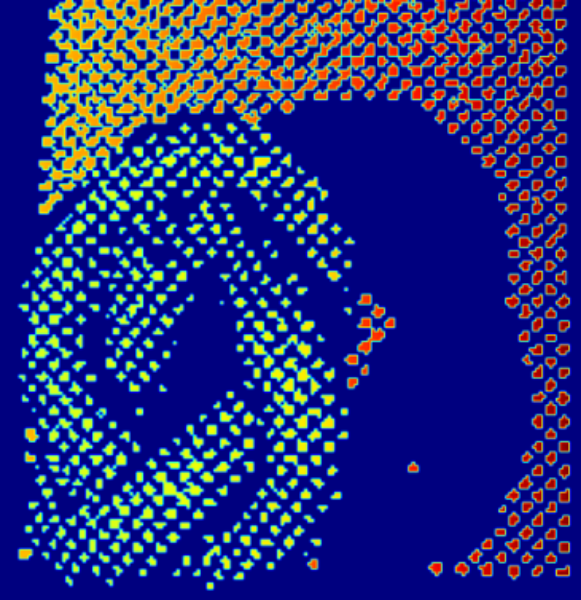}}
		{\includegraphics[width=0.090\linewidth]{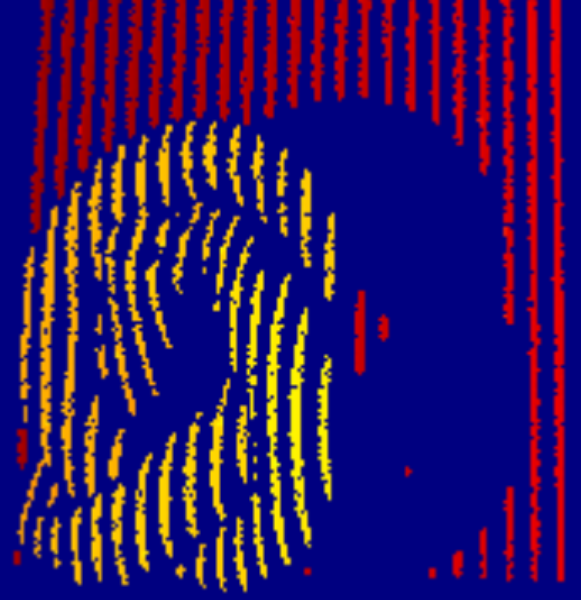}}
		{\includegraphics[width=0.090\linewidth]{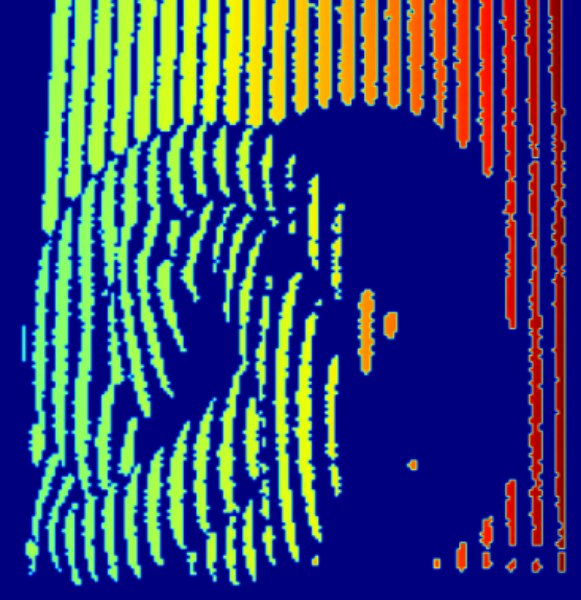}}
		{\includegraphics[width=0.090\linewidth]{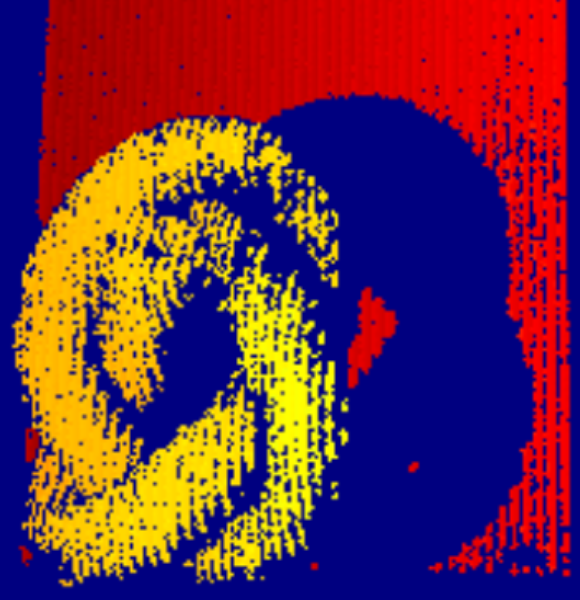}}
		{\includegraphics[width=0.090\linewidth]{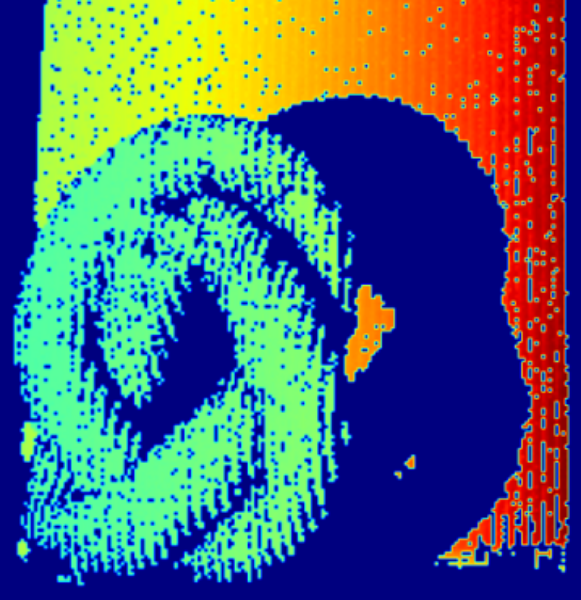}}
		\end{tabular}
		
		\begin{tabularx}{\linewidth}{p{0.001\linewidth}p{0.07\linewidth}p{0.075\linewidth}p{0.07\linewidth}p{0.07\linewidth}p{0.076\linewidth}p{0.066\linewidth}p{0.076\linewidth}p{0.066\linewidth}p{0.076\linewidth}p{0.066\linewidth}}
		& RMSE: Time: &
		19.11 \bms{12.57~ms} & 17.51 \bms{17.23~ms} &
		&
		328.98 \bms{12.57~ms} & 334.30 1.97 s &
		198.69 \bms{17.23~ms} & 196.33 2.70 s &
		161.86 \bms{12.57~ms} & 213.26 4.33 s \\
		\hline
		\addlinespace
		\end{tabularx}
		
		\begin{tabular}{ccccccccccc}
		\rotatebox{90}{\small{Duck}}
		{\includegraphics[width=0.090\linewidth]{Duck-bgr-D455.png}}
		{\includegraphics[width=0.090\linewidth]{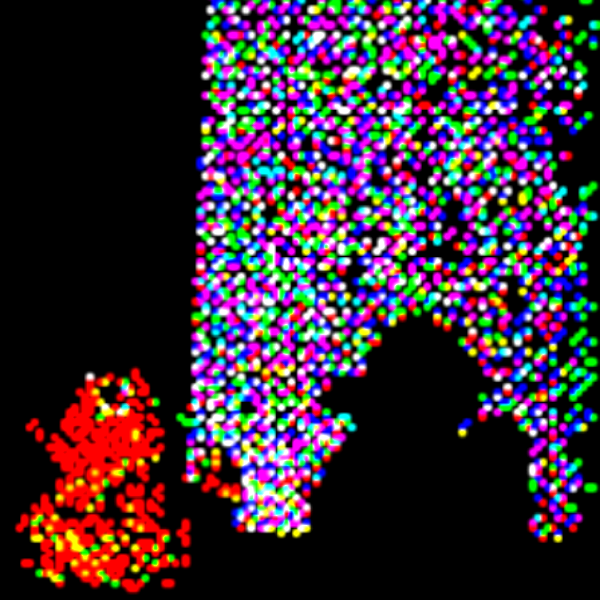}}
		{\includegraphics[width=0.090\linewidth]{Duck-bgr-M4L23-dense-58fps.png}}
		{\includegraphics[width=0.090\linewidth]{Duck-depth-D455.png}}
		{\includegraphics[width=0.090\linewidth]{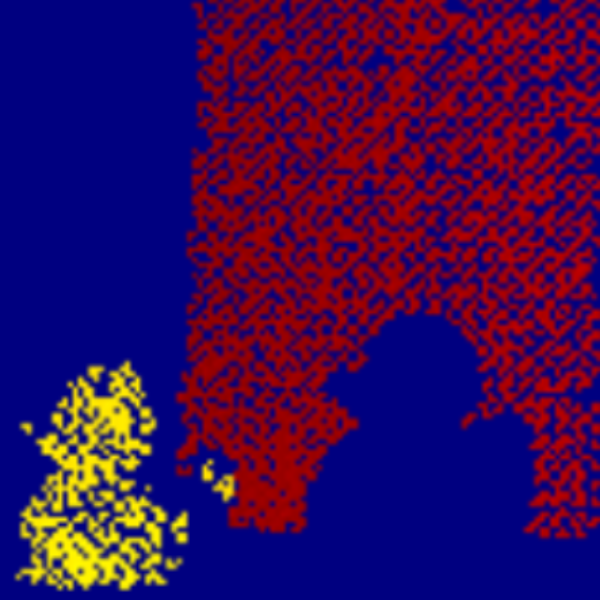}}
		{\includegraphics[width=0.090\linewidth]{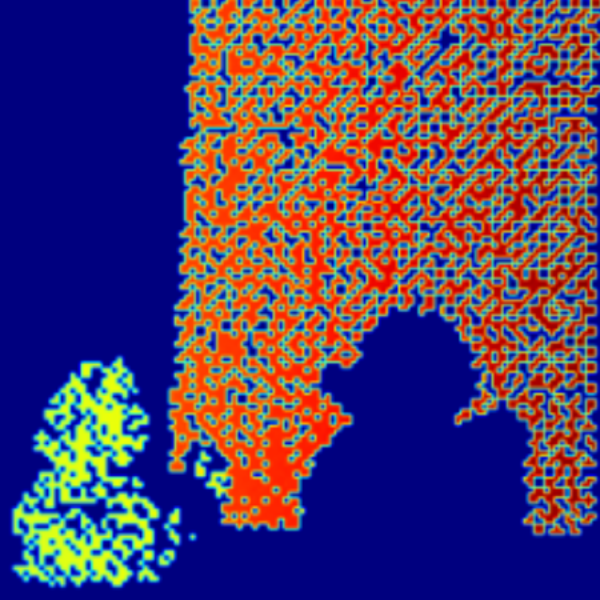}}
		{\includegraphics[width=0.090\linewidth]{Duck-depth-M4L23-dense-58fps.png}}
		{\includegraphics[width=0.090\linewidth]{Duck-depth-M4L23-dense-ESL.png}}
		{\includegraphics[width=0.090\linewidth]{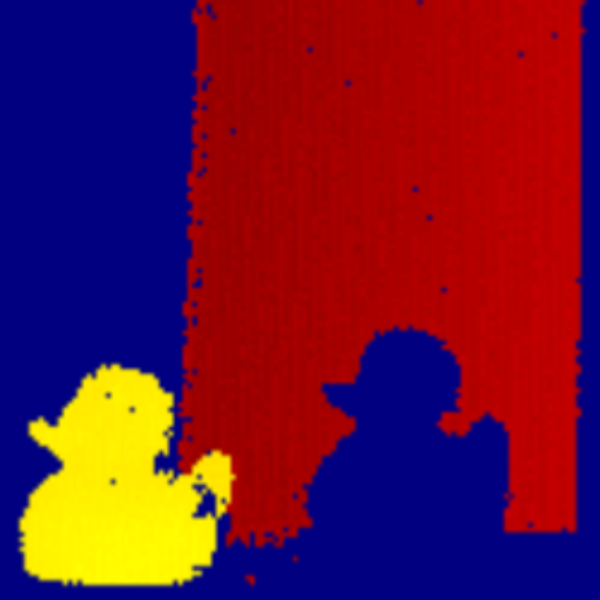}}
		{\includegraphics[width=0.090\linewidth]{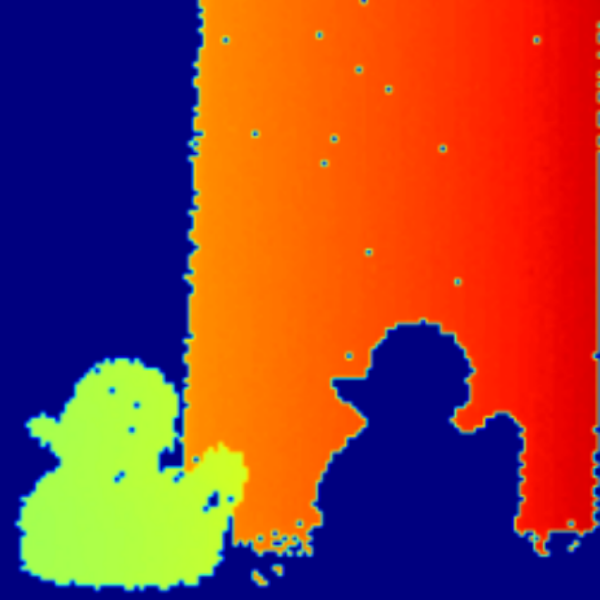}}
		\end{tabular}
		
		\begin{tabularx}{\linewidth}{p{0.001\linewidth}p{0.07\linewidth}p{0.075\linewidth}p{0.07\linewidth}p{0.07\linewidth}p{0.076\linewidth}p{0.066\linewidth}p{0.076\linewidth}p{0.066\linewidth}p{0.076\linewidth}p{0.066\linewidth}}
		& RMSE: Time: &
		18.64 \bms{12.57~ms} & 12.48 \bms{17.23~ms} &
		&
		189.59 \bms{12.57~ms} & 238.92 2.35 s &
		77.56 \bms{17.23~ms} & 74.10 1.55 s &
		85.28 \bms{12.57~ms} & 87.03 5.37 s \\
		\hline
		\addlinespace
		\end{tabularx}
		
		\begin{tabular}{ccccccccccc}
		\rotatebox{90}{\small{Bag}}
		{\includegraphics[width=0.090\linewidth]{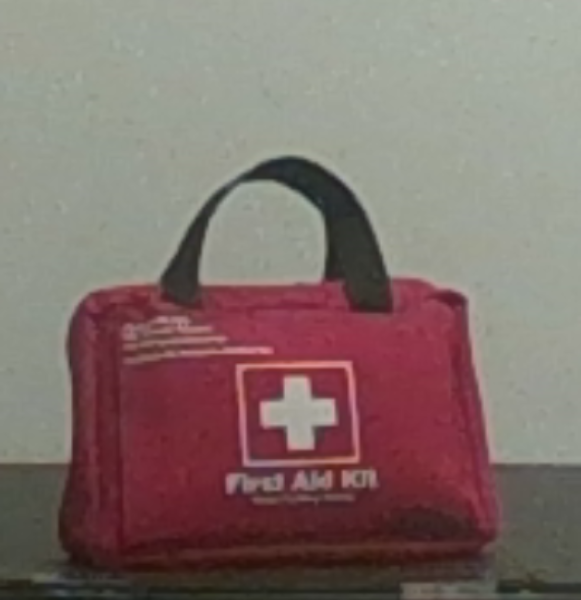}}
		{\includegraphics[width=0.090\linewidth]{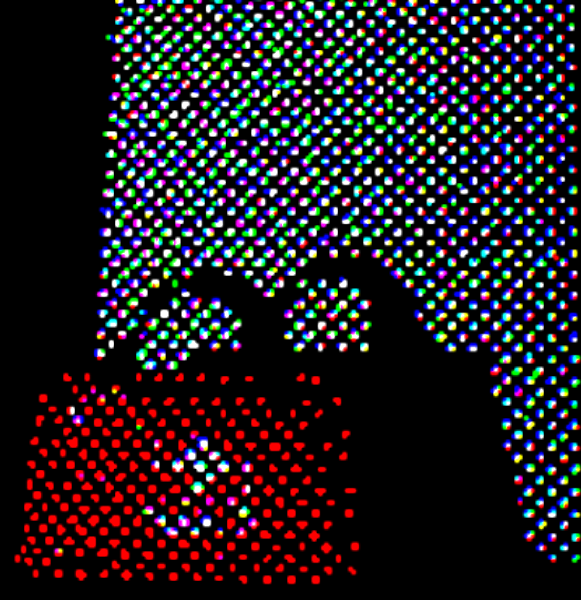}}
		{\includegraphics[width=0.090\linewidth]{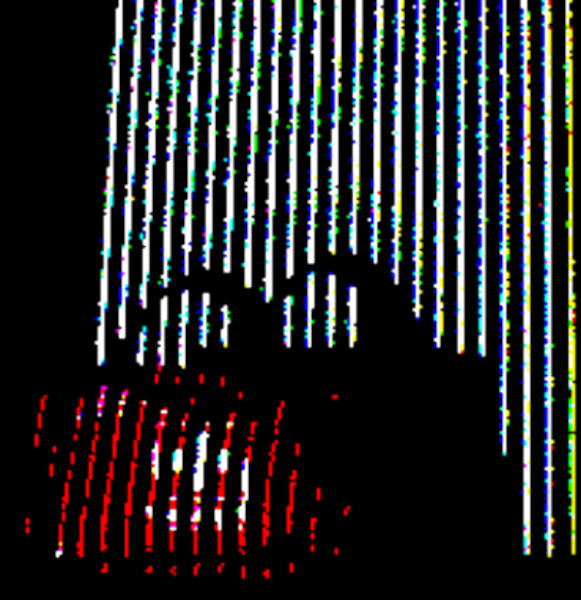}}
		{\includegraphics[width=0.090\linewidth]{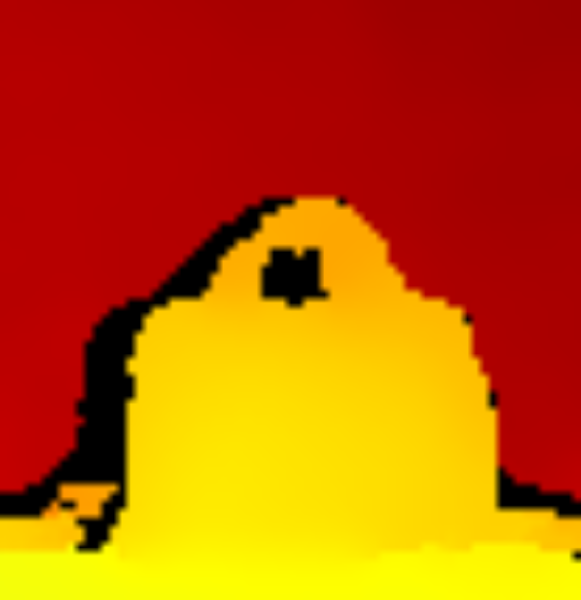}}
		{\includegraphics[width=0.090\linewidth]{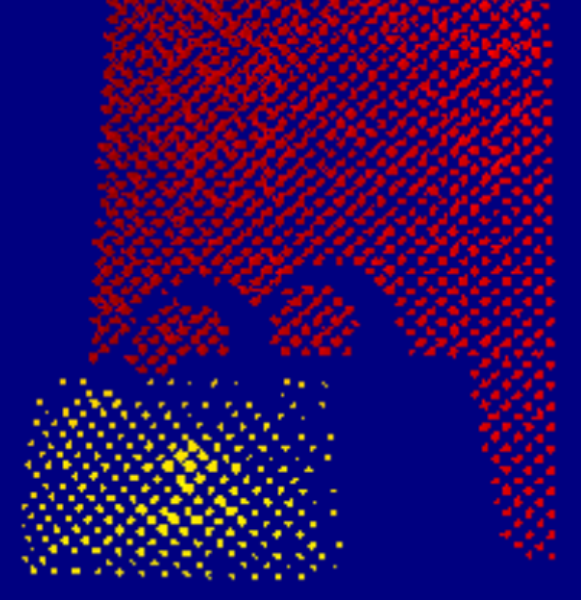}}
		{\includegraphics[width=0.090\linewidth]{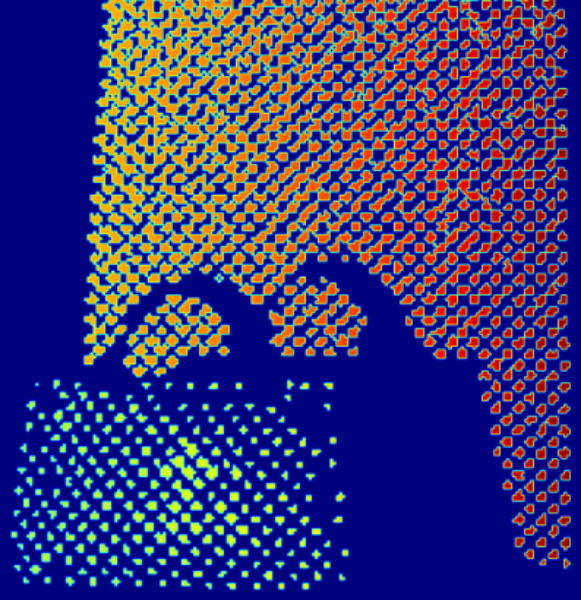}}
		{\includegraphics[width=0.090\linewidth]{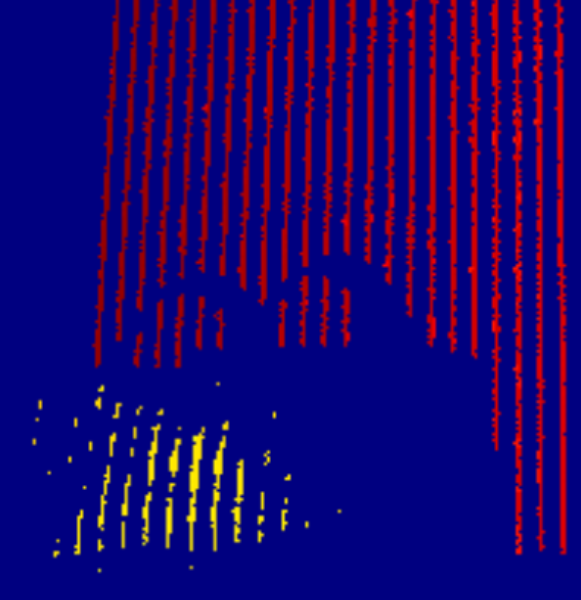}}
		{\includegraphics[width=0.090\linewidth]{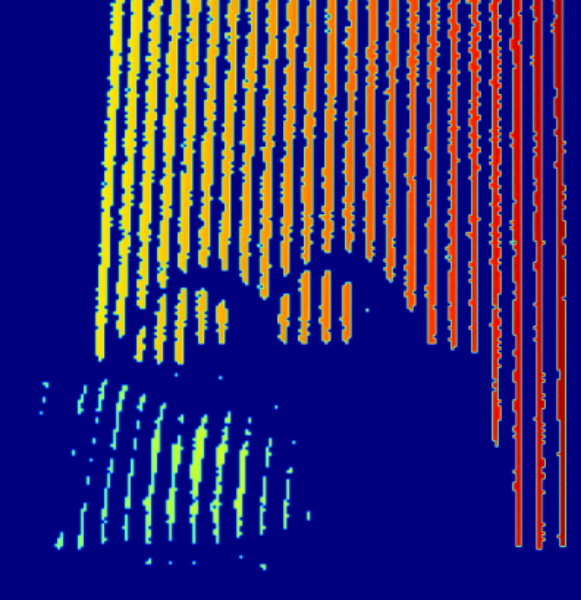}}
		{\includegraphics[width=0.090\linewidth]{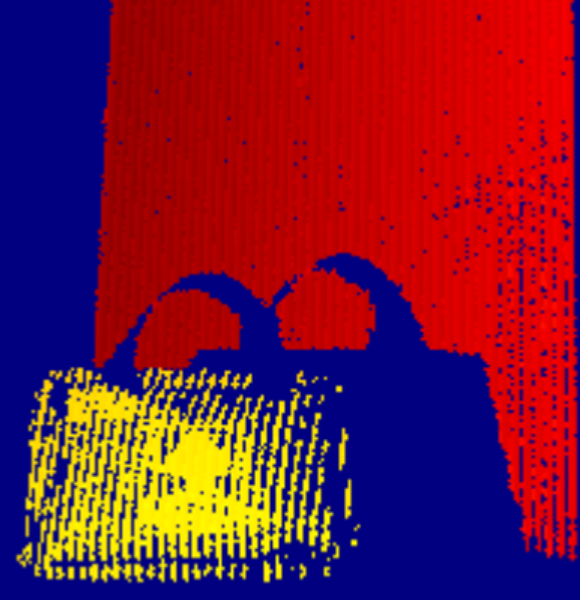}}
		{\includegraphics[width=0.090\linewidth]{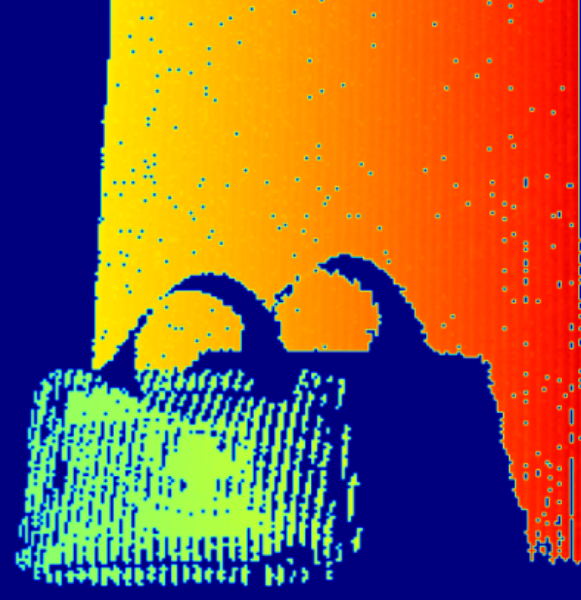}}
		\end{tabular}
		
		\begin{tabularx}{\linewidth}{p{0.001\linewidth}p{0.07\linewidth}p{0.075\linewidth}p{0.07\linewidth}p{0.07\linewidth}p{0.076\linewidth}p{0.066\linewidth}p{0.076\linewidth}p{0.066\linewidth}p{0.076\linewidth}p{0.066\linewidth}}
		& RMSE: Time: &
		22.38 \bms{12.57~ms} & 21.53 \bms{17.23~ms} &
		&
		371.73 \bms{12.57~ms} & 385.64 2.33 s &
		222.13 \bms{17.23~ms} & 189.69 2.14 s &
		326.28 \bms{12.57~ms} & 153.63 4.71 s \\
		\hline
		\addlinespace
		\end{tabularx}
		
		\begin{tabular}{ccccccccccc}
		\rotatebox{90}{\small{Circle}}
		{\includegraphics[width=0.090\linewidth]{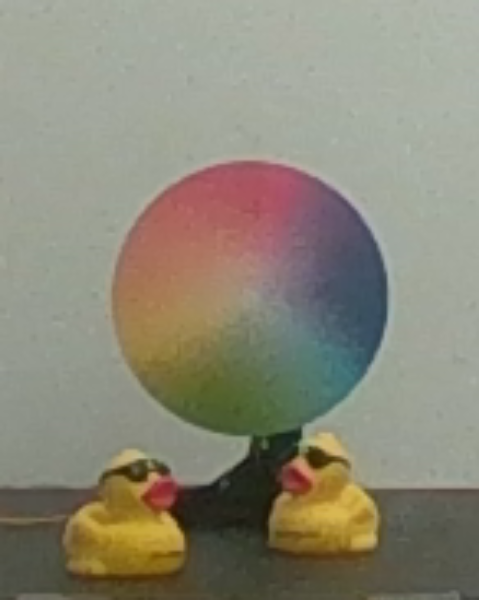}}
		{\includegraphics[width=0.090\linewidth]{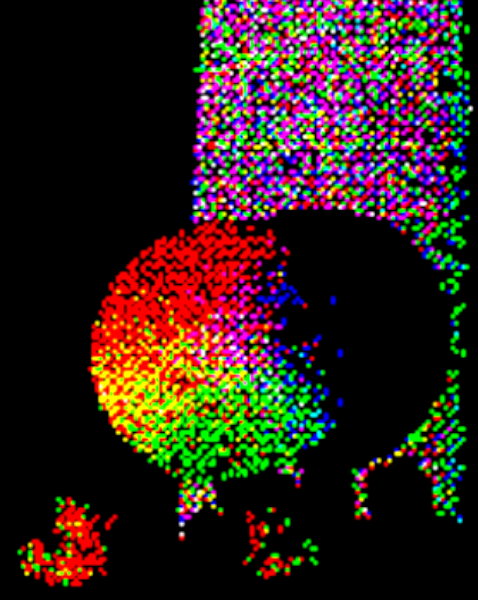}}
		{\includegraphics[width=0.090\linewidth]{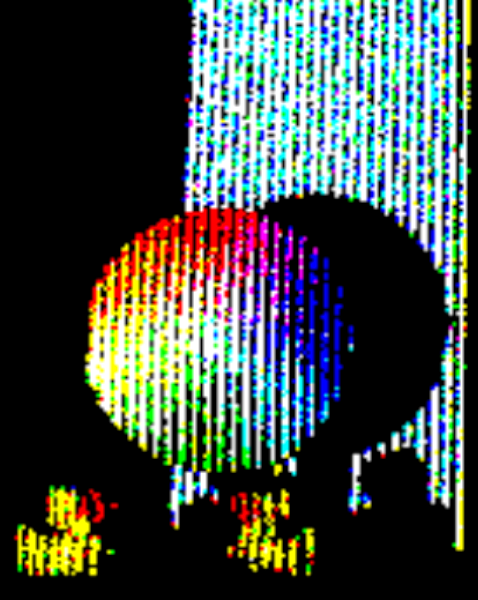}}
		{\includegraphics[width=0.090\linewidth]{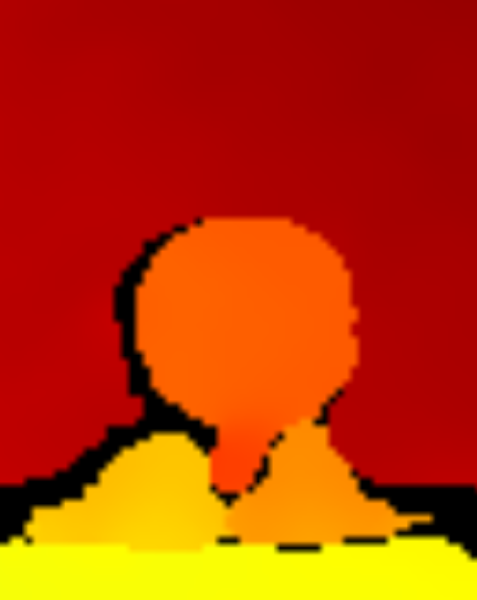}}
		{\includegraphics[width=0.090\linewidth]{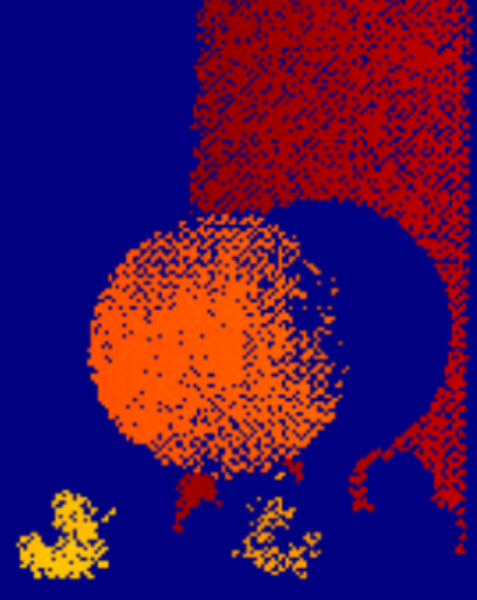}}
		{\includegraphics[width=0.090\linewidth]{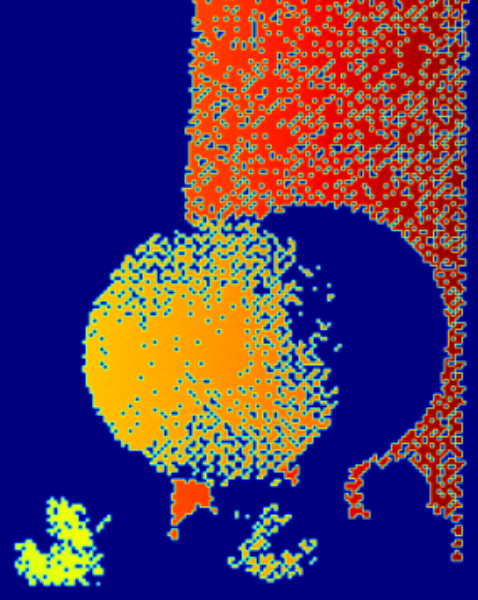}}
		{\includegraphics[width=0.090\linewidth]{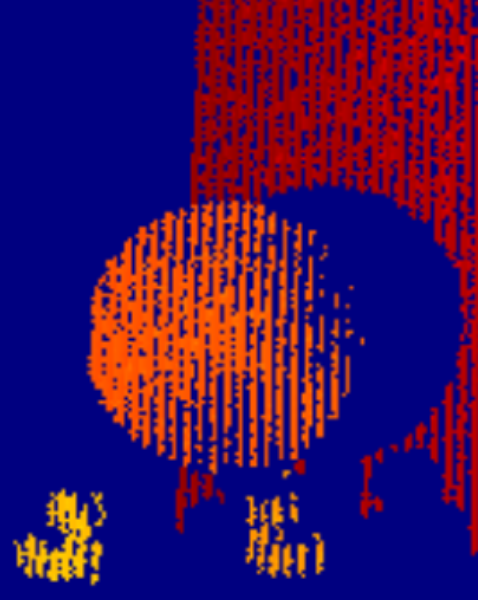}}
		{\includegraphics[width=0.090\linewidth]{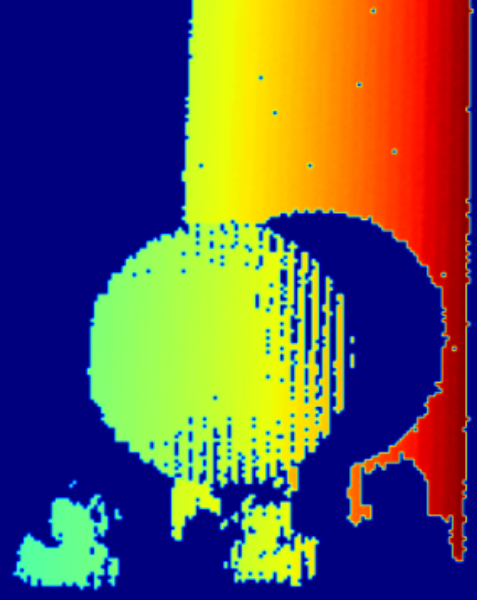}}
		{\includegraphics[width=0.090\linewidth]{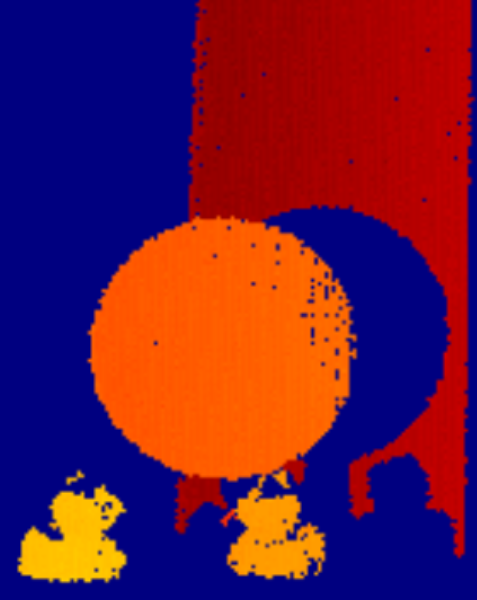}}
		{\includegraphics[width=0.090\linewidth]{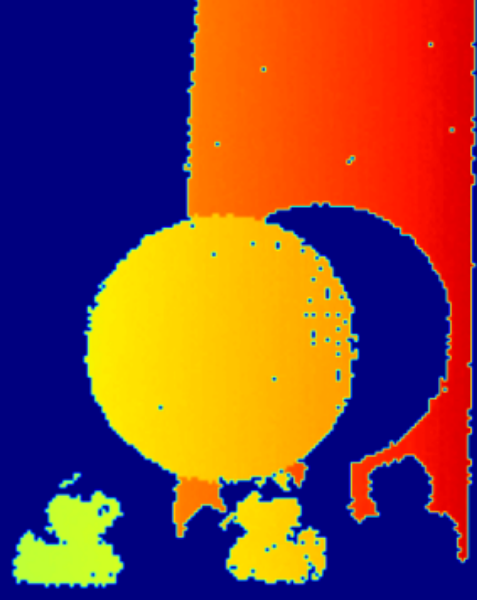}}
		\end{tabular}
		
		\begin{tabularx}{\linewidth}{p{0.001\linewidth}p{0.07\linewidth}p{0.075\linewidth}p{0.07\linewidth}p{0.07\linewidth}p{0.076\linewidth}p{0.066\linewidth}p{0.076\linewidth}p{0.066\linewidth}p{0.076\linewidth}p{0.066\linewidth}}
		& RMSE: Time: &
		19.03 \bms{12.57~ms} & 18.43 \bms{17.23~ms} &
		&
		166.77 \bms{12.57~ms} & 111.57 1.98 s &
		199.34 \bms{17.23~ms} & 170.33 2.59 s &
		83.11 \bms{12.57~ms} & 74.41 2.62 s \\
		\hline
		\addlinespace
		\end{tabularx}

		\begin{tabular}{ccccccccccc}
		\rotatebox{90}{\small{Stand}}
		{\includegraphics[width=0.090\linewidth]{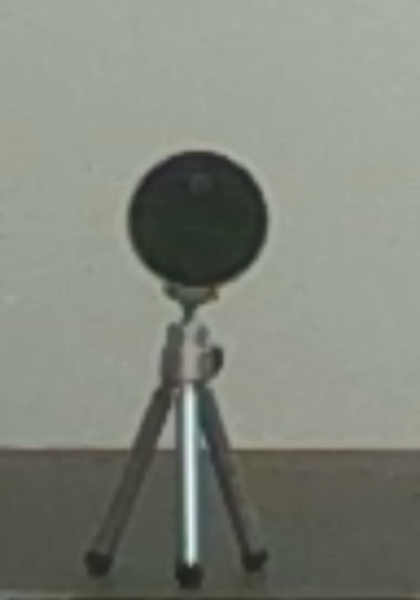}}
		{\includegraphics[width=0.090\linewidth]{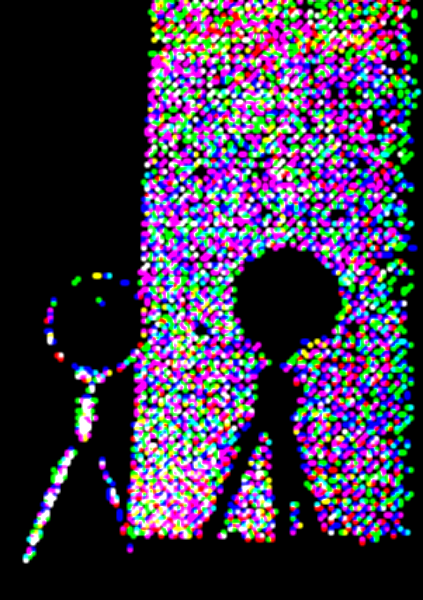}}
		{\includegraphics[width=0.090\linewidth]{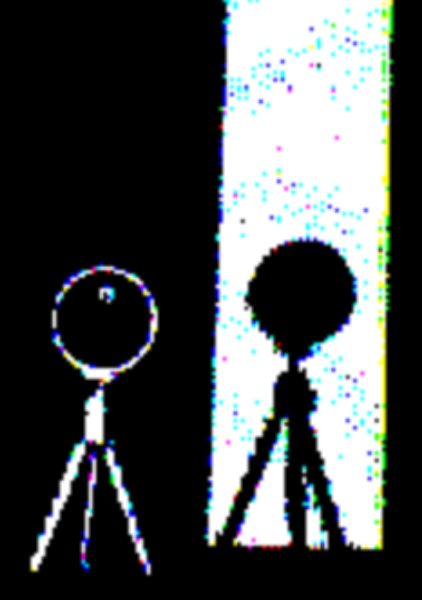}}
		{\includegraphics[width=0.090\linewidth]{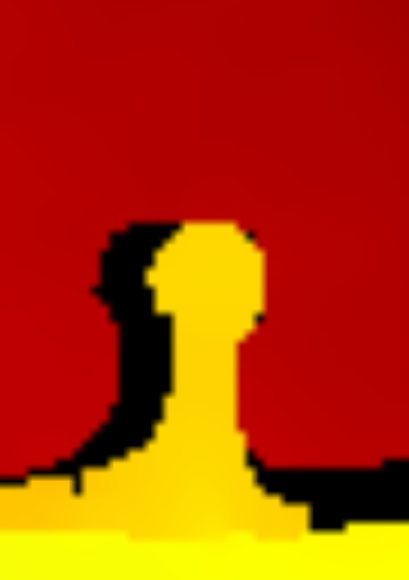}}
		{\includegraphics[width=0.090\linewidth]{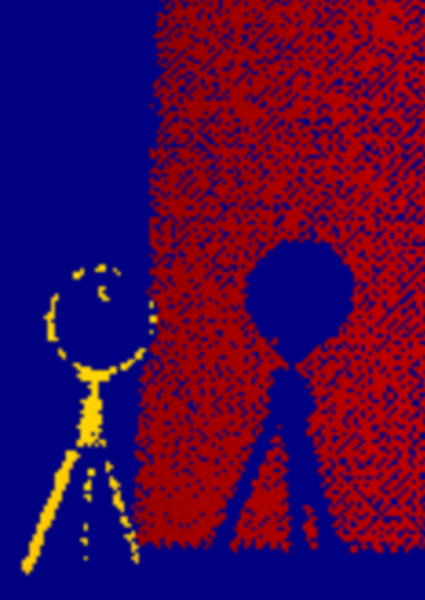}}
		{\includegraphics[width=0.090\linewidth]{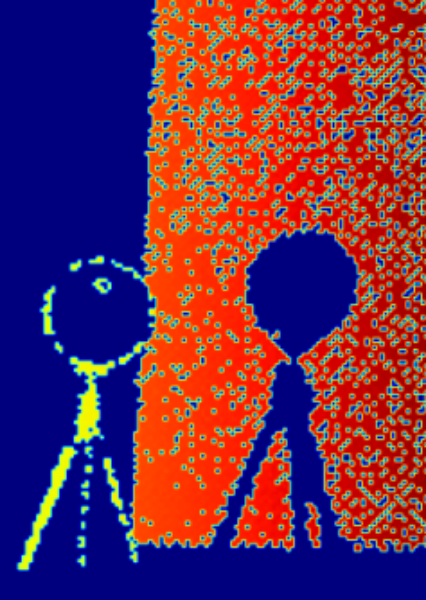}}
		{\includegraphics[width=0.090\linewidth]{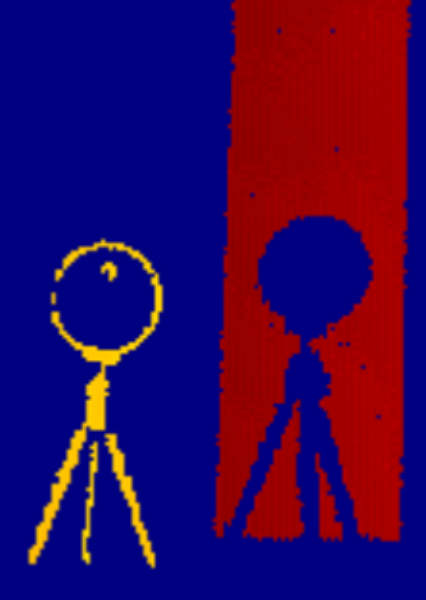}}
		{\includegraphics[width=0.090\linewidth]{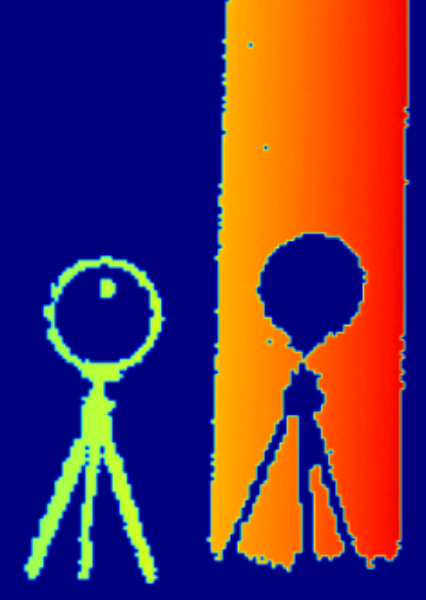}}
		{\includegraphics[width=0.090\linewidth]{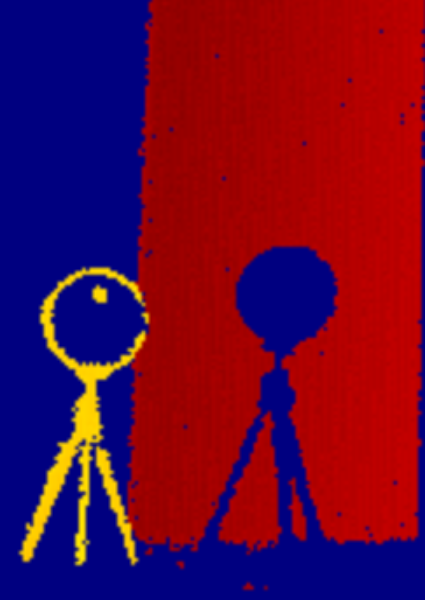}}
		{\includegraphics[width=0.090\linewidth]{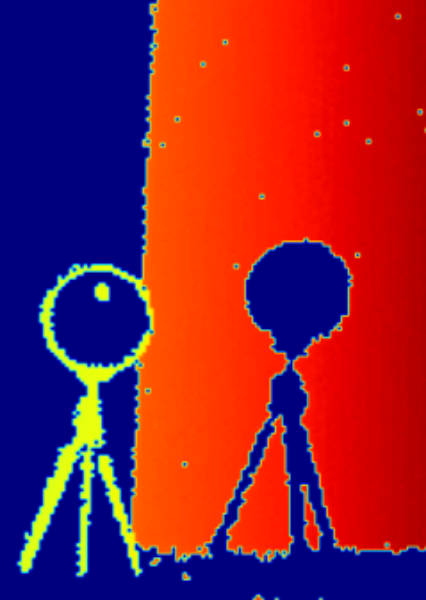}}
		\end{tabular}
		
		\begin{tabularx}{\linewidth}{p{0.001\linewidth}p{0.07\linewidth}p{0.075\linewidth}p{0.07\linewidth}p{0.07\linewidth}p{0.076\linewidth}p{0.066\linewidth}p{0.076\linewidth}p{0.066\linewidth}p{0.076\linewidth}p{0.066\linewidth}}
		& RMSE: Time: &
		20.46 \bms{12.57~ms} & 12.53 \bms{17.23~ms} &
		&
		181.56 \bms{12.57~ms} & 121.65 2.19 s &
		83.32 \bms{17.23~ms} & 100.22 1.60 s &
		85.25 \bms{12.57~ms} & 71.84 2.79 s \\
		\hline
		\end{tabularx}
		\caption{Comparison of color and depth detection for static scenes between RealSense D455, ESL, and E-RGB-D (ours). Noting that the available code for ESL needs to create a temporal map as a numpy array from a recorded raw file, we did not include the time required for these processes in our calculations. We only considered the time needed for calculating depth from those files. The operating system had an NVIDIA(R) GeForce RTX(TM) 2060 6GB GPU and an Intel Core i7-9750H CPU with 16GB of memory.}
		\label{fig:static_all}
\end{figure*}

\textbf{Static Environments:} To evaluate the outcome of the proposed method in
static situations, we have designed seven different setups (see
Figure~\ref{fig:static_all}). We utilized line and dot patterns in modes 3 and 4
(see Figure~\ref{fig:patterns}). It is necessary to mention that in mode 4, the
color and depth patterns are the same. However, in mode 3, different patterns
can be used (e.g., in mode 3, we utilized dot/line patterns to detect depth but
used dot patterns to detect color). To provide abbreviated names for each
pattern sequence, we used the following method: for example, 'M3L45' indicates
mode 3 with 45 lines in the pattern for depth detection.

Projecting 23 patterns, along with color patterns, in mode 3 takes 7.4 ms, while
projecting 23 patterns three times in mode 4 takes 17.23 ms. Projecting 45
patterns along with the color patterns in mode 3 takes 12.57 ms. To scan static
objects, we used the patterns M3L45, M3D45, and M4L23.

Each pattern falls into one of three categories: wide, normal, and dense. To
indicate the category, we used Coverage Percentage (CP). To calculate the CP, we
consider the area size between the first and last lines rather than the entire
projector plane. Otherwise, considering 45 lines dense or sparse would yield the
same CP. In general, a higher CP in our case indicates a denser pattern with a
smaller field of view (FOV), while a lower CP suggests a more sparse pattern
that covers a larger area with a higher FOV.

Our evaluation spanned three different speeds: 1, 26, and 58 fps. However, as
mentioned in Subsection~\ref{depth}, our system is an event-based model rather
than a frame-based one.

Detailed statistical analyses (e.g., Analysis of Variance (ANOVA) tables and raincloud figures) supporting these results are provided in the supplementary materials. For additional evaluation of color accuracy, including a Macbeth ColorChecker benchmark, we refer readers to our earlier work (see~\cite{Bajestani_2023_WACV}).

Figure~\ref{fig:static_duck} illustrates color and depth detection for the Duck setup using pattern M4L23. Our method achieves depth frame reconstruction about 90 times faster than the ESL method, while also simultaneously reconstructing color. Compared to RealSense, our method provides sharper and more accurate results. For a comprehensive comparison of all setups in a static scenario, refer to Figure~\ref{fig:static_all}.

\textbf{Dynamic Environments:}
To evaluate the outcome of the proposed method in dynamic situations, we have designed five different setups. To scan dynamic objects, we used the patterns M3D45, M3L23, and M3D23. We did not utilize pattern M4L23, although it has better output in terms of color in low-speed scanning. This is because considering the number of patterns that it needs to project, the total scanning speed could decrease, which is critical in dynamic conditions.

\begin{figure*}[htbp]
	\centering
	{\includegraphics[width=0.234\linewidth]{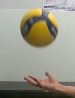}}
	\begin{tikzpicture}
		\node[inner sep=0pt] (img1) {\includegraphics[width=0.234\linewidth]{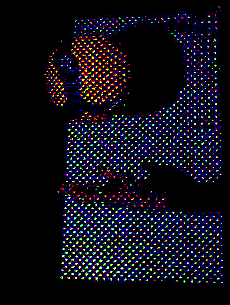}};
		\draw[orange, line width=2pt] (img1.north west) rectangle (img1.south east);
	\end{tikzpicture}
	\begin{tikzpicture}
		\node[inner sep=0pt] (img1) {\includegraphics[width=0.234\linewidth]{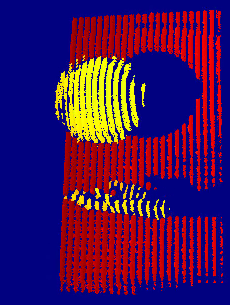}};
		\draw[green, line width=2pt] (img1.north west) rectangle (img1.south east);
	\end{tikzpicture}
	{\includegraphics[width=0.234\linewidth]{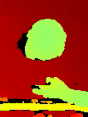}}
	\vspace{1pt}
	{\includegraphics[width=0.090\linewidth]{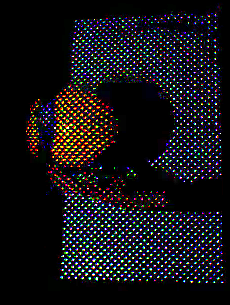}}
	{\includegraphics[width=0.090\linewidth]{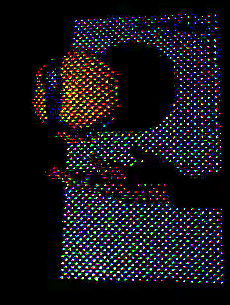}}
	{\includegraphics[width=0.090\linewidth]{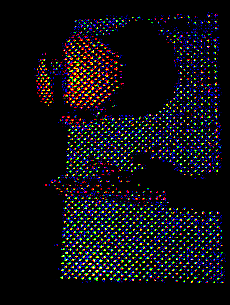}}
	{\includegraphics[width=0.090\linewidth]{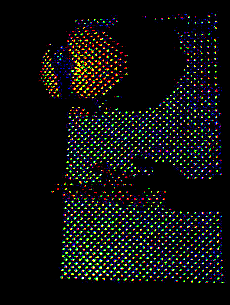}}
	{\includegraphics[width=0.090\linewidth]{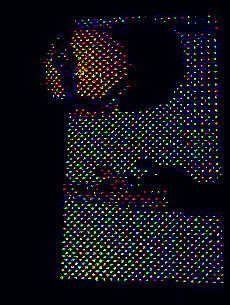}}
	\begin{tikzpicture}
		\node[inner sep=0pt] (img1) {\includegraphics[width=0.090\linewidth]{Ball_1_2_selected_bgr_06.png}};
		\draw[orange, line width=2pt] (img1.north west) rectangle (img1.south east);
	\end{tikzpicture}
	{\includegraphics[width=0.090\linewidth]{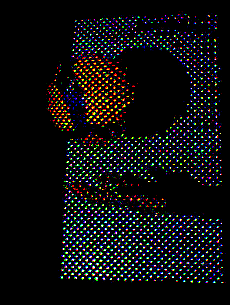}}
	{\includegraphics[width=0.090\linewidth]{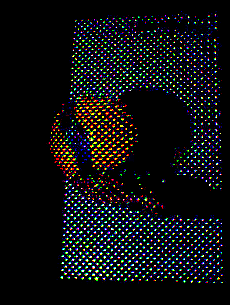}}
	{\includegraphics[width=0.090\linewidth]{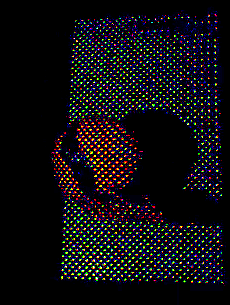}}
	{\includegraphics[width=0.090\linewidth]{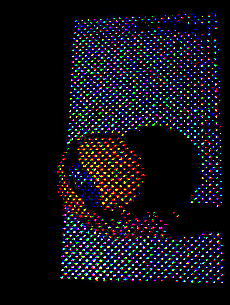}}
	\vspace{1pt}
	{\includegraphics[width=0.090\linewidth]{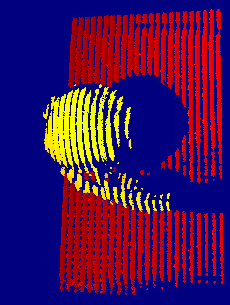}}
	{\includegraphics[width=0.090\linewidth]{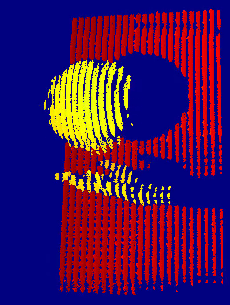}}
	{\includegraphics[width=0.090\linewidth]{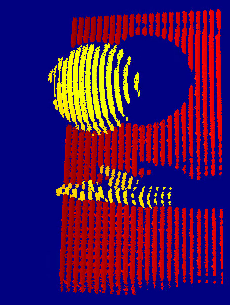}}
	{\includegraphics[width=0.090\linewidth]{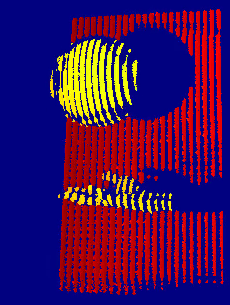}}
	{\includegraphics[width=0.090\linewidth]{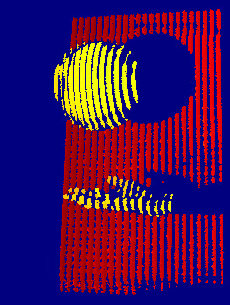}}
	\begin{tikzpicture}
		\node[inner sep=0pt] (img1) {\includegraphics[width=0.090\linewidth]{Ball_1_2_selected_depth_06.png}};
		\draw[green, line width=2pt] (img1.north west) rectangle (img1.south east);
	\end{tikzpicture}
	{\includegraphics[width=0.090\linewidth]{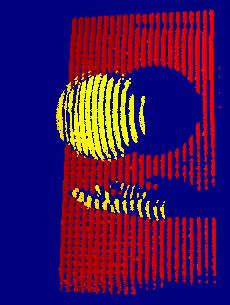}}
	{\includegraphics[width=0.090\linewidth]{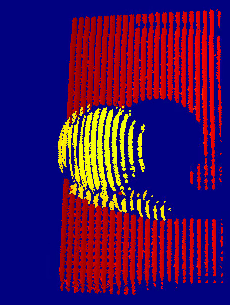}}
	{\includegraphics[width=0.090\linewidth]{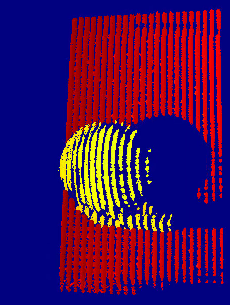}}
	{\includegraphics[width=0.090\linewidth]{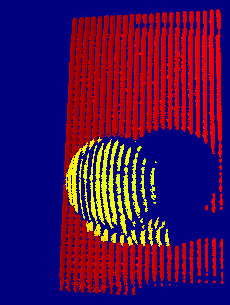}}

	\caption{Comparison of color and depth detection for dynamic scenes (setup with ball number one) between RealSense D455 (top row, right and left) and our system (middle, using pattern M3L23).}
	\label{fig:dynamic_ball}
\end{figure*}

Since there is no GT as mentioned in Subsection~\ref{baseline}, we could not provide FR or RMSE for dynamic experiments. Figure~\ref{fig:dynamic_ball} shows the color and depth detection of a volleyball ball being thrown up in front of the setup at a distance of approximately 1.5 m, using pattern M3L23. It also shows that the D455 sensor captured a slightly blurred RGB image, and the depth detection is not sharp and accurate.

\begin{figure*}[htbp]
	{
		{\includegraphics[width=0.090\linewidth]{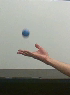}}
		{\includegraphics[width=0.090\linewidth]{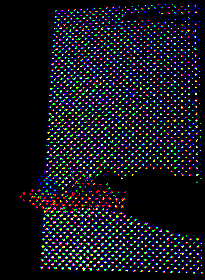}}
		{\includegraphics[width=0.090\linewidth]{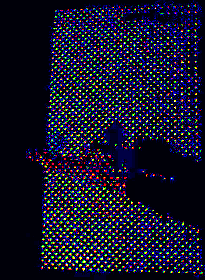}}
		{\includegraphics[width=0.090\linewidth]{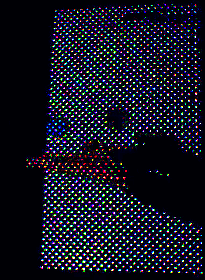}}
		{\includegraphics[width=0.090\linewidth]{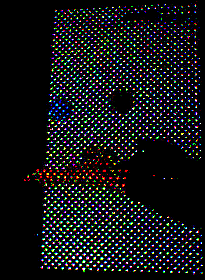}}
		{\includegraphics[width=0.090\linewidth]{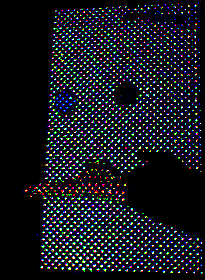}}
		{\includegraphics[width=0.090\linewidth]{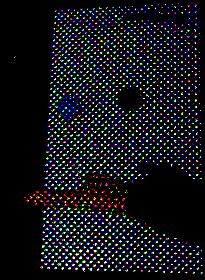}}
		{\includegraphics[width=0.090\linewidth]{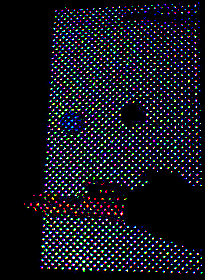}}
		{\includegraphics[width=0.090\linewidth]{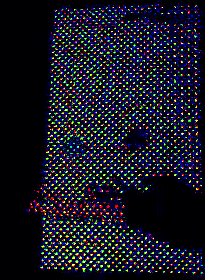}}
		{\includegraphics[width=0.090\linewidth]{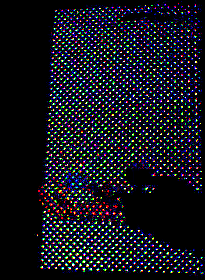}}
		
		{\includegraphics[width=0.090\linewidth]{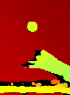}}
		{\includegraphics[width=0.090\linewidth]{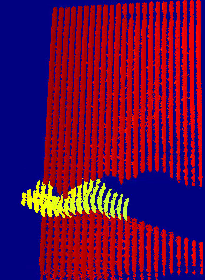}}
		{\includegraphics[width=0.090\linewidth]{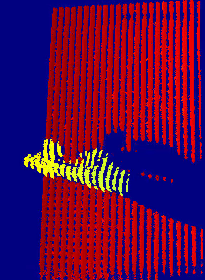}}
		{\includegraphics[width=0.090\linewidth]{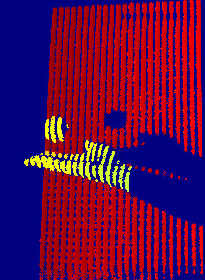}}
		{\includegraphics[width=0.090\linewidth]{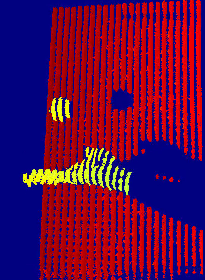}}
		{\includegraphics[width=0.090\linewidth]{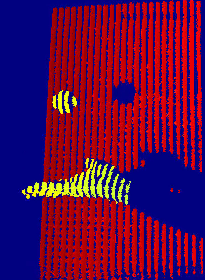}}
		{\includegraphics[width=0.090\linewidth]{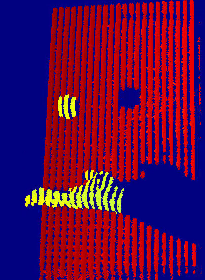}}
		{\includegraphics[width=0.090\linewidth]{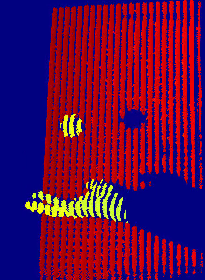}}
		{\includegraphics[width=0.090\linewidth]{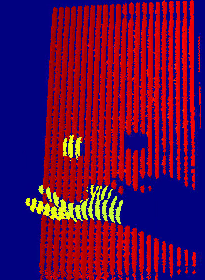}}
		{\includegraphics[width=0.090\linewidth]{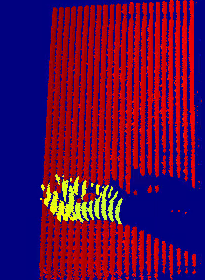}}
		
		{\includegraphics[width=0.090\linewidth]{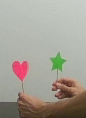}}
		{\includegraphics[width=0.090\linewidth]{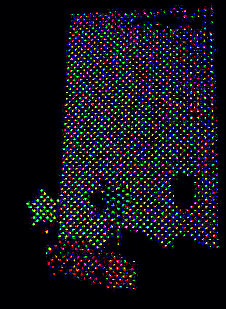}}
		{\includegraphics[width=0.090\linewidth]{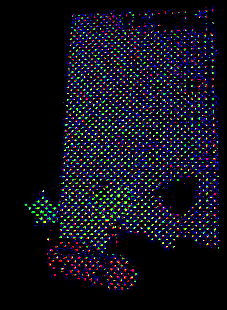}}
		{\includegraphics[width=0.090\linewidth]{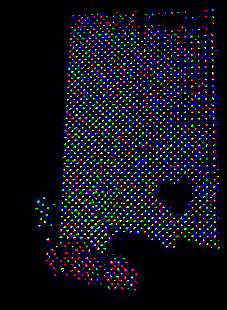}}
		{\includegraphics[width=0.090\linewidth]{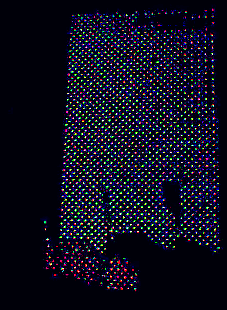}}
		{\includegraphics[width=0.090\linewidth]{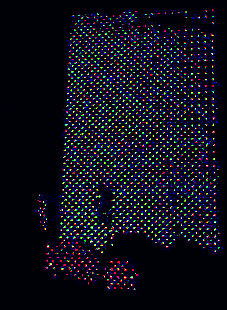}}
		{\includegraphics[width=0.090\linewidth]{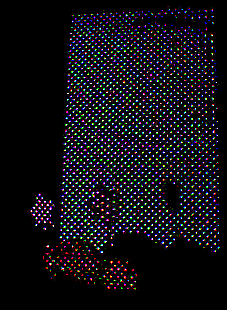}}
		{\includegraphics[width=0.090\linewidth]{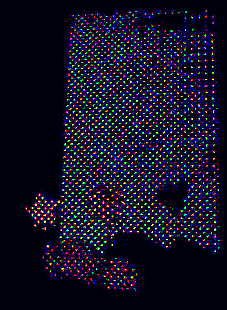}}
		{\includegraphics[width=0.090\linewidth]{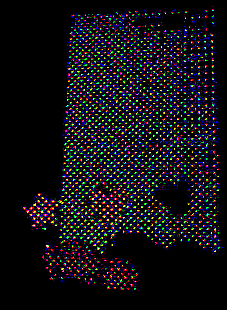}}
		{\includegraphics[width=0.090\linewidth]{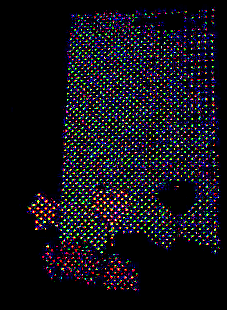}}
		
		{\includegraphics[width=0.090\linewidth]{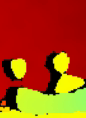}}
		{\includegraphics[width=0.090\linewidth]{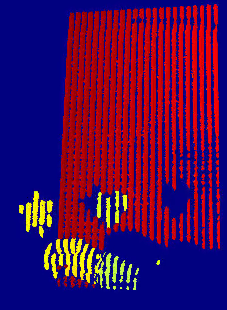}}
		{\includegraphics[width=0.090\linewidth]{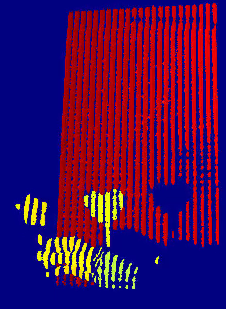}}
		{\includegraphics[width=0.090\linewidth]{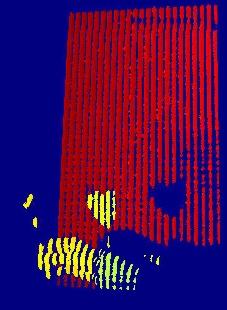}}
		{\includegraphics[width=0.090\linewidth]{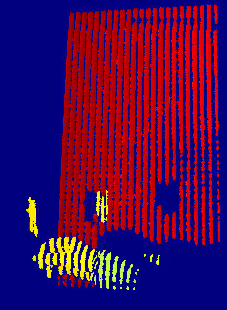}}
		{\includegraphics[width=0.090\linewidth]{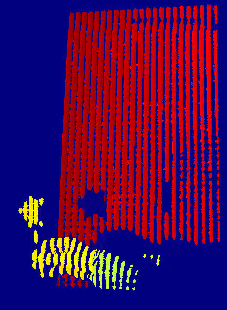}}
		{\includegraphics[width=0.090\linewidth]{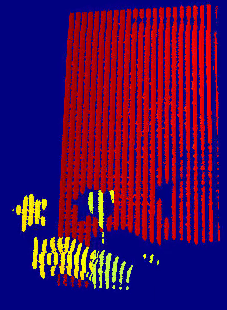}}
		{\includegraphics[width=0.090\linewidth]{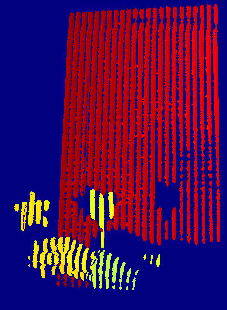}}
		{\includegraphics[width=0.090\linewidth]{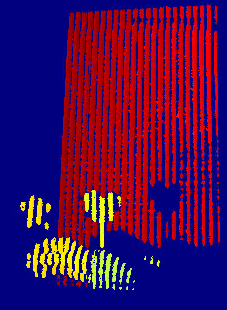}}
		{\includegraphics[width=0.090\linewidth]{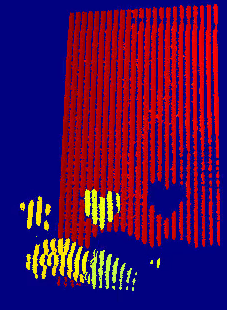}}
		
		{\includegraphics[width=0.090\linewidth]{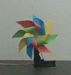}}
		{\includegraphics[width=0.090\linewidth]{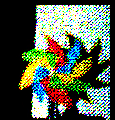}}
		{\includegraphics[width=0.090\linewidth]{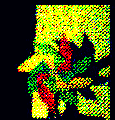}}
		{\includegraphics[width=0.090\linewidth]{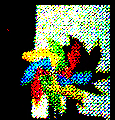}}
		{\includegraphics[width=0.090\linewidth]{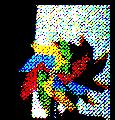}}
		{\includegraphics[width=0.090\linewidth]{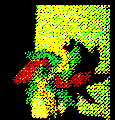}}
		{\includegraphics[width=0.090\linewidth]{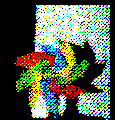}}
		{\includegraphics[width=0.090\linewidth]{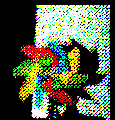}}
		{\includegraphics[width=0.090\linewidth]{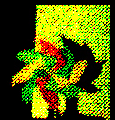}}
		{\includegraphics[width=0.090\linewidth]{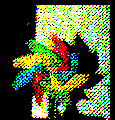}}
		
		{\includegraphics[width=0.090\linewidth]{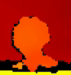}}
		{\includegraphics[width=0.090\linewidth]{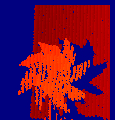}}
		{\includegraphics[width=0.090\linewidth]{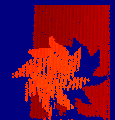}}
		{\includegraphics[width=0.090\linewidth]{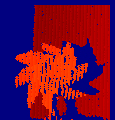}}
		{\includegraphics[width=0.090\linewidth]{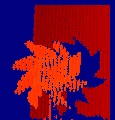}}
		{\includegraphics[width=0.090\linewidth]{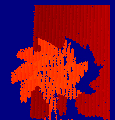}}
		{\includegraphics[width=0.090\linewidth]{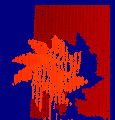}}
		{\includegraphics[width=0.090\linewidth]{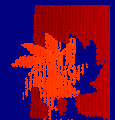}}
		{\includegraphics[width=0.090\linewidth]{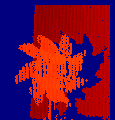}}
		{\includegraphics[width=0.090\linewidth]{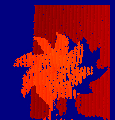}}
		
		{\includegraphics[width=0.090\linewidth]{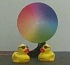}}
		{\includegraphics[width=0.090\linewidth]{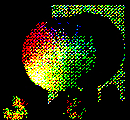}}
		{\includegraphics[width=0.090\linewidth]{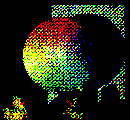}}
		{\includegraphics[width=0.090\linewidth]{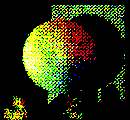}}
		{\includegraphics[width=0.090\linewidth]{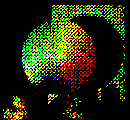}}
		{\includegraphics[width=0.090\linewidth]{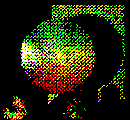}}
		{\includegraphics[width=0.090\linewidth]{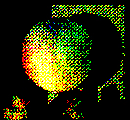}}
		{\includegraphics[width=0.090\linewidth]{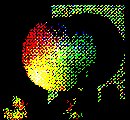}}
		{\includegraphics[width=0.090\linewidth]{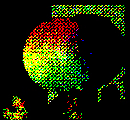}}
		{\includegraphics[width=0.090\linewidth]{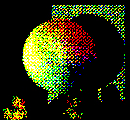}}
		
		{\includegraphics[width=0.090\linewidth]{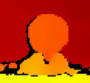}}
		{\includegraphics[width=0.090\linewidth]{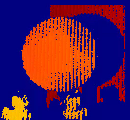}}
		{\includegraphics[width=0.090\linewidth]{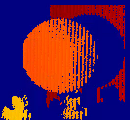}}
		{\includegraphics[width=0.090\linewidth]{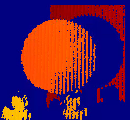}}
		{\includegraphics[width=0.090\linewidth]{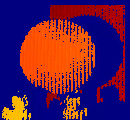}}
		{\includegraphics[width=0.090\linewidth]{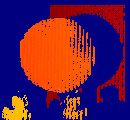}}
		{\includegraphics[width=0.090\linewidth]{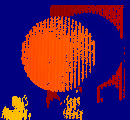}}
		{\includegraphics[width=0.090\linewidth]{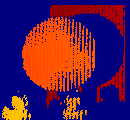}}
		{\includegraphics[width=0.090\linewidth]{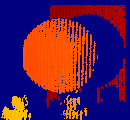}}
		{\includegraphics[width=0.090\linewidth]{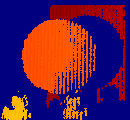}}
	}
	
	\caption{Series of images showing color and depth detection for dynamic scenes, scanned in 7.4 to 17.23 ms using the M3L23 pattern. The images in the first column from the left were captured by the RealSense D455 for comparison, which took 33 ms.}
	\label{fig:dynamic_all}
\end{figure*}

\section{Conclusion}\label{conclusion}
This study introduces a method for generating colored point clouds with adjustable speed and resolution, enabling depth map creation in challenging environments like dynamic and low-light conditions, even with stationary cameras or objects.

Results show that denser scanning patterns provide more accurate data for smaller fields of view (FOV), while sparser patterns cover larger areas. Using fewer patterns, we achieved a scanning time of 7.4 ms, significantly faster than the RealSense D455 (33 ms) and the ESL method (2 to 5 seconds). This method balances detail and speed effectively, offering improved speed control and enabling color reconstruction.

Using this setup, we achieved color scanning speeds up to 1.4~kHz (Mode 1, as investigated in our previous work) and pixel-based depth scanning speeds up to 4~kHz (Mode 2 with \( n=1 \)). This provides a continuous stream of annotated events with color and depth, along with detailed, colorful point cloud output. The method shows versatility across static and dynamic environments, introducing an innovative strategy for balancing resolution and acquisition speed.

Compared to prior methods that rely on frame accumulation or raster scanning, our approach offers a truly asynchronous pipeline with real-time event tagging. This allows color and depth sensing to occur independently of scene complexity or temporal batching, making it highly suitable for applications requiring low latency, high responsiveness, and continuous high-speed acquisition. This architectural advantage becomes especially critical in dynamic or resource-constrained environments.

An important feature of our system is its flexibility in structured light pattern design. While we demonstrate several effective patterns in this work, the system is capable of supporting custom encodings as long as they can be projected by the DLP hardware. Our software architecture is designed to accommodate such variations seamlessly, making the proposed method scalable and adaptable for a range of use cases, from mobile robotics to industrial metrology.

In light of the findings and outcomes of this study, several promising avenues for future research have emerged:
\begin{compactenum}
	\item \textbf{Optimizing resolution of color and depth detection} by projecting a denser pattern in a specific area could be achieved through the implementation of a movement detection algorithm during SL light downtime or by using a secondary camera to detect movement in parallel.
	\item \textbf{Increasing scanning speed} at the cost of higher bandwidth usage, while reducing resolution, can be achieved by projecting white-colored patterns when utilizing a color event camera (EC).
	\item \textbf{Optimizing depth measurement} based on the SL wavelength and the object's surface color by controlling the current of each LED of the projector.
	\item \textbf{Increasing the scanning range} by utilizing a Near-IR projector instead of capturing color.
	\item \textbf{Investigating predictive techniques} such as neighboring pixel tracking or learning-based interpolation to enhance the completeness of reconstructed color frames in regions with sparse event activity.
	
\end{compactenum}

\backmatter

\bmhead{Supplementary information}

Detailed statistical analyses supporting these findings, including Analysis of Variance (ANOVA) tables and raincloud figures, are presented in the supplementary materials, as mentioned in Subsection~\ref{results}. For further assessment of color accuracy, such as evaluations using a Macbeth ColorChecker benchmark, readers are directed to our previous study (\cite{Bajestani_2023_WACV}).

\bmhead{Acknowledgements}

This paper was produced by MIST Lab. thanks to the NSERC Discovery Grant 2019-05165, at the Department of Computer Engineering and Software Engineering, Polytechnique Montreal, Montreal, CA.


\bibliography{ergbd-ijcv-bibliography.bib}

@inproceedings{Bajestani_2023_WACV,
	author    = {Marjani-Bajestani, Seyed-Ehsan and Beltrame, Giovanni},
	title     = {Event-Based {RGB} Sensing With Structured Light},
	booktitle = {Proceedings of the {IEEE/CVF} Winter Conference on Applications of Computer Vision ({WACV})},
	year      = {2023},
	pages     = {5458--5467},
	url		  = {https://doi.org/10.1109/WACV56688.2023.00542}
}

@inproceedings{jason2011dlp,
	author = {Jason Geng},
	title = {{DLP-based structured light 3D imaging technologies and applications}},
	volume = {7932},
	booktitle = {Emerging Digital Micromirror Device Based Systems and Applications III},
	editor = {Michael R. Douglass and Patrick I. Oden},
	organization = {International Society for Optics and Photonics},
	publisher = {SPIE},
	pages = {79320B},
	year = {2011},
	URL = {https://doi.org/10.1117/12.873125}
}

@misc{helios,
	author       = {{LUCID Vision Labs}},
	title        = {{Helios Time-of-Flight (ToF) Camera}},
	year         = {2025},
	url          = {https://thinklucid.com/helios-time-of-flight-tof-camera/},
	note         = {Accessed: 2025-11-24}
}

@article{morar2020comprehensive,
	title={{A Comprehensive Survey of Indoor Localization Methods Based on Computer Vision}},
	author={Morar, Anca and Moldoveanu, Alin and Mocanu, Irina and Moldoveanu, Florica and Radoi, Ion Emilian and Asavei, Victor and Gradinaru, Alexandru and Butean, Alex},
	journal={Sensors},
	volume={20},
	number={9},
	pages={2641},
	year={2020},
	publisher={Multidisciplinary Digital Publishing Institute},
	url={https://doi.org/10.3390/s20092641}
}

@inproceedings{matsuda2015mc3d,
	author = {Matsuda, Nathan and Cossairt, Oliver and Gupta, Mohit},
	title = {MC3D: Motion Contrast 3D Scanning},
	volume = {},
	booktitle = {2015 IEEE International Conference on Computational Photography (ICCP)}, 
	organization = {IEEE},
	publisher = {IEEE},
	pages = {1--10},
	year = {2015},
	URL = {https://doi.org/10.1109/ICCPHOT.2015.7168370}
}

@inproceedings{gupta2012micro,
	title={Micro phase shifting},
	author={Gupta, Mohit and Nayar, Shree K},
	booktitle={2012 {IEEE} Conference on Computer Vision and Pattern Recognition},
	pages={813--820},
	year={2012},
	organization={IEEE},
	url={https://doi.org/10.1109/CVPR.2012.6247753}
}

@article{wu2020exposure,
	title={{An exposure fusion-based structured light approach for the 3D measurement of a specular surface}},
	author={Wu, Ke and Tan, Jie and Xia, Hailun and Liu, Chengbao},
	journal={{IEEE} Sensors Journal},
	year={2020},
	publisher={IEEE},
	url={https://doi.org/10.1109/JSEN.2020.3027317}
}

@inproceedings{tin20163d,
	title={{3D} reconstruction of mirror-type objects using efficient ray coding},
	author={Tin, Siu-Kei and Ye, Jinwei and Nezamabadi, Mahdi and Chen, Can},
	booktitle={2016 {IEEE} International Conference on Computational Photography ({ICCP})},
	pages={1--11},
	year={2016},
	organization={IEEE},
	url={https://doi.org/10.1109/ICCPHOT.2016.7492867}
}

@article{li2024learning,
	title={{Learning from General Diffuse Surfaces: An Event-driven Approach for High Dynamic Range Industrial Optical Measurement}},
	author={Li, Yuhui and Xu, Chen and Liu, Lilin},
	journal={Optics \& Laser Technology},
	volume={177},
	pages={111183},
	year={2024},
	publisher={Elsevier},
	url={https://doi.org/10.1016/j.optlastec.2024.111183}
}

@article{brandli2014adaptive,
	title={Adaptive pulsed laser line extraction for terrain reconstruction using a dynamic vision sensor},
	author={Brandli, Christian and Mantel, Thomas and Hutter, Marco and H{\"o}pflinger, Markus and Berner, Raphael and Siegwart, Roland and Delbruck, Tobi},
	journal={Frontiers in neuroscience},
	volume={7},
	pages={275},
	year={2014},
	publisher={Frontiers},
	url = {https://doi.org/10.3389/fnins.2013.00275}
}

@inproceedings{muglikar2021esl,
	title={{ESL}: Event-based Structured Light},
	author={Muglikar, Manasi and Gallego, Guillermo and Scaramuzza, Davide},
	booktitle={2021 International Conference on 3D Vision ({3DV})},
	pages={1165--1174},
	year={2021},
	organization={IEEE},
	url={https://doi.org/10.1109/3DV53792.2021.00124}
}

@inproceedings{morgenstern2023x,
	title={{X-maps}: Direct Depth Lookup for Event-based Structured Light Systems},
	author={Morgenstern, Wieland and Gard, Niklas and Baumann, Simon and Hilsmann, Anna and Eisert, Peter},
	booktitle={2023 {IEEE/CVF} Conference on Computer Vision and Pattern Recognition Workshops ({CVPRW})},
	pages={4006--4014},
	year={2023},
	url={https://doi.org/10.1109/CVPRW59228.2023.00418}
}

@article{tremeau2008color,
	title={Color in image and video processing: most recent trends and future research directions},
	author={Trémeau, Alain and Tominaga, Shoji and Plataniotis, KonstantinosN},
	journal={{EURASIP Journal on Image and Video Processing}},
	volume={2008},
	pages={1--26},
	year={2008},
	publisher={Springer},
	doi={10.1155/2008/581371}
}

@article{jang2020deep,
	title={{Deep Color Transfer for Color-Plus-Mono Dual Cameras}},
	author={Jang, Hae Woong and Jung, Yong Ju},
	journal={Sensors},
	volume={20},
	number={9},
	pages={2743},
	year={2020},
	publisher={Multidisciplinary Digital Publishing Institute},
	url={https://doi.org/10.3390/s20092743}
}

@inproceedings{levin2004colorization,
	title={Colorization using optimization},
	author={Levin, Anat and Lischinski, Dani and Weiss, Yair},
	booktitle={ACM SIGGRAPH 2004 Papers},
	pages={689--694},
	year={2004},
	url={https://doi.org/10.1145/1186562.1015780}
}

@inproceedings{zhang2016colorful,
	title={Colorful Image Colorization},
	author={Zhang, Richard and Isola, Phillip and Efros, Alexei A},
	booktitle={European conference on computer vision},
	pages={649--666},
	year={2016},
	organization={Springer},
	url={https://link.springer.com/chapter/10.1007/978-3-319-46487-9_40},
	doi={10.1007/978-3-319-46487-9_40}
}

@article{cohen2024colorful,
	title={Colorful image reconstruction from neuromorphic event cameras with biologically inspired deep color fusion neural networks},
	author={Cohen Duwek, Hadar and Ezra Tsur, Elishai},
	journal={Bioinspiration \& Biomimetics},
	year = {2024},
	month = {mar},
	publisher = {IOP Publishing},
	volume = {19},
	number = {3},
	pages = {036001},
	url = {https://doi.org/10.1088/1748-3190/ad2a7c},
}

@inproceedings{li2015design,
	title={{Design of an RGBW color VGA rolling and global shutter dynamic and active-pixel vision sensor}},
	author={Li, Chenghan and Brandli, Christian and Berner, Raphael and Liu, Hongjie and Yang, Minhao and Liu, Shih-Chii and Delbruck, Tobi},
	booktitle={2015 {IEEE} International Symposium on Circuits and Systems ({ISCAS})},
	pages={718--721},
	year={2015},
	organization={IEEE},
	url={https://doi.org/10.1109/ISCAS.2015.7168734}
}

@article{moeys2017sensitive,
	title={{A Sensitive Dynamic and Active Pixel Vision Sensor for Color or Neural Imaging Applications}},
	author={Moeys, Diederik Paul and Corradi, Federico and Li, Chenghan and Bamford, Simeon A and Longinotti, Luca and Voigt, Fabian F and Berry, Stewart and Taverni, Gemma and Helmchen, Fritjof and Delbruck, Tobi},
	journal={{IEEE} transactions on biomedical circuits and systems},
	volume={12},
	number={1},
	pages={123--136},
	year={2017},
	publisher={IEEE},
	url={https://doi.org/10.1109/TBCAS.2017.2759783}
}

@inproceedings{moeys2017color,
	title={Color temporal contrast sensitivity in dynamic vision sensors},
	author={Moeys, Diederik Paul and Li, Chenghan and Martel, Julien NP and Bamford, Simeon and Longinotti, Luca and Motsnyi, Vasyl and Bello, David San Segundo and Delbruck, Tobi},
	booktitle={2017 {IEEE} International Symposium on Circuits and Systems (ISCAS)},
	pages={1--4},
	year={2017},
	organization={IEEE},
	url={https://doi.org/10.1109/ISCAS.2017.8050412}
}

@article{bayer1976color,
	title={Color imaging array},
	author={Bayer, Bryce E},
	journal={United States Patent 3,971,065},
	year={1976}
}

@article{khashabi2014joint,
	title={{Joint Demosaicing and Denoising via Learned Nonparametric Random Fields}},
	author={Khashabi, Daniel and Nowozin, Sebastian and Jancsary, Jeremy and Fitzgibbon, Andrew W},
	journal={{IEEE} Transactions on Image Processing},
	volume={23},
	number={12},
	pages={4968--4981},
	year={2014},
	publisher={IEEE},
	url={https://doi.org/10.1109/TIP.2014.2359774}
}

@article{ramanath2005color,
	title={Color image processing pipeline},
	author={Ramanath, Rajeev and Snyder, Wesley E and Yoo, Youngjun and Drew, Mark S},
	journal={{IEEE} Signal Processing Magazine},
	volume={22},
	number={1},
	pages={34--43},
	year={2005},
	publisher={IEEE},
	url={https://doi.org/10.1109/MSP.2005.1407713}
}

@article{marcireau2018event,
	title={{Event-Based Color Segmentation With a High Dynamic Range Sensor}},
	author={Marcireau, Alexandre and Ieng, Sio-Hoi and Simon-Chane, Camille and Benosman, Ryad B},
	journal={Frontiers in neuroscience},
	volume={12},
	pages={135},
	year={2018},
	publisher={Frontiers},
	url={https://doi.org/10.3389/fnins.2018.00135}
}

@article{marrugo2020state,
	title={State-of-the-art active optical techniques for three-dimensional surface metrology: a review},
	author={Marrugo, Andres G and Gao, Feng and Zhang, Song},
	journal={{J. Opt. Soc. Am. A (JOSA A)}},
	volume={37},
	number={9},
	pages={B60--B77},
	year={2020},
	publisher={Optical Society of America},
	url={http://doi.org/10.1364/JOSAA.398644}
}

@inproceedings{martel2018active,
	title={{An Active Approach to Solving the Stereo Matching Problem using Event-Based Sensors}},
	author={Martel, Julien NP and M{\"u}ller, Jonathan and Conradt, J{\"o}rg and Sandamirskaya, Yulia},
	booktitle={2018 {IEEE} International Symposium on Circuits and Systems ({ISCAS})},
	pages={1--5},
	year={2018},
	organization={IEEE},
	url={https://doi.org/10.1109/ISCAS.2018.8351411}
}

@inproceedings{weikersdorfer2014event,
	title={{Event-based 3D SLAM with a depth-augmented dynamic vision sensor}},
	author={Weikersdorfer, David and Adrian, David B and Cremers, Daniel and Conradt, J{\"o}rg},
	booktitle={2014 {IEEE} International Conference on Robotics and Automation ({ICRA})},
	pages={359--364},
	year={2014},
	organization={IEEE},
	url={https://doi.org/10.1109/ICRA.2014.6906882}
}

@article{willomitzer2017single,
	title={{Single-shot 3D motion picture camera with a dense point cloud}},
	author={Willomitzer, Florian and H{\"a}usler, Gerd},
	journal={Optics express},
	volume={25},
	number={19},
	pages={23451--23464},
	year={2017},
	publisher={Optical Society of America},
	url = {http://doi.org/10.1364/OE.25.023451}
}

@inproceedings{zhao2019miniature,
	title={{Miniature 3D Depth Camera for Real-time Reconstruction}},
	author={Zhao, Zilong and Gu, Feifei and Xie, Pengju and Cao, Huazhao and Song, Zhan},
	booktitle={2019 {IEEE} International Conference on Robotics and Biomimetics ({ROBIO})},
	pages={1769--1776},
	year={2019},
	organization={IEEE},
	url={https://doi.org/10.1109/ROBIO49542.2019.8961795}
}

@article{leroux2018event,
	title={Event-based structured light for depth reconstruction using frequency tagged light patterns},
	author={Leroux, T and Ieng, S-H and Benosman, Ryad},
	journal={arXiv preprint arXiv:1811.10771},
	year={2018},
	url={https://doi.org/10.48550/arXiv.1811.10771}
}

@article{mangalore2020neuromorphic,
	title={{Neuromorphic Fringe Projection Profilometry}},
	author={Mangalore, Ashish Rao and Seelamantula, Chandra Sekhar and Thakur, Chetan Singh},
	journal={{IEEE} Signal Processing Letters},
	volume={27},
	pages={1510--1514},
	year={2020},
	publisher={IEEE},
	url={https://doi.org/10.1109/LSP.2020.3016251}
}

@article{li2024event,
	title={{Event-driven Fringe Projection Structured Light 3D Reconstruction based on Time-frequency Analysis}},
	author={Li, Yuhui and Jiang, Heng and Xu, Chen and Liu, Lilin},
	journal={{IEEE} Sensors Journal},
	year={2024},
	publisher={IEEE},
	url={https://doi.org/10.1109/JSEN.2024.3349432}
}

@article{lu2024sge,
	title={{SGE}: Structured Light System Based on Gray Code with an Event Camera},
	author={Lu, Xingyu and Sun, Lei and Gu, Diyang and Xu, Zhijie and Wang, Kaiwei},
	journal={arXiv preprint arXiv:2403.07326},
	year={2024},
	url={https://doi.org/10.48550/arXiv.2403.07326}
}

@inproceedings{wang2022enhancing,
	title={{Enhancing Event-based Structured Light Imaging with a Single Frame}},
	author={Wang, Huijiao and Liu, Tangbo and He, Chu and Li, Cheng and Liu, Jianzhuang and Yu, Lei},
	booktitle={2022 {IEEE} International Conference on Multisensor Fusion and Integration for Intelligent Systems ({MFI})},
	pages={1--7},
	year={2022},
	organization={IEEE},
	url={https://doi.org/10.1109/MFI55806.2022.9913845}
}

@article{yu2020high,
	title={{High Sensitivity Fringe Projection Profilometry Combining Optimal Fringe Frequency and Optimal Fringe Direction}},
	author={Yu, Jin and Gao, Nan and Zhang, Zonghua and Meng, Zhaozong},
	journal={Optics and Lasers in Engineering},
	volume={129},
	pages={106068},
	year={2020},
	publisher={Elsevier},
	url={http://doi.org/10.1016/j.optlaseng.2020.106068}
}

@misc{kinectv1,
	author = {{Microsoft Kinect}},
	title = {Microsoft Kinect for Xbox 360},
	year = {2010},
	url = {https://www.microsoft.com/en-us/download/details.aspx?id=40278},
	note = {Accessed: 2025-12-01}
}

@misc{intel3d,
	author = {{Intel RealSense}},
	title = {Stereo depth solutions},
	year = {2025},
	url = {https://realsenseai.com/},
	note = {Accessed: 2025-11-23}
}

@misc{orbbec3d,
	author = {{Orbbec, Astra Series}},
	title = {Intelligent computing for everyone, everywhere},
	year = {2025},
	url = {https://www.orbbec.com/products/structured-light-camera/astra-2/},
	note   = {Accessed: 2025-11-23}
}

@article{huang2021high,
	title={{High-speed structured light based 3D scanning using an event camera}},
	author={Huang, Xueyan and Zhang, Yueyi and Xiong, Zhiwei},
	journal={Optics Express},
	volume={29},
	number={22},
	pages={35864--35876},
	year={2021},
	publisher={Optica Publishing Group},
	url={https://doi.org/10.1364/OE.437944}
}

@inproceedings{gupta2013structured,
	title={Structured light in sunlight},
	author={Gupta, Mohit and Yin, Qi and Nayar, Shree K},
	booktitle={2013 {IEEE} International Conference on Computer Vision ({ICCV})},
	pages={545--552},
	year={2013},
	url={https://doi.org/10.1109/ICCV.2013.73}
}

@misc{prophesee,
	author       = {{Prophesee}},
	title        = {Evaluation Kit},
	year         = {2024},
	url          = {https://www.prophesee.ai/event-based-evk/},
	note         = {Accessed: 2025-01-15}
}

@misc{dlpLC4500,
	author       = {{Texas Instruments}},
	title        = {Digital Light Processing LightCrafter 4500},
	year         = {2024},
	url          = {https://www.ti.com/tool/DLPLCR4500EVM},
	note         = {Accessed: 2025-01-15}
}

@article{opencv_library,
	author = {Bradski, G.},
	journal = {Dr. Dobb's Journal of Software Tools},
	title = {{The OpenCV Library}},
	year = {2000},
	url={https://github.com/opencv}
}

@book{learning_opencv3,
	title={{Learning OpenCV 3}},
	author={Kaehler, Adrian and Bradski, Gary},
	year={2017},
	publisher={O'Reilly Media},
	isbn = {978-1491937990},
	note = {ISBN-10: 1491937998},
	url={https://www.oreilly.com/library/view/learning-opencv-3/9781491937983/}
}

@article{luo2014simple,
	title={A simple calibration procedure for structured light system},
	author={Luo, Huafen and Xu, Jing and Binh, Nguyen Hoa and Liu, Shuntao and Zhang, Chi and Chen, Ken},
	journal={Optics and Lasers in Engineering},
	volume={57},
	pages={6--12},
	year={2014},
	publisher={Elsevier},
	url={http://doi.org/10.1016/j.optlaseng.2014.01.010}
}

@article{wang2020temporal,
	title={{Temporal Matrices Mapping Based Calibration Method for Event-Driven Structured Light Systems}},
	author={Wang, Guijin and Feng, Chenchen and Hu, Xiaowei and Yang, Huazhong},
	journal={{IEEE} Sensors Journal},
	year={2020},
	publisher={IEEE},
	url={https://doi.org/10.1109/JSEN.2020.3016833}
}

@article{allen2019raincloud,
	title={Raincloud plots: a multi-platform tool for robust data visualization},
	author={Allen, Micah and Poggiali, Davide and Whitaker, Kirstie and Marshall, Tom Rhys and van Langen, Jordy and Kievit, Rogier A},
	journal={Wellcome open research},
	volume={4},
	year={2019},
	publisher={The Wellcome Trust},
	url={https://doi.org/10.12688/wellcomeopenres.15191.2}
}

@misc{jasp2024,
	author = {{JASP Team}},
	title = {{JASP (Version 0.18.3)}},
	year = {2024},
	howpublished = {Computer software},
	url = {https://jasp-stats.org/}
}

@inproceedings{Bajestani_2025_RoboCup,
	author    = {Marjani-Bajestani, Seyed-Ehsan and Beltrame, Giovanni},
	title     = {{Event-Based Vision for Robot Soccer}},
	booktitle = {{RoboCup 2024: Robot World Cup XXVII}},
	editor    = {Barros, Edna and Hanna, Josiah P. and Okada, Hiroyuki and Torta, Elena},
	publisher = {Springer Nature Switzerland},
	address   = {Cham},
	year      = {2025},
	month     = {April},
	pages     = {309--317},
	url       = {https://doi.org/10.1007/978-3-031-85859-8_26},
	isbn      = {978-3-031-85859-8}
}

\clearpage

\appendix

\section{Supplementary Materials}\label{secA1}

\begin{figure}[H]
	\centering
	\includegraphics[width=0.97\linewidth]{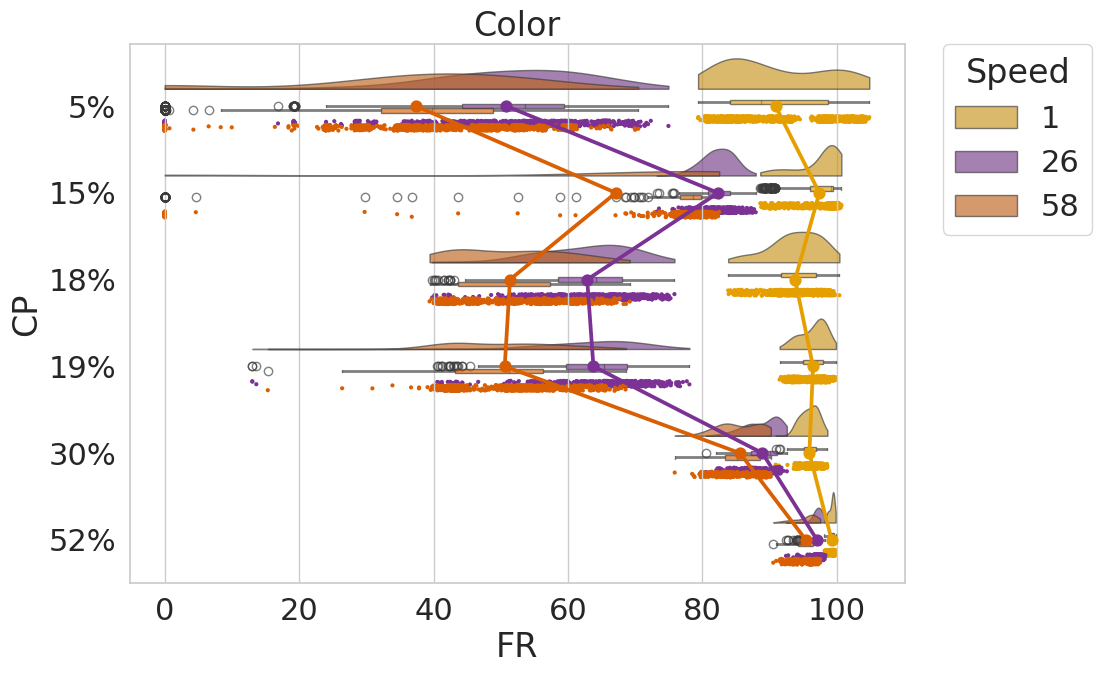}
	\includegraphics[width=0.97\linewidth]{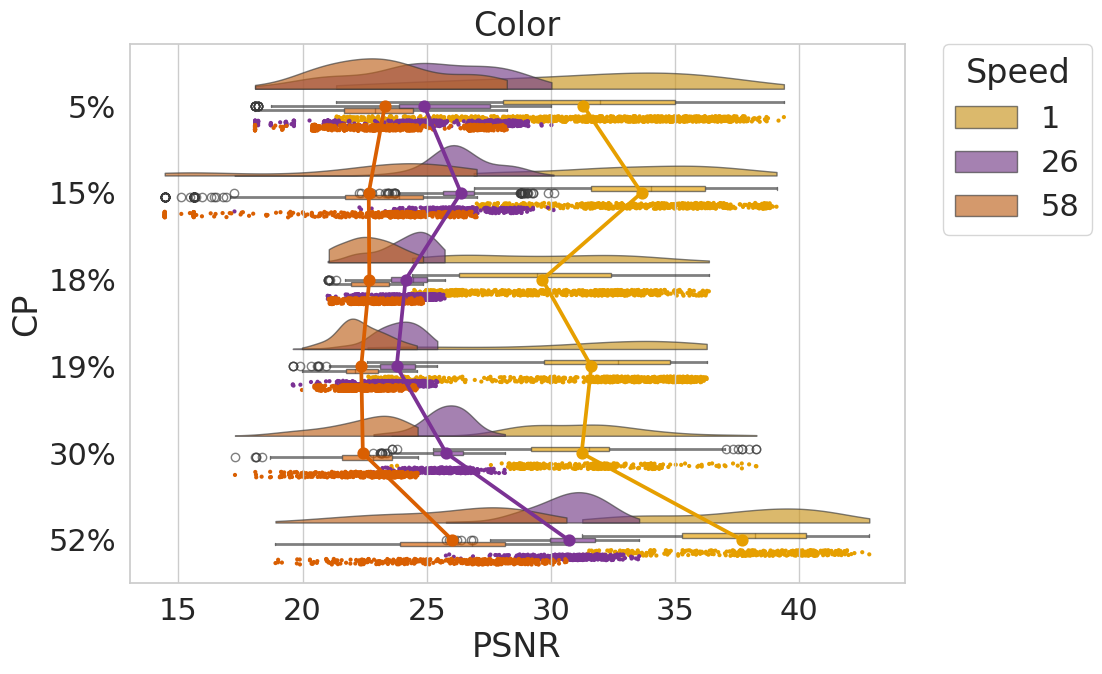}
	\includegraphics[width=0.97\linewidth]{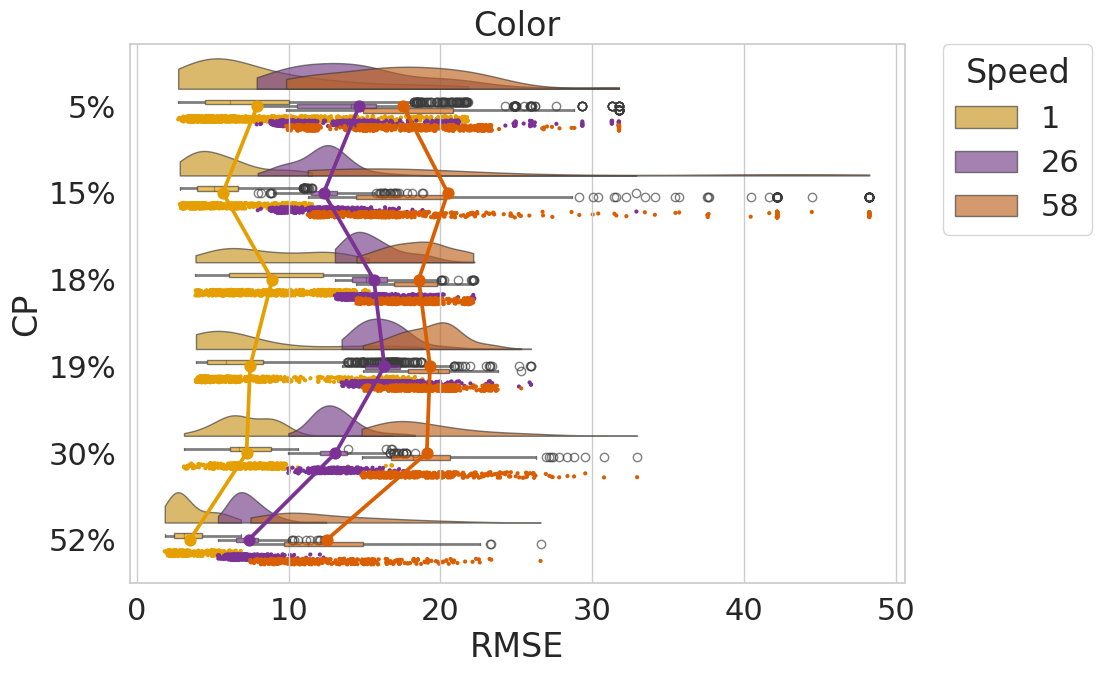}
	\caption{Comparison of color detection for all setups at different speeds.}
	\label{fig:static_raincloud-bgr}
\end{figure}

\begin{figure}[tbp]
	\centering
	\includegraphics[width=0.95\linewidth]{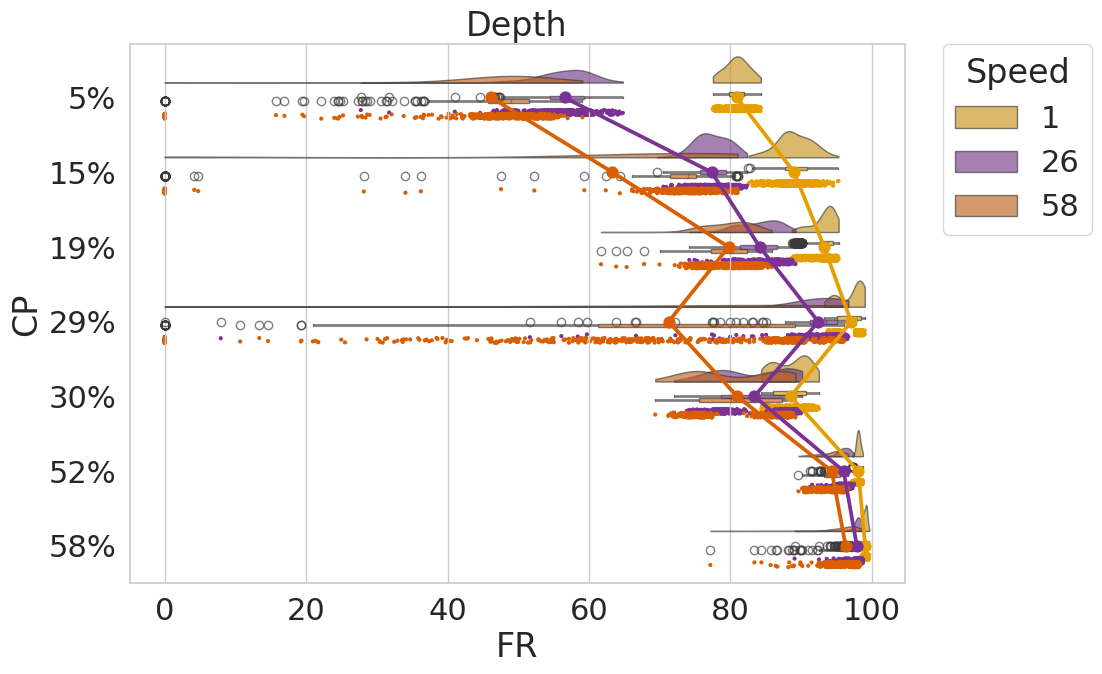}
	\includegraphics[width=0.95\linewidth]{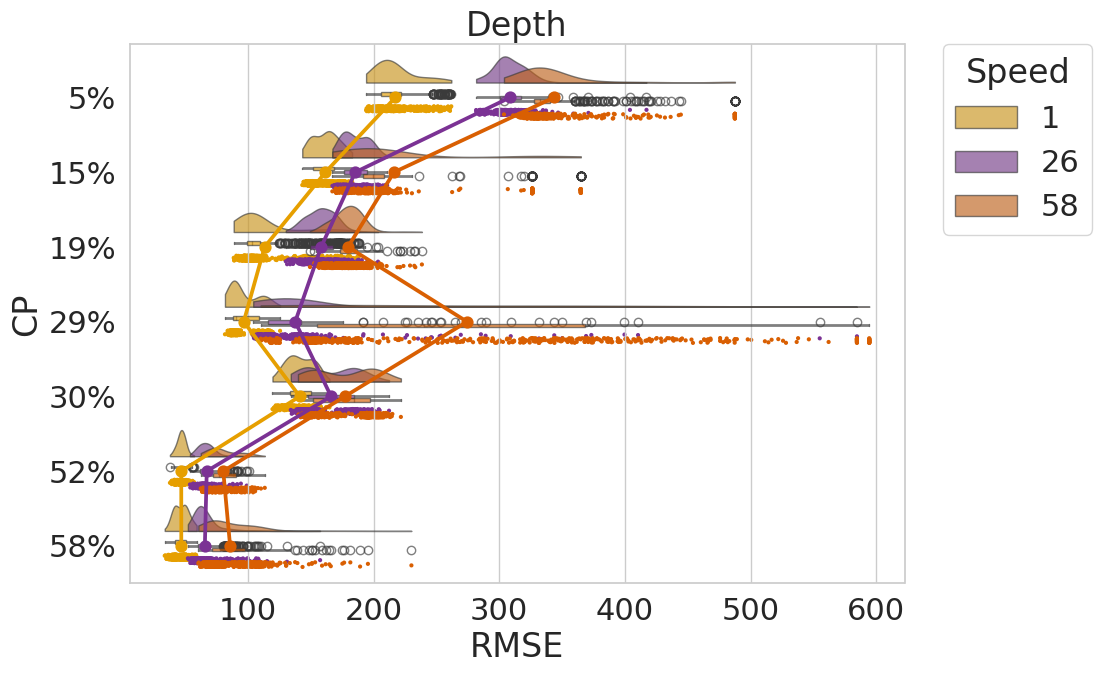}
	\caption{Comparison of depth detection for all setups at different speeds.}
	\label{fig:static_raincloud-depth}
\end{figure}

Figures~\ref{fig:static_raincloud-bgr} and~\ref{fig:static_raincloud-depth} are	raincloud plots~\cite{allen2019raincloud} of accuracy metrics for color detection (PNSR, RMSE, FR) over more than 25k
frames of data.

\begin{figure}[tbp]
	\centering
	\begin{tabularx}{\linewidth}{p{0.1\linewidth}p{0.28\linewidth}p{0.28\linewidth}p{0.28\linewidth}}
		& \small PSNR & \small RMSE & \small FR\\
	\end{tabularx}	
	\begin{tabular}{cccc}
		\small\rotatebox{90}{Standardized Residuals} &
		{\includegraphics[width=0.25\linewidth]{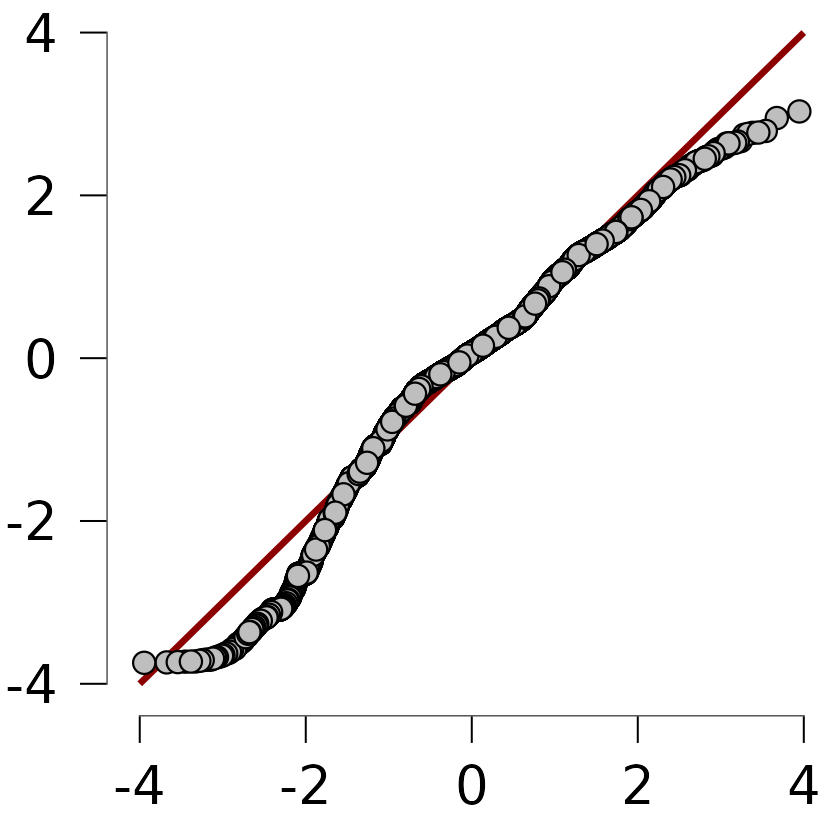}} &
		{\includegraphics[width=0.25\linewidth]{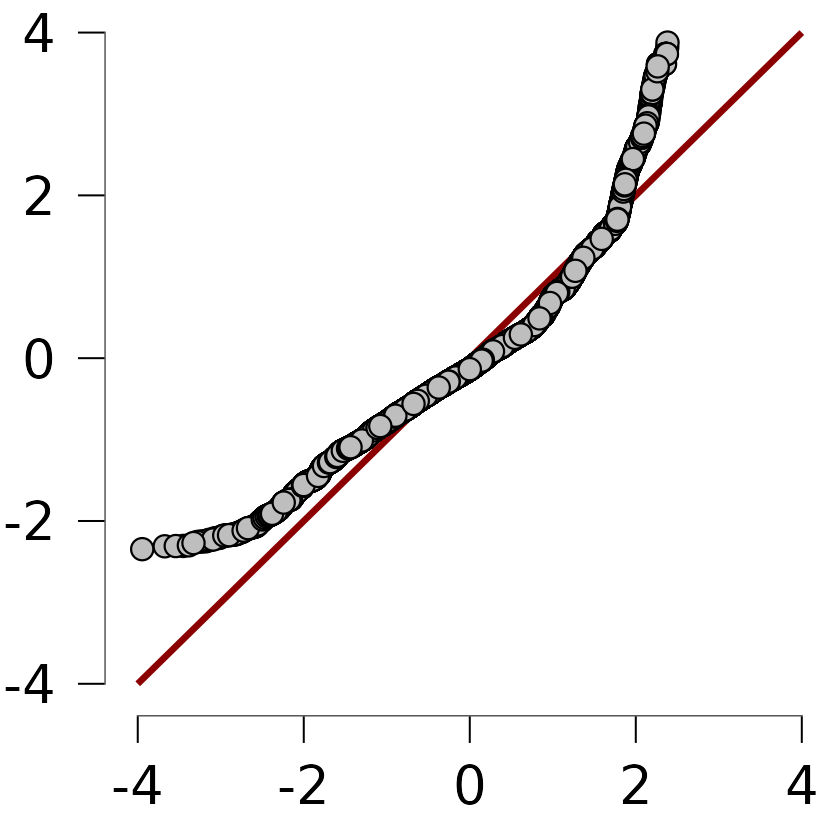}} &
		{\includegraphics[width=0.25\linewidth]{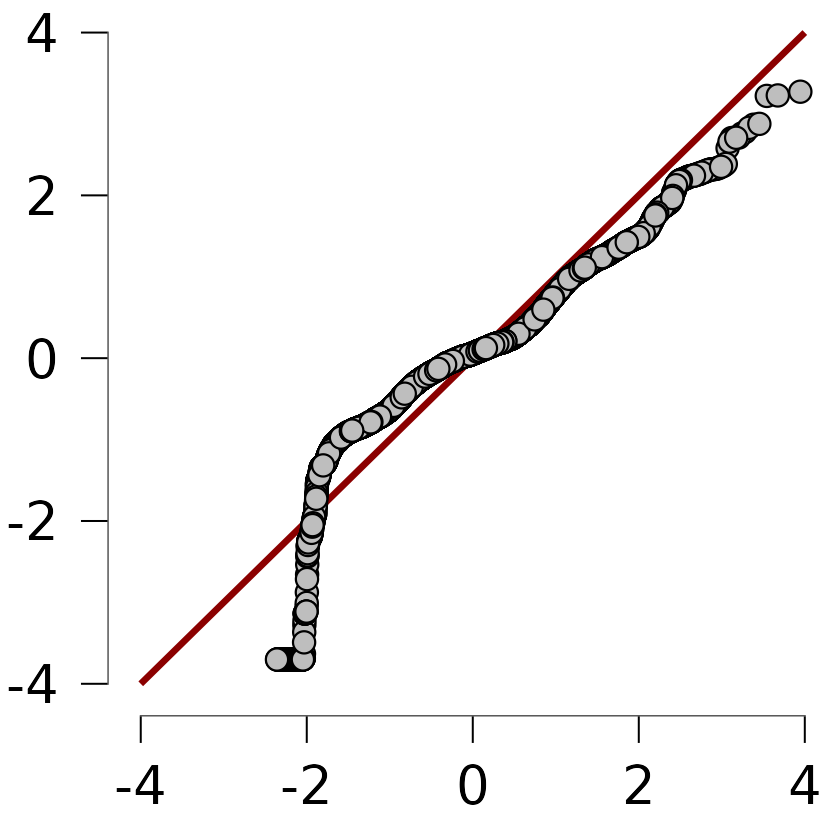}} \\
	\end{tabular}
	\begin{tabularx}{\linewidth}{p{0.3\linewidth}p{0.7\linewidth}}
		& \small Theoretical Quantiles\\
	\end{tabularx}
	\caption{Quantile-Quantile plots of the color detection dataset.}
	\label{fig:Q-Q}
\end{figure}

\begin{figure}[tbp]
	\centering
	\begin{tabularx}{\linewidth}{p{0.3\linewidth}p{0.28\linewidth}p{0.28\linewidth}}
		& \small RMSE & \small FR\\
	\end{tabularx}	
	\begin{tabular}{ccc}
		\small\rotatebox{90}{Standardized Residuals} &
		{\includegraphics[width=0.40\linewidth]{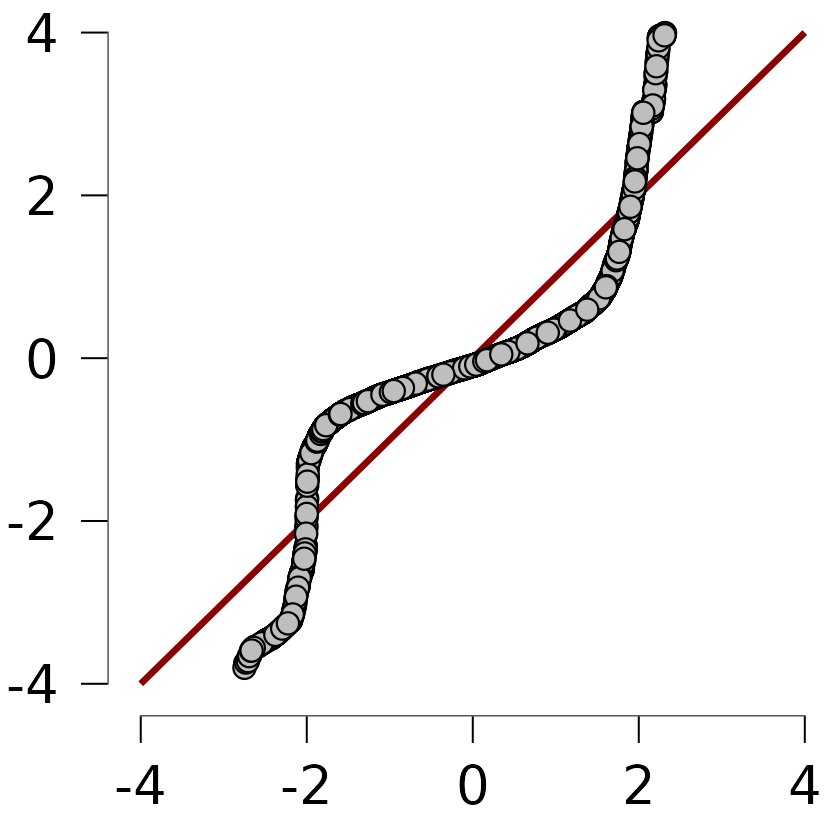}} &
		{\includegraphics[width=0.40\linewidth]{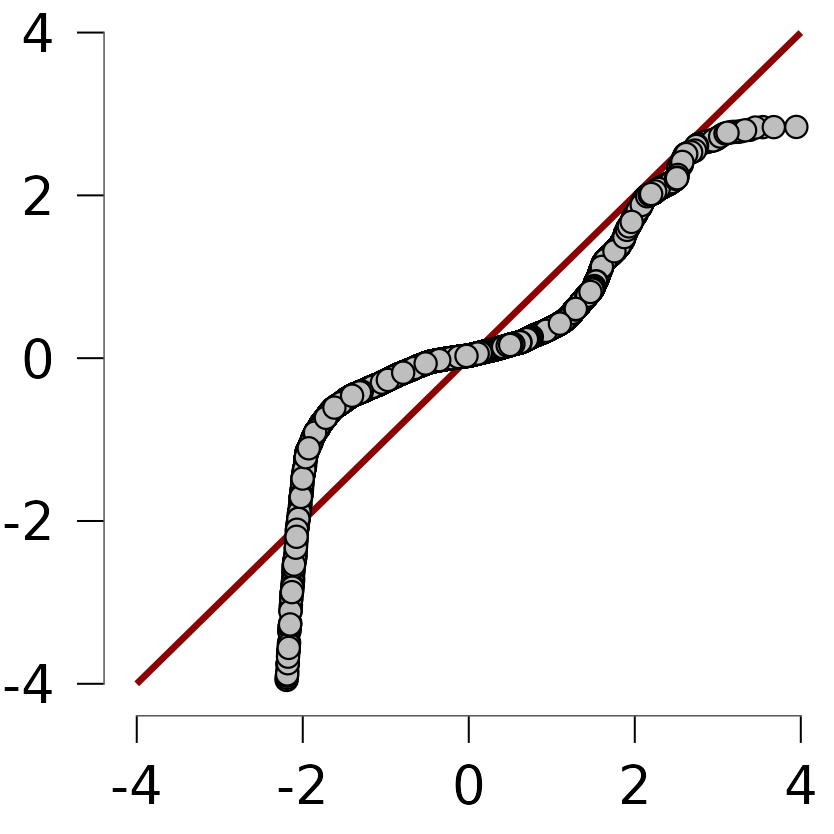}} \\
	\end{tabular}
	\begin{tabularx}{\linewidth}{p{0.3\linewidth}p{0.7\linewidth}}
		& \small Theoretical Quantiles\\
	\end{tabularx}
	\caption{Quantile-Quantile plots of the depth detection dataset.}
	\label{fig:Q-Q-depth}
\end{figure}

Analysis of Variance (ANOVA) (Tables~\ref{tab:ANOVA-PSNR-RGB}
to~\ref{tab:ANOVA-FR-Depth}) shows significant effects of both the Speed
and CP factors (as well as their interaction) on the dependent variables (PSNR,
RMSE, and FR). Residuals meet the normality assumption, supported by Q-Q plots
in Figures~\ref{fig:Q-Q} and~\ref{fig:Q-Q-depth}, validating the model. These results
confirm the statistical validity of our work. ANOVA tables were generated using JASP~\cite{jasp2024}.

\begin{table*}[htbp]
	\centering
	\caption{Analysis of Variance (ANOVA) for the Color PSNR}
	\label{tab:ANOVA-PSNR-RGB}
	\begin{tabular}{lrrrrr}
		\toprule
		Cases & Sum of Squares & df & Mean Square & F & p  \\
		\midrule
		Speed & $164027.653$ & $2$ & $82013.826$ & $11569.337$ & $<$ .001  \\
		CP & $33584.097$ & $5$ & $6716.819$ & $947.513$ & $<$ .001  \\
		Speed * CP & $6169.135$ & $10$ & $616.913$ & $87.025$ & $<$ .001  \\
		Residuals & $89192.490$ & $12582$ & $7.089$ &  &   \\
		\bottomrule
	\end{tabular}
\end{table*}

\begin{table*}[htbp]
	\centering
	\caption{Analysis of Variance (ANOVA) for the Color RMSE}
	\label{tab:ANOVA-RMSE-RGB}
	\begin{tabular}{lrrrrr}
		\toprule
		Cases & Sum of Squares & df & Mean Square & F & p  \\
		\midrule
		Speed & $225230.139$ & $2$ & $112615.070$ & $7267.745$ & $<$ .001  \\
		CP & $42003.544$ & $5$ & $8400.709$ & $542.150$ & $<$ .001  \\
		Speed * CP & $12562.719$ & $10$ & $1256.272$ & $81.075$ & $<$ .001  \\
		Residuals & $194960.457$ & $12582$ & $15.495$ &  &  \\
		\bottomrule
	\end{tabular}
\end{table*}

\begin{table*}[htbp]
	\centering
	\caption{Analysis of Variance (ANOVA) for the Color FR}
	\label{tab:ANOVA-FR-RGB}
	\begin{tabular}{lrrrrr}
		\toprule
		Cases & Sum of Squares & df & Mean Square & F & p  \\
		\midrule
		Speed & \mbox{$1.814\times10^{+6}$} & $2$ & $906934.152$ & $8884.279$ & $<$ .001  \\
		CP & \mbox{$1.897\times10^{+6}$} & $5$ & $379446.459$ & $3717.037$ & $<$ .001  \\
		Speed * CP & $658311.623$ & $10$ & $65831.162$ & $644.879$ & $<$ .001  \\
		Residuals & \mbox{$1.284\times10^{+6}$} & $12582$ & $102.083$ &  &  \\
		\bottomrule
	\end{tabular}
\end{table*}

\begin{table*}[htbp]
	\centering
	\caption{Analysis of Variance (ANOVA) for the Depth RMSE}
	\label{tab:ANOVA-RMSE-Depth}
	\begin{tabular}{lrrrrr}
		\toprule
		Cases & Sum of Squares & df & Mean Square & F & p  \\
		\midrule
		Speed & \mbox{$1.136\times10^{+7}$} & $2$ & \mbox{$5.680\times10^{+6}$} & $4288.971$ & $<$ .001  \\
		CP & \mbox{$6.380\times10^{+7}$} & $6$ & \mbox{$1.063\times10^{+7}$} & $8028.498$ & $<$ .001  \\
		Speed * CP & \mbox{$6.595\times10^{+6}$} & $12$ & $549596.715$ & $414.970$ & $<$ .001  \\
		Residuals & \mbox{$1.666\times10^{+7}$} & $12579$ & $1324.425$ & & \\
		\bottomrule
	\end{tabular}
\end{table*}

\begin{table*}[htbp]
	\centering
	\caption{Analysis of Variance (ANOVA) for the Depth FR}
	\label{tab:ANOVA-FR-Depth}
	\begin{tabular}{lrrrrr}
		\toprule
		Cases & Sum of Squares & df & Mean Square & F & p  \\
		\midrule
		Speed & $518315.917$ & $2$ & $259157.958$ & $3401.744$ & $<$ .001  \\
		CP & \mbox{$1.689\times10^{+6}$} & $6$ & $281427.178$ & $3694.053$ & $<$ .001  \\
		Speed * CP & $329420.234$ & $12$ & $27451.686$ & $360.335$ & $<$ .001  \\
		Residuals & $958316.716$ & $12579$ & $76.184$ &  &  \\
		\bottomrule
	\end{tabular}
\end{table*}




\end{document}